\documentclass[10pt,journal,compsoc]{IEEEtran}

\usepackage{times}
\usepackage{epsfig}
\usepackage{graphicx}
\usepackage{amsmath}
\usepackage{amsthm}
\usepackage{amssymb}
\usepackage{subfigure}
\usepackage{dsfont}
\usepackage{algorithmic}
\usepackage{stfloats}
\usepackage{color}
\usepackage{soul}
\usepackage{algorithm,algorithmic}

\usepackage{siunitx}
\usepackage{booktabs}

\usepackage[pagebackref=false,breaklinks=true,letterpaper=true,colorlinks,bookmarks=false]{}%{hyperref}
\graphicspath{{images/}}

\newcommand{\bx}{\mathbf{x}}

\DeclareMathOperator*{\argmin}{arg\,min}
\DeclareMathOperator*{\argmax}{arg\,max}

\newcommand{\RS}{{\bf Rand}}
\newcommand{\rpRS}{{\bf pRand}}
\newcommand{\FU}{{\bf FEnt}}
\newcommand{\CU}{{\bf CEnt}}
\newcommand{\PFU}{{\bf p*FEnt}}
\newcommand{\rPFU}{{\bf pFEnt}}
\newcommand{\PCU}{{\bf p*CEnt}}
\newcommand{\rPCU}{{\bf pCEnt}}
\newcommand{\FMinMax}{{\bf FMnMx}}
\newcommand{\FMinmargin}{{\bf FMnMar}}
\newcommand{\FEntS}{{\bf FEntS}}
\newcommand{\FEntC}{{\bf FEntC}}
\newcommand{\CEntS}{{\bf CEntS}}
\newcommand{\CEntC}{{\bf CEntC}}
\newcommand{\pFEntS}{{\bf p*FEntS}}
\newcommand{\pFEntC}{{\bf p*FEntC}}
\newcommand{\pCEntS}{{\bf p*CEntS}}
\newcommand{\rpCEntS}{{\bf pCEntS}}
\newcommand{\pCEntC}{{\bf p*CEntC}}

\newcommand{\featun}{{\bf FUn}}
\newcommand{\geomun}{{\bf GUn}}
\newcommand{\combun}{{\bf CUn}}

\begin{document}

\author{Ksenia~Konyushkova,
	Raphael~Sznitman,
	and~Pascal~Fua,~\IEEEmembership{Fellow,~IEEE,}% <-this % stops a space
	\IEEEcompsocitemizethanks{\IEEEcompsocthanksitem K.~Konyushkova and P.~Fua are with the Computer Vision Laboratory, EPFL, Lausanne, Switzerland. 
  E-mail: FirstName.LastName@epfl.ch
  \IEEEcompsocthanksitem R.~Sznitman is with  the ARTORG Center, 
  University of Bern, Bern, Switzerland.
  E-mail: raphael.sznitman@artorg.unibe.ch}
}
%\thanks{Manuscript received June 19, 2015; revised September 17, 2014.}}

\title{Geometry in Active Learning for Binary and Multi-class Image Segmentation}

\IEEEtitleabstractindextext{
	
	\begin{abstract}

We propose  an active  learning approach to image segmentation that  exploits geometric  priors to speed up and streamline the  annotation process. 
It can be applied for both background-foreground and multi-class segmentation tasks  in \num{2}D images and \num{3}D image volumes. 
Our approach combines geometric smoothness priors in the image space with more traditional uncertainty measures to estimate which pixels or voxels are the most informative, and thus should to be annotated next. 
For multi-class settings, we additionally introduce two novel criteria for uncertainty.
In the \num{3}D case, we use the resulting uncertainty measure to select voxels lying on a planar patch, which makes batch annotation much more convenient for the end user compared to the setting where voxels are randomly distributed in a volume. 
The planar patch is found using a branch-and-bound algorithm that looks for a \num{2}D patch in a \num{3}D volume where the most informative instances are located.
We evaluate  our approach on  Electron Microscopy and Magnetic  Resonance image volumes,  as  well  as  on  regular images of  horses  and  faces. We  demonstrate a substantial performance increase over other approaches thanks to the use of geometric priors.

\end{abstract}

}

\maketitle

%%\IEEEtitleabstractindextext{
%	\input{00-abstract}
%	% Note that keywords are not normally used for peerreview papers.
%	\begin{IEEEkeywords}
%		Active Learning, Multi-Class Active Learning, Image Segmentation, Branch-and-Bound
%	\end{IEEEkeywords}
%%}

\section{Introduction}
\label{sec:introduction}

Machine learning techniques are a key component of modern approaches to image segmentation, making the need for sufficient amounts of training data critical.
As far as images of everyday scenes are concerned, this is addressed by compiling large training databases and obtaining---at a high cost---the ground truth via crowd-sourcing~\cite{Kovashka16,Long13,Lin14a}.
By contrast, in specialized domains such as biomedical imaging, this is not always  an option  because only experts, whose time is scarce  and precious, can annotate images reliably. 
This stands  in the  way  of wide  acceptance  of many state-of-the-art segmentation algorithms, which are formulated in terms of a classification problem and  require large amounts of annotated data for training. 
The problem is even more acute for multi-class segmentation, which requires even larger training sets and more sophisticated interfaces to produce them~\cite{Joshi12}.

Active learning (AL)  is an established way to reduce  the annotation workload by automatically deciding  which parts of  the image  an annotator should  label to train the system with the minimal amount of manual intervention.
However,  most  AL  techniques used  in  computer  vision~\cite{Sun15a, Kading15, Beluch18} are designed for general classification tasks.
As such, these methods do not account for the specific difficulties or exploit the  opportunities that arise  when annotating individual  pixels in \num{2}D images and \num{3}D voxels  in image volumes.
Moreover, multi-class  classification has been studied relatively little in the AL setting despite its importance in numerous applications.

\begin{figure}[t]
 \begin{center}
\begin{tabular}{cc}
  \hspace{-0.3cm}\includegraphics[height=0.4\linewidth]{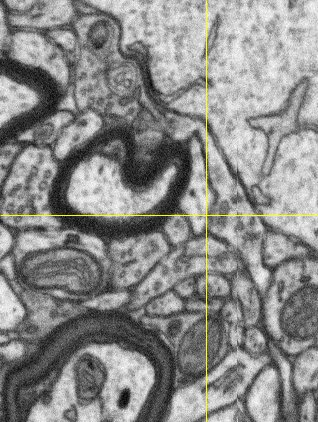}&
  \hspace{-0.3cm}\includegraphics[height=0.4\linewidth]{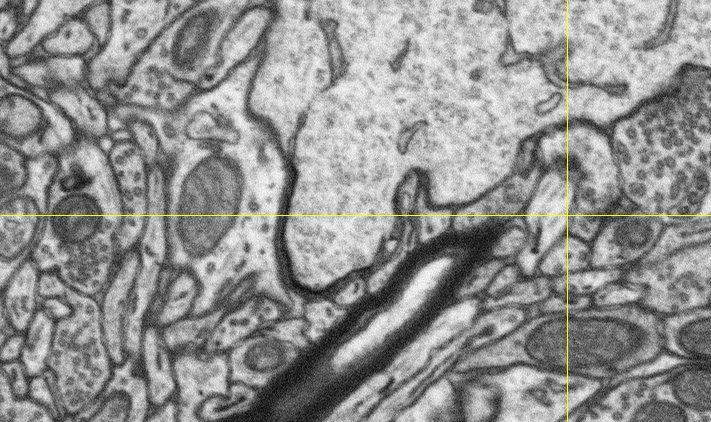}\\[-0.1cm]
  (yz)&(xy)\\
  \hspace{-0.3cm}\includegraphics[width=0.302\linewidth]{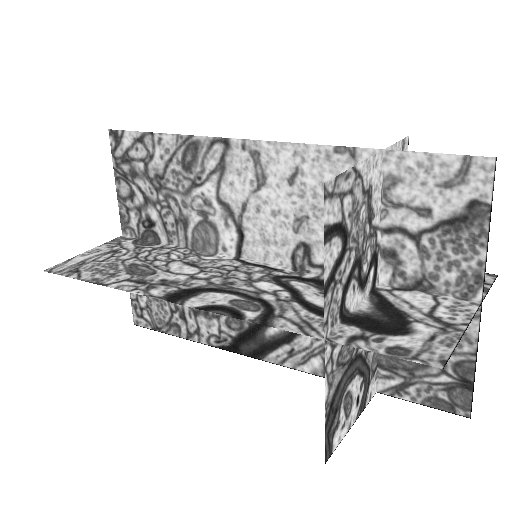}&
  \hspace{-0.3cm}\includegraphics[height=0.302\linewidth]{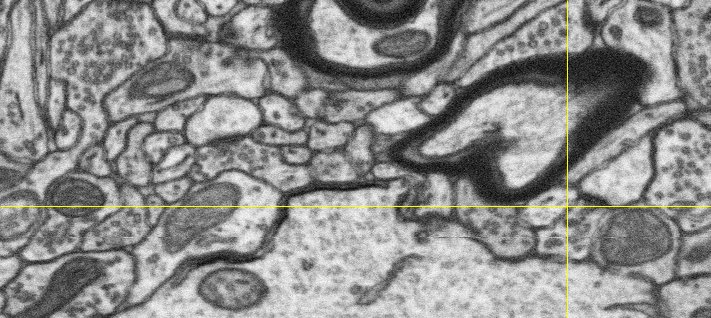}\\[-0.1cm]
  volume cut&(xz)\\
\end{tabular}
 \end{center}
\caption{Interface  of  the  FIJI   Visualization  API~\cite{Schmid10},  which  is
  extensively used to interact with 3D  image stacks. The user is presented with
  three  orthogonal planar  slices of  the stack.  While effective  when working
  slice  by slice,  this is  extremely cumbersome  for random  access to  voxels
  anywhere in the 3D stack, which is what a naive AL implementation
  would require.    }
\label{fig:01-fijiinterface}
\end{figure}
\vspace{0mm}

\begin{figure}[h]
  \begin{center}
      \includegraphics[width=0.4\linewidth]{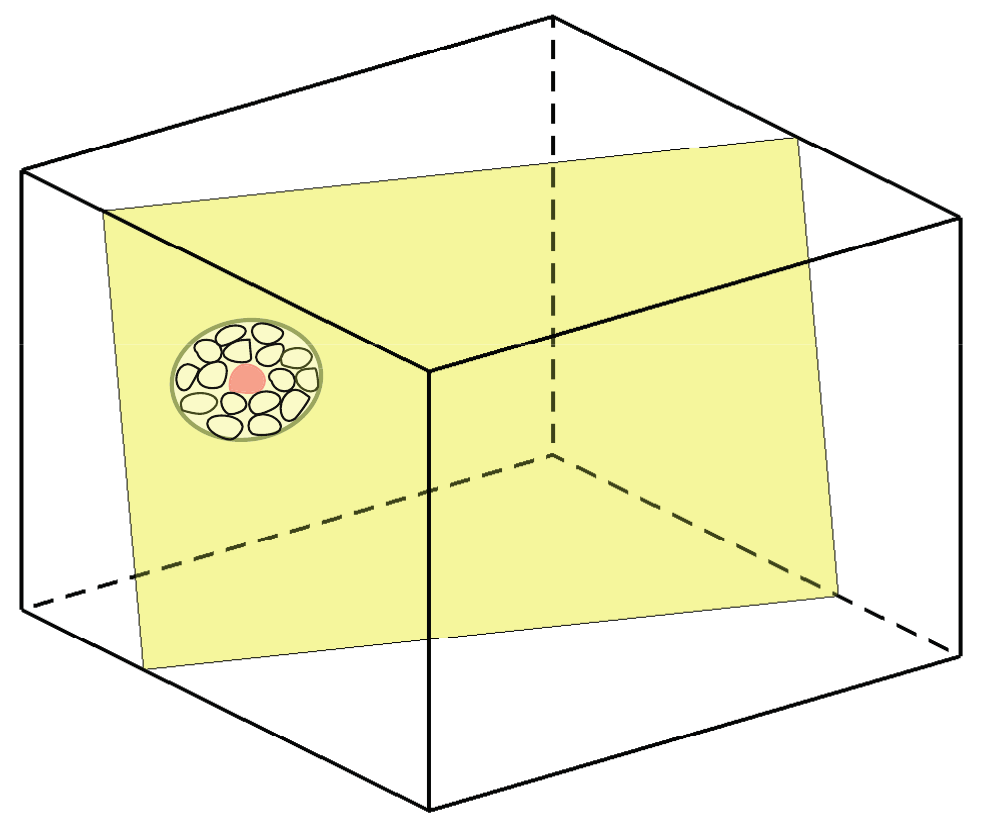} \\
      \includegraphics[width=0.49\linewidth]{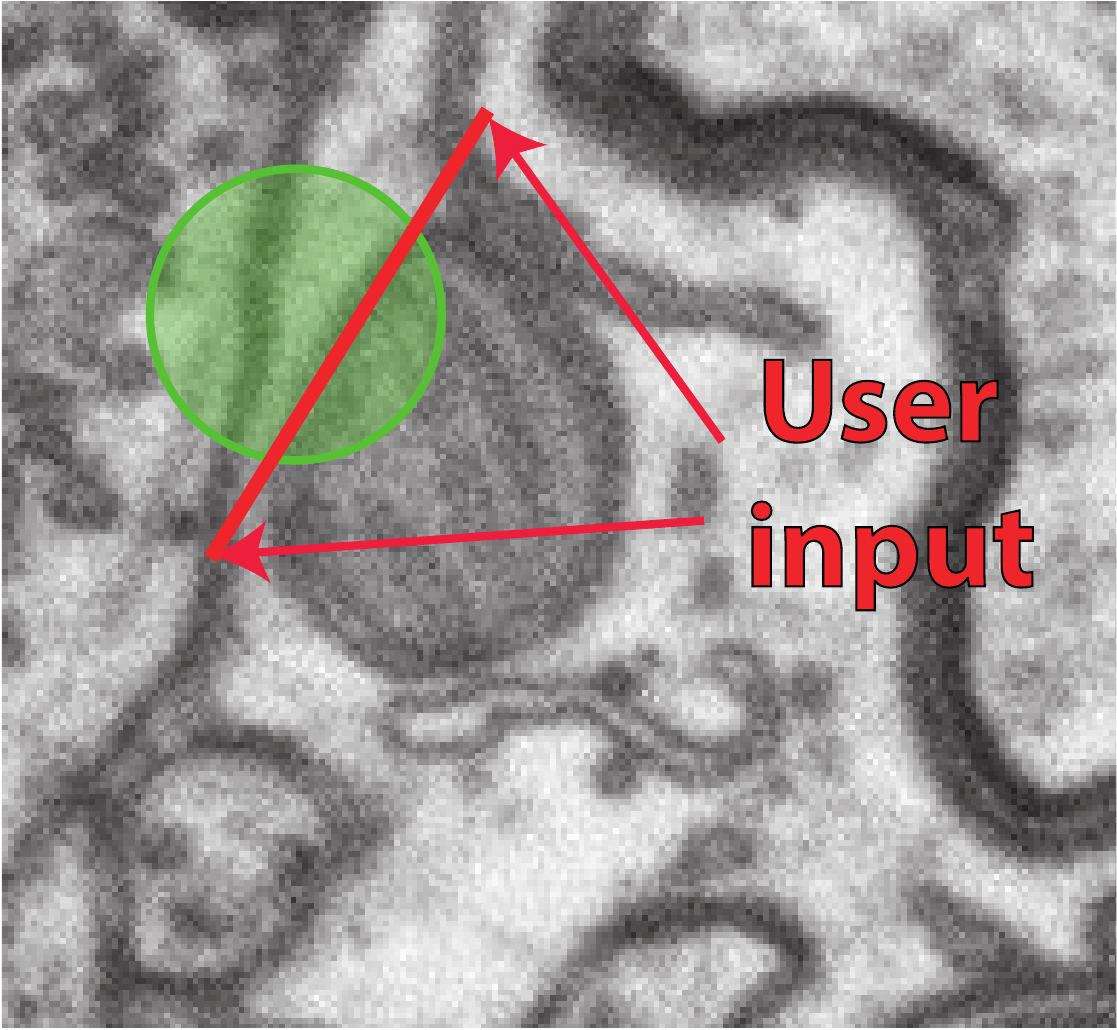}
      \includegraphics[width=0.49\linewidth]{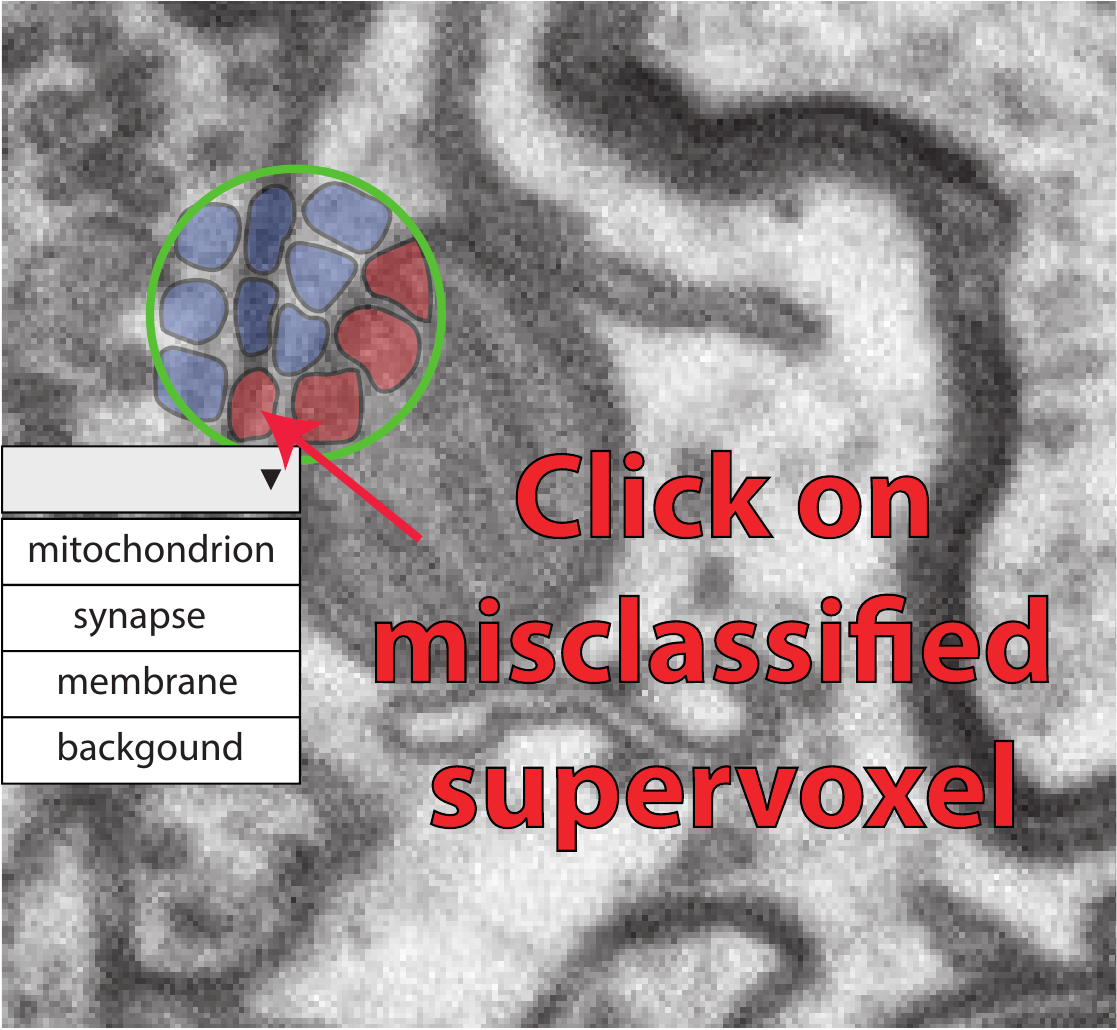}
  \end{center}
  \caption{Our approach to  annotation.  
Top row: The system selects an  optimal plane in an arbitrary orientation and presents the user  with a  patch that is  easy to annotate. 
The area to annotate is shown as  part of the full  3D stack.  
Bottom row: user interface, the planar patch the  user would see.
Left: in case of two classes present in the patch, it  could be annotated by clicking twice  to specify the  red segment that forms  the boundary between the  inside and outside of  a target object within the green circle. 
Right: the other way to annotate data is to correct mistakes in the current prediction. Supervoxels predicted to be mitochonria are shown in red, background in blue. If a user clicks on the misclassified supervoxel he can select the correct class among proposed.
Best viewed in color as most figures in this paper.}
\label{fig:01-planeInterface}
\end{figure}

In this paper we deal with image segmentation algorithms which require laborious annotations in the form of object masks.
\num{3}D stacks such as those depicted by Fig.~\ref{fig:01-fijiinterface} are common in the biomedical field and are
particularly challenging, because it is difficult for users to quickly figure out what they are looking at and annotate data efficiently.  
In this paper, we therefore introduce a novel approach to AL that is geared towards segmenting \num{3}D stacks while accounting  for geometric constraints of region shapes and thus making the annotation process convenient. 
Our approach  applies both  to  background-foreground and  multi-class segmentation of ordinary \num{2}D  images and \num{3}D volumes.
Our main contributions are as follows:
\begin{itemize}

\item  We exploit geometric priors  to select the image data for annotation more effectively, both for background-foreground and multi-class segmentation.
      
\item We define novel uncertainty measures for multi-class AL, which can be combined with the above-mentioned geometric priors.

\item We  streamline the annotation  process in  \num{3}D volumes so  that annotating them is no  more cumbersome than annotating ordinary \num{2}D  images, as depicted by Fig.~\ref{fig:01-planeInterface}.
    
\end{itemize}

The ideas on geometric uncertainty measures first appeared in \cite{konyushkova15}. 
Here, we extend them to the multi-class case and present in details the optimal branch-and-bound procedure for batch-mode AL.
In the  remainder of this paper,  we first  review current  approaches to  binary and  multi-class AL  and discuss why they are not necessarily the most effective when dealing with pixels and voxels. 
We then  give a short overview of our method before discussing in details the use of geometric  priors and how we  search for an  optimal cutting plane  to simplify  the annotation  process.
We then provide extensive experiments. 
We start by evaluating multi-class AL on image classification tasks and testing our geometric uncertainty with different classifiers.
Then, we compare our results against those of state-of-the-art techniques in a few challenging cases in image segmentation. 
Finally, we provide the experiments that illustrate the role of human intuition in the labelling procedure.

\vspace{-0mm}

\section{Related work and motivation}
\label{sec:related}

Image segmentation, both background-foreground and semantic segmentation, is a fundamental problem in computer vision. 
Training data for learning-based segmentation algorithms usually comes in a form of pixel-wise masks which are known to be very tedious to produce. 
In this paper, we are concerned with situations where domain experts are available to annotate images to train image segmentation algorithms.  
When experts' time is the most expensive/scarcest resource, AL is the technique of choice because it  seeks the smallest possible set of training samples to annotate    for effective model instantiation~\cite{Settles12}.

{\bf{Active learning (AL)}~~}
Typical AL strategies for query selection rely on uncertainty sampling~\cite{Lewis94}, which works remarkably well in many cases~\cite{Luo13, Sun15a}, query-by-committee, which does not require probability estimates~\cite{Giladbachrach05, Beluch18}, expected model change~\cite{Freytag14, Kading15}, or measuring information in the Fisher matrix~\cite{Hoi06}.
While  there is  a wide  range of  literature on  AL for  binary classification, multi-class problems are considered less often.   
Multi-class scenarios  are often approached  by reducing the problem to one-vs.-all binary classification~\cite{Kapoor07,Luo04}.  
Alternative methods rely on the expected reduction in misclassification risk and are independent of the number of classes~\cite{Joshi12}.    
Unfortunately, they can run in reasonable time only when they are combined with a classifier that has an easy model update rule for adding new samples. 
On the other side, for uncertainty  sampling,  one needs  to  redefine  the  notion of  uncertainty  or disagreement~\cite{Settles12, Joshi09, Long15a, Yang15, Jain09, Korner06}. 
Three ways to define the most uncertain datapoint are introduced in~\cite{Settles12}: (\num{1}) maximum entropy of  posterior probability distribution over classes, (\num{2}) minimal probability  of selected class and (\num{3}) minimal gap between the two most probable classes. 
There are many works relying on one of the above criteria or on combining them.
This includes selection uncertainty~\cite{Jain09}, posterior distribution entropy~\cite{Joshi09,Yang15},  the  combination of entropy, minimum margin and exploration criteria~\cite{Long15a}, and all three strategies of ~\cite{Settles12} together~\cite{Korner06}.

Recent work has demonstrated the effectiveness of data-driven AL approaches~\cite{Konyushkova17, Bachman17}. 
These methods learn strategies from available annotated data. 
It learns what kind of datapoints are the most beneficial for training the model given the current state of the classification problem. 
Then, past experience helps to derive more effective selection strategies.

{\bf{Batch-mode AL}~~}
Many interactive annotation pipelines suffer from long model update times which are necessary at every iteration.
Batch-mode selection has become a standard way to increase efficiency by asking the expert to annotate more than one sample at every iteration~\cite{Settles11, Yang13b, Luo13, Elhamifar13, Altaie14, Hasan15}. 
This procedure amortises the total retraining time over many annotations and enables to annotate data in parallel by several annotators.
Besides, in some situations it is easier for humans to provide labels to groups of examples~\cite{Johnson15}.
Density-based AL strategies often deal with the question how to form batches that ensure the diversity of the selection~\cite{Elhamifar13, Hasan15, Mosinska16}.
Moreover, batches can be formed with hierarchical clustering~\cite{Wigness15} and annotator's cognitive efforts can be taken into account~\cite{Vondrick13}.

{\bf{AL and image priors}~~} The AL techniques have been used for tasks in computer vision such as image classification~\cite{Joshi09, Kapoor07, Joshi12, Beluch18}, visual recognition~\cite{Long15a, Luo04}, semantic segmentation~\cite{Vezhnevets12,Iglesias11}, and foreground segmentation~\cite{Jain16b}.  
However, selection strategies are rarely designed to take advantage  of image specificities when labelling individual pixels or voxels, such as the fact that a neighborhood of pixels/voxels tends to have homogeneous labels.
The segmentation  methods presented in~\cite{Li11,Iglesias11,Zhou04} do take such geometric constraints into account for classification purposes, but not to guide AL, as we do.

Recently several authors realised the need to account for image properties in the AL selection for various computer vision tasks.
For example, in human pose estimation the uncertainty depends on the spatial distribution of the detected body joints~\cite{Liu12b}. 
In brain connectome reconstruction, an algorithm of Plaza\cite{Plaza16} can benefit from priors on how synapses can be situated in an image volume to result in a feasible reconstruction. 
Some methods~\cite{Hasan15, Jain16b, Paul17} account for the influence of neighbouring instances in AL selection by connecting datapoints in a graph as we do. 
However, the serious difference to our approach is that they add edges between datapoints in a graph based on their {\em feature similarity} and not their {\em geometric proximity} as we do.

{\bf{Human-computer interactions for segmentation}~~}
To understand how the cost of the annotation influences the final segmentation, Zlateski~\cite{Zlateski18} study the performance of a convolutional neural network depending on the amount and coarseness of the training labels.
In order to reach the same prediction quality, the number of annotations can be traded for their precision.
However, the performance improves when more time is spent on annotations.

Instead of pixel-wise masks some works try to adapt cheaper data modalities.
Scribbles (sparsely provided annotated pixels) are known to be very user-friendly to annotate images and video~\cite{Lin16a, Xu15, Nagaraja15}.
Scribbles annotations can be propagated from labelled to unlabelled pixels using a graphical model~\cite{Lin16a}.
In video segmentation, scribbles are propagated though the video while preserving its consistency~\cite{Nagaraja15}.
Another cheap annotation modality for segmentation is point-clicks~\cite{Bearman16, Bell15, Wang14f}.
Point-clicks on the object of interest can be incorporated into a weakly-supervised CNN with a special form of loss function~\cite{Bearman16}.
If an algorithm computes many hypotheses of the segmentation, the annotator can click on the object boundaries to eliminate wrong hypotheses~\cite{Jain16c}.
Polygons and pixel-wise masks are complementary label modalities: one-to-one correspondence between them can be easily established by assigning a mask to the area inside a polygon or by approximating the borders of a mask by a polygon. 
Then, the segmentation can be obtained either by predicting a class of pixels or by predicting the vertices of polygon with supervised~\cite{Castrejon17} or reinforcement learning~\cite{Acuna18}.
The polygon prediction task can involve the annotator to correct wrongly predicted vertices.

{\bf{Batch-mode AL and image priors}~~}
Batch-mode AL has been mostly investigated in terms of  semantic queries without due consideration to the fact that, in image segmentation, it is much easier for annotators to quickly label many samples  in  a localized  image  patch  than  having  to annotate  random image locations. 
We believe that for the efficient annotation pipeline in image segmentation, it is necessary to join the benefits of batch-mode AL and human-computer interactions.
If samples are distributed randomly in a \num{3}D  volume, it is  extremely cumbersome to labels them using current image display tools  such as the popular FIJI platform depicted by Fig.~\ref{fig:01-fijiinterface}. 
Thus, in  \num{3}D image  volumes~\cite{Li11,Iglesias11,Gordillo13}, it  is important  to provide the annotator  with a patch in  a well-defined plane, such as  the one shown  in Fig.~\ref{fig:01-planeInterface}. 
The technique of~\cite{Top11b} is an exception in  that it  asks users to label objects of interest in a plane of maximum  uncertainty.  
Our approach is similar, but has several distinctive features.  
First, the procedure we use to find the plane requires far fewer parameters to be set, as discussed in Sec.~\ref{sec:BatchGeom}.
Second, we search for the most uncertain patch in the plane and do not require the user to annotate the  whole plane.  
Finally, our approach can be used in conjunction with  an ergonomic interface that requires at most three mouse clicks per iteration when two classes are involved. 
Also, as  we show in the result section,  our method combined with geometric smoothness priors outperforms the earlier one.
\section{Approach}

We  begin by  broadly outlining  our framework,  which is  set in  a traditional AL context.  That is,  we  wish to  train  a classifier  for segmentation purposes,  but have initially  only few labelled and  many unlabelled training samples at our disposal.

AL seeks to find, iteratively and adaptively, a small set of training samples to be annotated for effective model training~\cite{Settles12}. 
In practice, this means that instead of asking an oracle to annotate all the data, we carefully select which datapoints should be labelled next based on what we know so far. 
The intelligent selection of data to be annotated can help to reach a good model performance using fewer labels.

\begin{figure}[t]
  \begin{center}
      \includegraphics[width=0.4\linewidth]{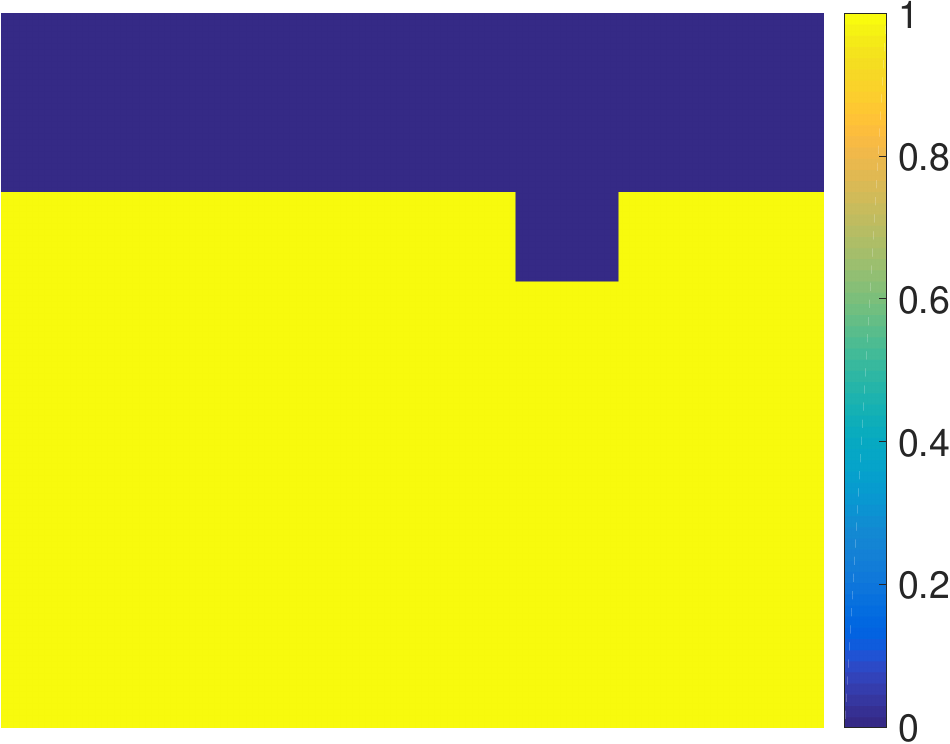} \hspace{10mm}
      \includegraphics[width=0.4\linewidth]{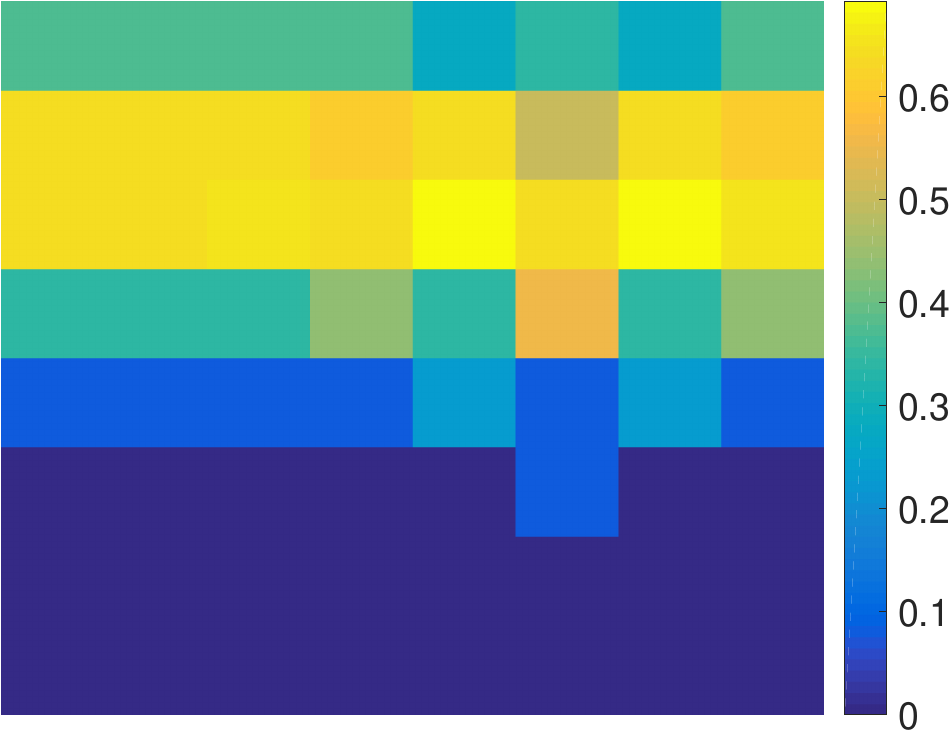}
  \end{center}
  \caption{Geometry-based uncertainty score. Left: predicted binary classification map for an $8 \times 8$ image. In this example the classifier assigns the pixels coloured in yellow to class $1$ with probability $1$ and pixels coloured in blue to class $0$, also with probability $1$. Feature uncertainty has the lowest possible uncertainty value for all pixels as the classifier is certain of its predictions. Right: geometric uncertainty score of Section~\ref{sec:GeomUncert}. The area of transition between the two classes is given a high geometric uncertainty score. Its maximum is reached where the boundary is not smooth.}
\label{fig:03-intuition}
\end{figure}

In our work the AL procedure is set in the context of image segmentation.
Since segmentation of \num{3}D volumes  is computationally expensive, supervoxels have been  extensively used  to speed  up the  process~\cite{Andres08,Lucchi11b}.  
In the remainder of this section and in Sec.~\ref{sec:GeomActive}, we will refer almost solely to supervoxels for simplicity but the definitions apply equally to superpixels when dealing with \num{2}D images.
We formulate our problem in terms of classifying supervoxels as a part of a specific target  object.  
As such, we start by  oversegmenting the image volume using the SLIC algorithm~\cite{Achanta12} and  computing for each resulting supervoxel $s_i$ a feature vector $\bx_i$. 
When dealing when with ordinary \num{2}D images, we simply replace the \num{3}D supervoxels with \num{2}D superpixels, which SLIC can also produce.
Our AL  problem thus involves  iteratively finding  the next set  of supervoxels
that  should be  labelled by  an expert  to improve  segmentation performance  as
quickly as possible.
To this end, our algorithm proceeds as follows:
\begin{enumerate}
  
\item Train a classifier on the labelled  supervoxels $S_L$ and  use  it  to  predict the class probabilities for the remaining  supervoxels  $S_U$.
  
\item Score $S_U$ on the basis  of a novel uncertainty function that we introduce in Sec.~\ref{sec:GeomActive}. It is inspired by the geometric properties of images in which semantically meaningful regions tend to have smooth boundaries. 
Fig.~\ref{fig:03-intuition} illustrates its behaviour given a simple prediction map: Non-smooth regions between various classes tend to be assigned the highest uncertainty scores.
 
\item In volumes, select a \num{2}D plane that contains a patch with the most uncertain supervoxels, as shown in Fig.~\ref{fig:01-planeInterface} and, in regular images, select a patch around the most uncertain superpixel. 
The expert can  then effortlessly label an indicated \num{2}D patch without having to examine the image data from multiple perspectives, as would be the case otherwise and as depicted by Fig.~\ref{fig:01-fijiinterface}. 
Furthermore, we can then design a simple interface that lets the user label supervoxel or superpixel batches with just a few mouse clicks, as shown in Fig.~\ref{fig:01-planeInterface} and described in Sec.~\ref{sec:exp}.

\item Sets $S_L$ and $S_U$ are updated and the  process is repeated until the segmentation quality is satisfactory.

\end{enumerate}

Compared to the standard AL procedure, our contribution lies in the way how we defined the uncertainty measure by relying on image priors (Sec.~\ref{sec:GeomActive}) and how we select a batch for annotation by designing an ergonomic interface to jointly present informative datapoints (Sec.~\ref{sec:BatchGeom}).

%% -*- mode: latex; mode: reftex; mode: flyspell; mode: auto-fill; TeX-master: "../top.tex"; -*-
% !TEX root =../top.tex 
% !TEX spellcheck = en-US

\section{Geometry-based active learning}
\label{sec:GeomActive}

Most AL  methods were  developed for  general tasks  and operate  exclusively in feature  space,  thus ignoring  the  geometric  properties  of images  and  more specifically their  geometric consistency. 
We start from Uncertainty Sampling (US). 
It is designed to focus the annotators' attention on samples for which  image features do not yet provide enough information for the classifier to decide what label to assign them. It selects samples that are {\it uncertain} in feature space to be annotated first so that classifier is updated with the largest amount of information. 
In short, US suggests labelling samples that are the most uncertain for the classifier, for example, closest to the classifier's decision boundary.
We will refer to this family of approaches as  {\it Feature  Uncertainty} (\featun{}). 
These methods are both effective and computationally inexpensive, thus, they are chosen as a basis of our work.
However, they do not account for image geometry to clue which samples may be mislabelled. 

To remedy this, we first introduce the  concept of {\it Geometric Uncertainty} (\geomun{}) and then show how to combine it with \featun.  
Our basic insight is that supervoxels that are assigned a label different from that of their neighbours ought to  be considered  more carefully  than those that  are assigned  the same label, as illustrated by  Fig.~\ref{fig:03-intuition}.
In this \num{2}D toy example, standard uncertainty in the feature space is low for all pixels because the classifier is confident in its predictions. 
On the contrary, in terms of geometry-based uncertainty measure, pixels near classification boundaries are uncertain and those near irregular parts of the boundary are even more uncertain. 
This corresponds to the intuition that object boundaries should be smooth.

We express both kinds of uncertainties in terms of entropy so that we  can combine them in a principled way. 
Doing this in multi-class segmentation case requires a new criterion for feature uncertainty, which we introduce below. 

\subsection{Uncertainty measures}
\label{sec:UncertMeas}

For each supervoxel $s_i$ characterised by feature vector $\bx_i$ and each possible label $\hat{y} \in Y$, let $p(y_i=\hat{y} \mid \bx_i)$ be  the probability that the label $y_i$ of $s_i$ is $\hat{y}$. 
In  this section we are not  concerned with the question  of how this probability  is obtained.  
For background-foreground  segmentation,  we  take  $Y$ to  be  $\{0,1\}$.   
In  the multi-class scenario, $Y$  is a larger set, such as  $\{$background, hair, skin, eyes, nose, mouth$\}$ for face  segmentation.  
We   describe  below  three entropy-based uncertainty  measures. 
We start  from the well-known Shannon Entropy of the predicted distribution over all possible classes and we then introduce two novel uncertainty measures which are both entropic in nature, but account for different properties of the predicted probability distribution.

\subsubsection{Total Entropy}
\label{sec:ShannonEntropy}

The simplest and most common way to estimate uncertainty is to
compute the Shannon entropy $\mathcal{H}$ of the total probability distribution over classes
\begin{align}
 \mathcal{H}(\{p(y_i=\hat{y} \mid \bx_i)\}) = \\ -\sum_{\hat{y} \in Y} p(y_i=\hat{y} \mid \bx_i) \log
 p(y_i=\hat{y} \mid \bx_i) \nonumber ,
 \label{eq:ShannonEntropy}
\end{align}
which we  will refer to as  {\it Total Entropy}. 
By definition, it is not restricted to the binary case and can be used straightforwardly in the multi-class scenario as well.

\subsubsection{Selection Entropy}
\label{sec:SelectEntropy}

When there are more than two elements  in $Y$, another way to evaluate uncertainty is to consider the label $b_1$ with highest probability: $p(y_i=b_1 \mid \bx_i) \ge p(y_i=b_j \mid \bx_i) \forall b_j \in Y, b_j \ne b_1$, against all others taken together: $p(y_i=\overline{b_1} \mid \bx_i) = \sum_{b_j \in Y \setminus b_1 } p(y_i=b_j \mid \bx_i)$.
For $b_k \in \{b_1, \overline{b_1}\}$ this yields a probability distribution
\begin{align}
 p_s = p(y_i=b_k \mid \bx_i).
\end{align}
Then, we compute the entropy  of the resulting probability distribution over two classes as {\it Selection Entropy} $\mathcal{H}_s$
\begin{align}
 \mathcal{H}_s = \mathcal{H}(p_s).
\label{eq:SelectEntropy}
\end{align} 
This definition of uncertainty is motivated by our desire to  minimize the number of
misclassified samples by concentrating on the classifier's decision output.
Notice that Selection Entropy avoids choosing the datapoints with the equal probability assigned to every class: $p(y_i=b_j \mid \bx_i) = p(y_i=b_l \mid \bx_i) \forall b_j, b_l \in Y$, when the number of classes is greater than two. 
This makes sense in practice because an example that is confused between all classes of the multi-class problem is likely to be an outlier.

\subsubsection{Conditional Entropy}
\label{sec:ConditEntropy}

Let $b_1$ and $b_2$ be two highest ranked classes for a supervoxel $s_i$, i.e. classes with the highest predicted probabilities: $p(y_i=b_1 \mid \bx_i) >  p(y_i=b_2 \mid \bx_i) > p(y_i=b_j \mid \bx_i), \forall b_j \neq b_1, b_2$. 
Another way to evaluate uncertainty in a multi-class scenario is to consider how much more likely class $b_1$ is than class $b_2$.
We assume that one of them is truly the correct class $y_i^*$, and we condition on this fact. 
For $\forall b_k \in \{b_1, b_2 \}$ this yields
\begin{align}
 p_c = p(y_i=b_k \mid \bx_i, y_i^* \in & \{b_1, b_2\}) = \\ &\frac{p(y_i=b_k \mid \bx_i)}{p(y_i=b_1  \mid  \bx_i)+p(y_i=b_2  \mid \bx_i)}
 \nonumber.
\end{align}
We then take the {\it Conditional Entropy} uncertainty to be the Shannon entropy of this probability distribution, which is
\begin{align}
 \mathcal{H}_c =  
 \mathcal{H}(p_c).
 \label{eq:ConditEntropy}
\end{align}
This definition of uncertainty is  motivated by the fact that the  classifier is rarely confused about  all possible  classes. 
More typically,  there are  two classes that are hard to distinguish and we want to focus on those \footnote{For example, when trying to  recognize digits from \num{0}  to \num{9}, it  is unusual to find  samples that resemble all possible classes with equal probability, but there are many cases in which \num{3} and \num{5} are not easily distinguishable. 
According to Selection Entropy, an example that is equally likely to be any of the digits should be avoided as a potential outlier.}

% !TEX root =../top.tex 
% !TEX spellcheck = en-US

\begin{figure*}[htbp]
\begin{center}
\includegraphics[width=0.9\linewidth]{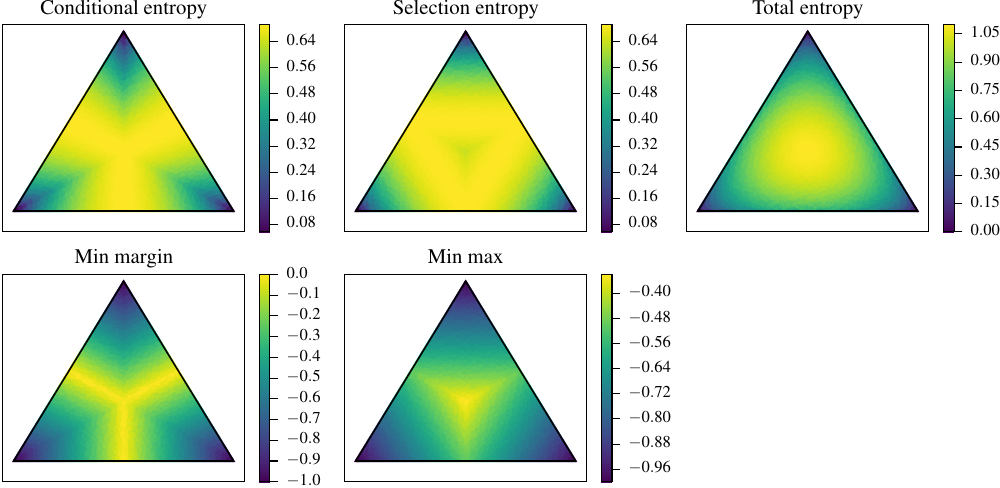}
\end{center}
\vspace{-0.2cm}
   \caption{Measures of Feature Uncertainty in a three-class problem. In each triangle the color denotes the uncertainty as a function of the three probabilities assigned to each class, which sum to 1. The three corners correspond to a point with probability \num{1} belonging to one of the three classes and therefore no uncertainty. By contrast, the center point can belong to any class with equal probability and therefore yields maximal uncertainty.  For better comparison, we inverted some values such that yellow corresponds to higher uncertainty and dark blue to the lower uncertainty. Top: entropy-based measures of Sec.~\ref{sec:UncertMeas}, Bottom: Measures proposed in~\cite{Settles12}.}
\label{fig:06-multiintuition}
\end{figure*}

\subsection{Feature Uncertainty (\featun)}
\label{sec:FeatUncert}

In  practice, we  estimate $p(y_i=\hat{y}  \mid \bx_i)$  by means  of a  classifier trained  using  parameters  $\theta$  and we  denote  the  distribution probability  by  $p_{\theta}$. 
Then,  any of  the  uncertainty  measures  from Sec.~\ref{sec:UncertMeas}  can  be  applied  to  the  probability  distribution $p_{\theta}(y_i=\hat{y} \mid \bx_i) \forall \hat{y}  \in Y$ resulting in {\it Feature Total Entropy} $\mathcal{H}$  from Eq.~\eqref{eq:ShannonEntropy},  {\it Feature  Selection Entropy} $\mathcal{H}_{s}$ from   Eq.~\eqref{eq:SelectEntropy}  and   {\it  Feature   Conditional  Entropy} $\mathcal{H}_{c}$  from Eq.~\eqref{eq:ConditEntropy}.
While all Feature Uncertainty measures are equivalent in the binary classification case, they behave quite differently in a multi-class scenario, as shown in the top row of Fig.~\ref{fig:06-multiintuition}. 
Furthermore, even though our {\it Selection Entropy} and {\it Conditional Entropy} measures are in the same spirit as the {\it Min margin} and {\it Min max} measures of~\cite{Settles12} (bottom row of Fig.~\ref{fig:06-multiintuition}), their selection is still different and they enable the combination with geometric priors, as we will shown in Sec.~\ref{sec:CombinedUncert}.
Next, we  will refer to  any one  of these three uncertainty measures as the Feature Uncertainty $\mathcal{H}^{\theta}$.

\subsection{Geometric Uncertainty}
\label{sec:GeomUncert}

Estimating the  uncertainty as  described above does not explicitly account for correlations between neighbouring supervoxels. 
To account  for  them, we  can estimate  the entropy of a different probability, specifically the probability that supervoxel $s_i$  belongs  to class  $\hat{y}$  given  the  classifier predictions  of  its neighbours and which we denote $p_G(y_i=\hat{y})$.

\begin{figure}[t]
\begin{center}
   \includegraphics[width=0.9\linewidth]{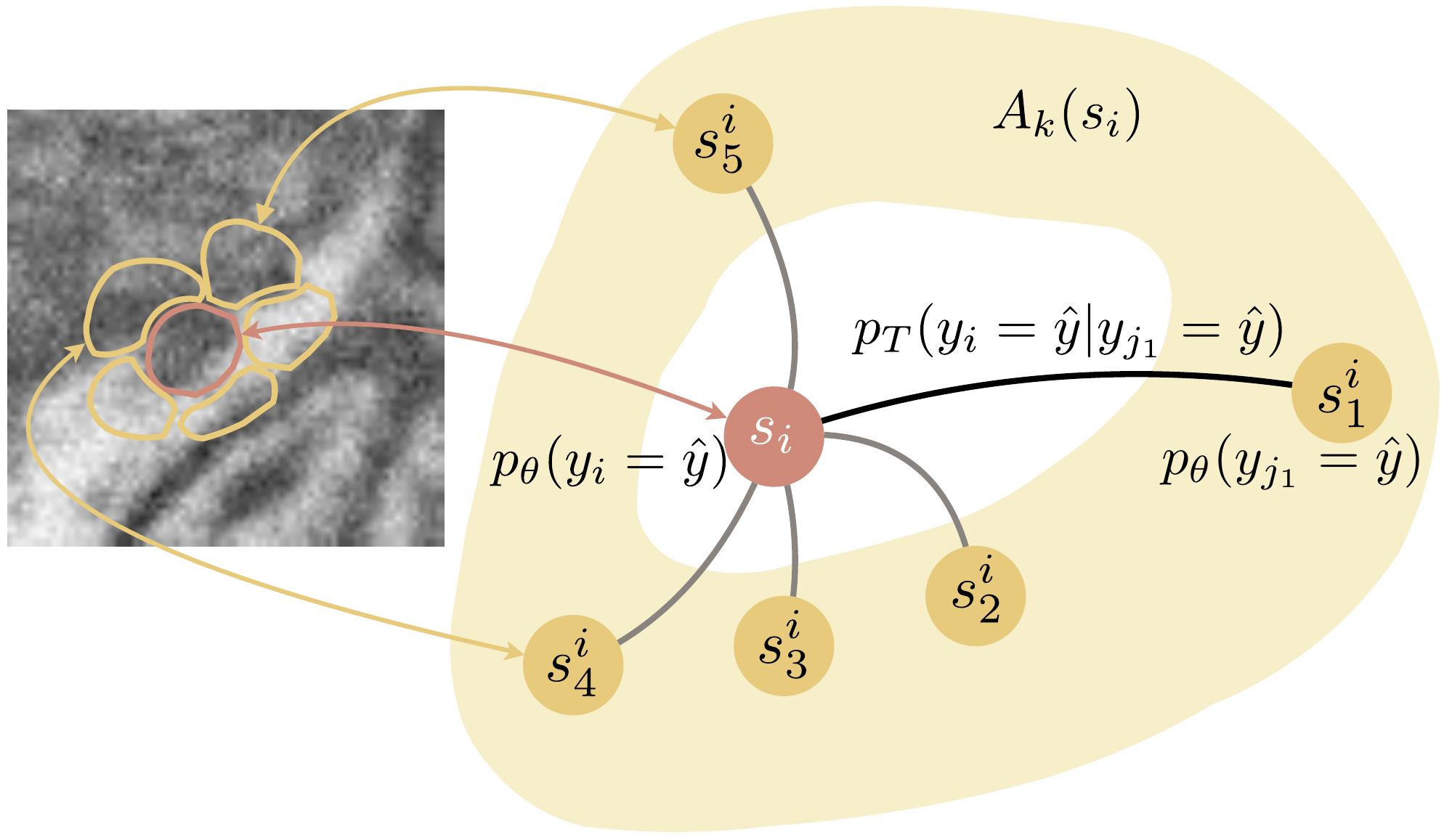}
\end{center}
   \caption{Image represented as a graph. We treat supervoxels as nodes in the graphs and edge weights between them reflect the probability of transition of the same label to a neighbour. 
   Supervoxel $s_i$ has $k$ neighbours from $A_k(i) = \{s_{1}^i, s_{2}^i, .., s_{k}^i\}$, 
$p_T(y_i=\hat{y} \mid y_j=\hat{y})$ is the probability of  node $s_i$ having  the same label as  node $s_j^i$, $p_{\theta}(y_i=\hat{y} \mid \bx_i)$ is the probability that $y_i$ , class of $s_i$, is $\hat{y}$, given only the corresponding feature vector $\bx_i$}
\label{fig:04-graph}
\end{figure}

To this  end, we treat  the supervoxels of  a single image  volume as nodes  of a directed
weighted  graph  $G$  whose  edges connect neighbouring  supervoxels,  as  depicted  in
Fig.~\ref{fig:04-graph}. We let $A_k(s_i)=\{s^i_{1},s^i_{2}, .., s^i_{k}\}$ be the set of $k$ nearest neighbours of $s_i$ and assign a weight to each edge inversely proportional to the Euclidean distance between the voxel centers. 
This approximation is the most precise when the supervoxels are close to being spherical, which is the case when using SLIC algorithm of~\cite{Achanta12}.
For each node $s_i$, we normalize the weights of all incoming edges so that   their    sum   is   one.
This allows to treat this quantity as the probability $p_T(y_i=\hat{y} \mid y_j=\hat{y})$ of node $s_i$ having  the same label as node $s_j^i \in A_k(s_i)$. 
It reflects our intuition that the closer two nodes are, the more likely they are to have the same label. 

To define $p_G(y_i=\hat{y})$ we use a random walk procedure on graph $G$~\cite{lovasz93}, as it reflects  well our
smoothness  assumption and has been extensively used for image segmentation purposes~\cite{Grady06,Top11b}. Given  the  $p_T(y_i=\hat{y} \mid y_j=\hat{y})$  transition probabilities,  we  can compute the probabilities $p_G$ iteratively by initially taking $p^0_{G}(y_i=\hat{y}) $ to be $p_\theta(y_j=\hat{y} \mid \bx_j)$
and then iteratively computing
  \begin{equation}
  p^{\tau+1}_{G}(y_i=\hat{y})                     =                     \sum_{s_j\in
    A_k(s_i)}p_T(y_i=\hat{y} \mid y_j=\hat{y})p^{\tau}_{G}(y_j=\hat{y}).
\label{eq:GeomProb}
  \end{equation}
\noindent
Note that $p_\theta(y_j=\hat{y} \mid \bx_j)$, $p^0_{G}(y_i=\hat{y})$ and $p^{\tau+1}_{G}(y_i=\hat{y})$ are  vectors whose dimension is  the cardinality of $Y$,  the set of all  possible labels.  
The procedure described by Eq.~\eqref{eq:GeomProb} is essential for defining the geometric uncertainty as it propagates the labels  of individual  supervoxels into  their neighborhood. 
Then, the supervoxels that are surrounded by neighbours whose predictions contradict to each other would receive contradicting scores and $p_{G}(y_i=\hat{y})$ would result in high uncertainty. 
The number  of iterations, $\tau_{max}$, defines  the radius of the neighborhood  involved in the computation of $p_G$ for $s_i$, thus encoding smoothness priors. 
When segmenting some objects of interest, $\tau_{max}$ should be chosen approximately as the average size of these objects. 
This enables score propagation inside the object, but does not favour propagation from parts that are far outside of object boundaries.
In our experiments, $\tau_{max}$ value varies between \num{10} and \num{20} depending on the application. 
Fig.~\ref{fig:03-intuition} shows the result of this computation for a simple \num{8x8} image  with initial prediction of a classifier as shown on the left and $k=4$ neighbours with equal edge weights.
We apply $\tau_{max}=4$ iterations and the resulting geometric uncertainty on the right shows how smoothness prior is reflected in the uncertainty: non-smooth boundaries receive the highest uncertainty score.

Given these probabilities, we can use the any of the uncertainty measures of Sec.~\ref{sec:ShannonEntropy}, ~\ref{sec:SelectEntropy} and~\ref{sec:ConditEntropy} to  compute the Geometric Uncertainty $\mathcal{H}^G$ for the probability distribution $p_{G}(y_i=\hat{y} \mid \bx_i) \forall \hat{y} \in Y$ as {\it Geometric Total Entropy} $\mathcal{H}$, {\it Geometric Selection Entropy} $\mathcal{H}_{s}$ and {\it Geometric Conditional Entropy} $\mathcal{H}_{c}$, respectively.

\subsection{Combining Feature and Geometric Entropy}
\label{sec:CombinedUncert}

Finally, given a trained classifier, we can estimate both \featun{} and \geomun{}.  To  use them  jointly, we  should in  theory estimate  the joint
probability distribution  $p_{\theta,G}(y_i=\hat{y} \mid x_i)$ and  the corresponding
joint entropy.   
As this is not modelled by our classification procedure, we take advantage of  the fact that  the joint  entropy is upper  bounded by the  sum of
individual entropies  $\mathcal{H}^{\theta}$ and  $\mathcal{H}^{G}$. 
Thus, for each supervoxel, we take the {\it Combined Uncertainty (\combun)} to be
\begin{equation}
  \mathcal{H}^{\theta,G} = \mathcal{H}^{\theta} + \mathcal{H}^G,
  \label{eq:CombEntropy}
\end{equation}
that is the upper bound of the joint entropy.  
The  same rule can be equally applied  to the Total entropy and entropy-based functions Selection Entropy and Conditional Entropy.
If we directly sum up feature and geometric scores defined by {\it Min margin} and {\it Min max} of Settles~\cite{Settles12}, the resulting measure does not have a physical meaning. 
The combined uncertainty measure is more appealing thanks to its upper bound interpretation\footnote{Besides, in practice, we confirmed in preliminary experiments that the proposed way to combine uncertainties results in faster learning rate than the combination of {\it Min margin} and {\it Min max} scores.}.

In practice, using this measure means that supervoxels that individually receive uncertain  predictions {\em and}  are in  areas of non-smooth transition between  classes will  be considered first. 
Note that the AL method of~\cite{Mosinska16} based on Zhou's propagation~\cite{Zhou04}, which is similar to the one we use, operates exclusively on $\mathcal{H}^G$. 
However, we experimentally observed on our datasets that considering the upper bound on the joint entropy from Eq.~\ref{eq:CombEntropy} results in a significant improvement in the learning curve compared to single Feature or Geometric uncertainty alone.
Our MATLAB implementation of {\it Combined Total Entropy} on \num{10} volumes of resolution  \num{176x170x220} of  the MRI dataset  from Sec.~\ref{sec:MRI} takes \num{1.4}s  per iteration (\num{2.3} GHz Intel Core i\num{7}, \num{64}-bit). 
The possibility of real-time performance is extremely important in the interactive applications.
\section{Batch-mode geometry query selection}
\label{sec:BatchGeom}

\begin{figure*}[h]
\begin{center}
\begin{tabular}{ccc}
   \includegraphics[width=0.3\linewidth]{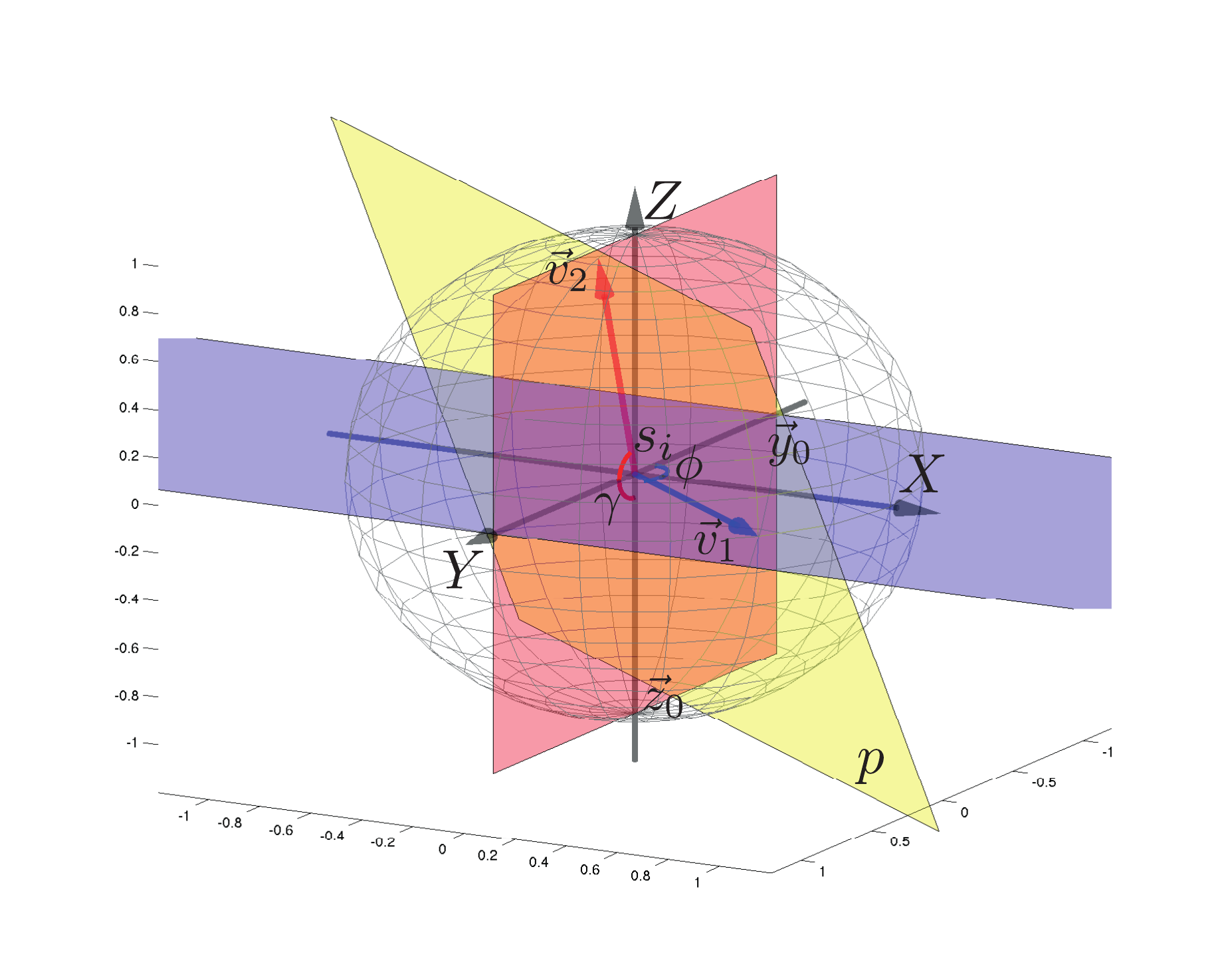} &
   \includegraphics[width=0.3\linewidth]{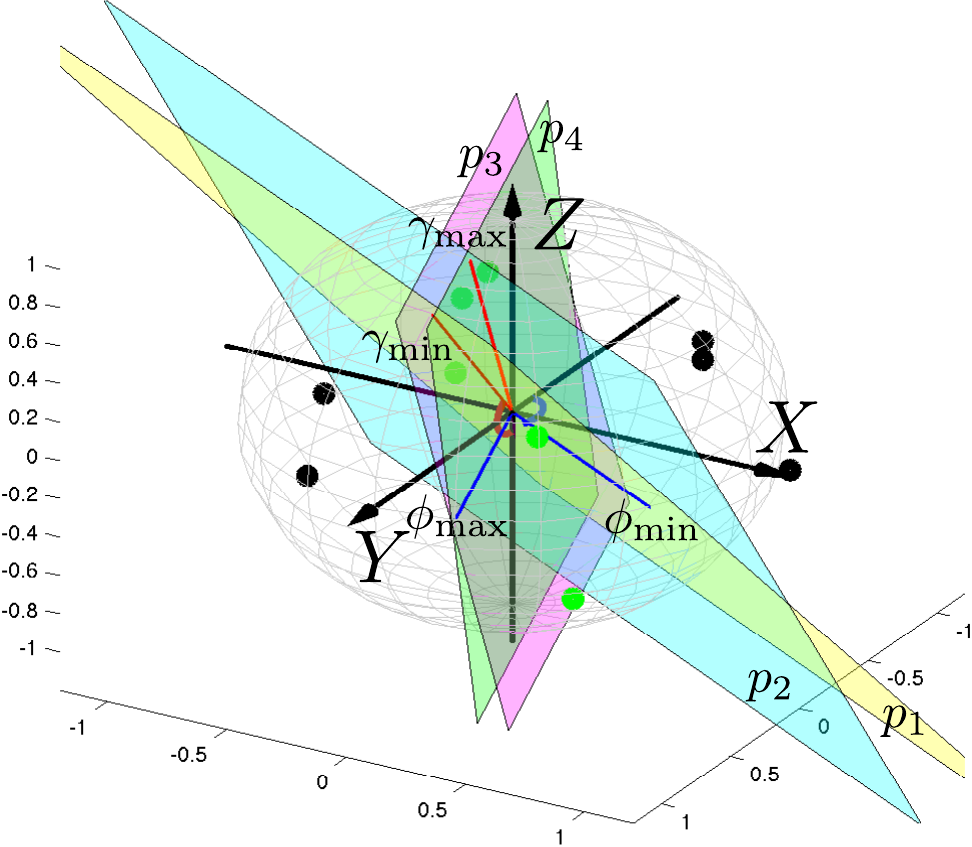} &
   \includegraphics[width=0.3\linewidth]{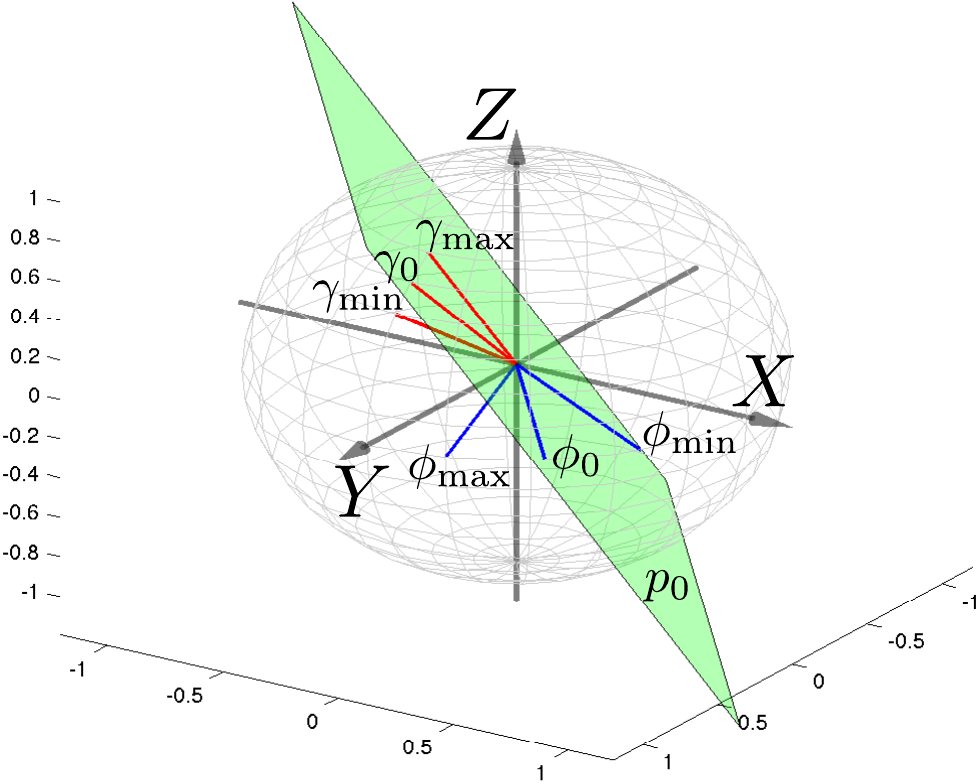} \\
   (a) & (b) & (c)
\end{tabular}
\end{center}
\vspace{-4mm}
   \caption{
   a) Parametrizing the search space. A circular patch is defined as the intersection of a  plane with a sphere. Plane $p_i$ (yellow) is parametrised by two angles, $\phi$ and $\gamma$; $\phi$ is the angle between the negative component of axis $-Y$ and plane intersection with $XY$ (blue), similarly, $\gamma$ is the angle between $-Z$ and plane intersection with $YZ$ (red). 
   b) Corridor. A corridor is a union of the areas between planes $p_1$ and $p_4$ as well as between $p_2$ and $p_3$. The green points depict supervoxels included in corridor $\Omega$ while black points depict supervoxels outside of it. 
   c) Bounding function and corridor splitting procedure. The score of the plane $p_0$ is less or equal to the score of all the points included between two planes $p_{\min}$ and $p_{\max}$: $U(p_0)<U(\Omega)$. We split the corridor $\Omega$ into corridors $[\phi_{\min}, \phi_0) \times [\gamma_{min}, \gamma_0)$, $[\phi_{\min}, \phi_0) \times [\gamma_0, \gamma_{\max})$, $[\phi_0, \phi_{\max}) \times [\gamma_{min}, \gamma_0)$ and $[\phi_0, \phi_{\max}) \times [\gamma_0, \gamma_{\max})$ and evaluate their uncertainty values. Among all available sectors we select a sector with the highest value to be split next. Best seen in color.}
   \label{fig:05-boundingFunction}
\end{figure*}

The  straightforward way  to exploit  the \combun{}  from Sec.~\ref{sec:CombinedUncert} would be  to pick  the most uncertain  supervoxel, ask the  expert to  label it, retrain the  classifier, and iterate.  
A more effective  way is  to find appropriately-sized batches of uncertain supervoxels and ask the expert to label
them  all   at  once  before   retraining  the  classifier.   
As   discussed  in Sec.~\ref{sec:related}, this  is referred to  as batch-mode selection, which usually reduces the time-complexity of AL. 
However, a naive implementation  would  force the  user to  randomly  view and  annotate several supervoxels in \num{3}D volumes regardless of where they are. 
This would not be user friendly as they would have to navigate a large image volume at each iteration.

In this  section, we therefore  introduce an  approach to select an uncertain  planar patch in \num{3}D volumes and allow the user to quickly label supervoxels within it, as shown in Fig.~\ref{fig:01-planeInterface}.   
We  allow annotator to  only consider circular regions  within planar  patches such as  the one depicted in Figs.~\ref{fig:01-planeInterface} and~\ref{fig:06-selection}.  
These can be  understood as  the intersection of  a sphere with  a plane  of arbitrary orientation.

Recall from  Sec.~\ref{sec:GeomActive},  that we  can  assign to  each supervoxel  $s_i$ an  uncertainty estimate  $U(s_i)$ in  one of  several ways.
Whichever one  we choose,  finding the circular  patch of  maximal uncertainty $p^\ast$ can be formulated as finding
\begin{equation}
p^\ast =  \argmax_{p} \sum_{s_j\in p} U(s_j),
\label{eq:optimization1}
\end{equation}
where the  summation occurs  over the  voxels that intersect  the plane  and are within the sphere.

Since Eq.~\eqref{eq:optimization1}  is linear  in $U(s_j)\geq  0$, we  design a branch-and-bound procedure to finding  the plane that yields  the largest  uncertainty.  
It  recursively eliminates  whole subsets  of planes and  quickly converges to the  correct solution. 
Whereas an  exhaustive search would be excruciatingly slow, our current MATLAB implementation on MRI dataset takes $0.024$s  per plane search with the same settings as in Sec.~\ref{sec:CombinedUncert}.
This means that  a C implementation of the entire pipeline would be real-time, which  is critical for  acceptance of such an interactive  method by users.

As  discussed  above,  this  procedure  could  be  used  in conjunction with any  one of the uncertainty measures defined  in the previous section.  
We will as shown in Sec.~\ref{sec:exp} that it  is most beneficial when   used   in   combination    with   the   geometry-aware   criterion   of Sec.~\ref{sec:CombinedUncert}. We  describe  our  branch-and-bound  plane-finding procedure  in  more  detail below and in Section~\ref{sec:2Dprocedure} we show a simplified example of this procedure in $2$D.

\subsection{Parametrizing the search space}
\label{sec:ParamSeachSpace}

Let us  consider a spherical volume centered at  supervoxel $s_i$, such as the  one depicted by  Fig.~\ref{fig:05-boundingFunction}a.
Since the  SLIC superpixels/supervoxels  are always  roughly circular/spherical, any supervoxel  $s_j$ can be well  approximated by a sphere of radius $\kappa$, set to a constant for a  particular dataset, and its center $w_j$.  
We will refer to this approximation as $\hat{s}_j$. Then,  every $\hat{s}_j=(w_j,\kappa)$ is characterized by its center $w_j$  and the common radius $\kappa$.
   
Let $\hat{S}_i^r$  be the set  of supervoxels within the distance $r$ from $\hat{s}_i$, that is,
\begin{equation}
\hat{S}_i^r = \{ \hat{s}_j = (w_j, \kappa) \mid \Vert w_j-w_i \Vert \le r \}.
\end{equation}
If we  take the  desired patch  size to  be $r$,  we can  then operate exclusively on the elements of  $\hat{S}_i^r$. 
Let $P_i$  be the set  of all planes passing through the center of  $\hat{s}_i$. 
As we will  see below, our procedure requires defining  planes, area splits of  approximately equal size, and supervoxel membership  to certain areas and planes. 
We parametrize planes in $P_i$ as follows.

Let  us consider  a plane  $p  \in P_i$,  such as  the  one shown  in yellow  in Fig.~\ref{fig:05-boundingFunction}a.  
It  intersects the $XY$ plane  along   a  line  characterized   by  a  vector  $\vec{v}_1$,   shown  in blue. 
Without loss  of generality, we can choose the  orientation of $\vec{v}_1$ such that its  $X$ coordinate is positive  and we denote by $\phi$  the angle between the negative component of axis $-Y$  and $\vec{v}_1$. 
Similarly, let us consider the  intersection of  $p$ with  $YZ$  plane and  characterize it  by the  vector $\vec{v}_2$ (shown in  red) with a  positive $Y$ coordinate.  
Now let $\gamma$ be the  angle between $-Z$ and  $\vec{v}_2$.  
We can now  parametrize the plane $p$ by  the two angles  $\phi \in  [0, \pi)$ and  $\gamma \in [0,  \pi)$ because there is  one and only one  plane passing through two  intersecting lines.  
We will refer  to $(\phi, \gamma)$  as the plane's angular  coordinates.  
Finally, let  $C_i^r(p)$ be  the  set  of  supervoxels  $\hat{s}_j  \in \hat{S}_i^r$ lying on $p$, that is,
\begin{equation}
C_i^r(p) =  \{ \hat{s}_j \in  \hat{S_i^r} \mid \text{distance(} p,  w_j \text{)}
\le 2 \kappa \}.
\end{equation}

The  set $P_i$ can  be represented by  the Cartesian product  $[0, \pi) \times  [0,   \pi)$  of  the   full  ranges   of  $\phi$  and   $\gamma$.  
Let $\Phi=[\phi_{\min}, \phi_{\max})$  and $\Gamma=[\gamma_{\min}, \gamma_{\max})$ be two angular intervals. 
We  will refer  to a  set of  planes with  angular coordinates  in $\Phi \times \Gamma$ as the {\it corridor} $\Omega =  \Phi \times \Gamma$, as illustrated by Fig.~\ref{fig:05-boundingFunction}b. 
The boundaries of this corridor are defined by four planes shown in Fig.~\ref{fig:05-boundingFunction}b: $p_1 = (\phi_{\min}, \gamma_{\min})$, $p_2 = (\phi_{min}, \gamma_{\max})$, $p_3 = (\phi_{\max}, \gamma_{min})$ and $p_4 = (\phi_{\max}, \gamma_{\max})$.

\subsection{Searching for the best bisecting plane}
    
\subsubsection{Uncertainty of planes and corridors}
\label{sec:planeSectorUncertainty}

Recall  that we  assign  to  each supervoxel  $\hat{s}_j$  an uncertainty  value $U(\hat{s}_j)\geq 0$. We take the uncertainty of plane $p$ to be
\begin{equation}
U(p) = \sum\limits_{\hat{s}_j \in C_i^r(p)} U(\hat{s}_j).
\end{equation}
Finding a circular patch $p^*$ of maximum uncertainty then amounts to finding
\begin{equation}
p^* = (\phi^*, \gamma^*) = \underset{p \in P_i}{\arg\max}\ U(p).
\label{eq:optimization}
\end{equation}

Similarly, we  define the uncertainty of  a corridor as the sum of  the  uncertainty values  of  all  supervoxels lying  between  the four  planes  bounding it, between $p_1$ and $p_4$, and between $p_2$ and $p_3$ as depicted by green points in Fig.~\ref{fig:05-boundingFunction}b.
We therefore write 
\begin{equation}
U(\Omega) = \sum\limits_{\hat{s}_j \in C_i^r(\Omega)} U(\hat{s}_j),
\label{eq:sectorUncertainty}
\end{equation}
where $C_i^r(\Omega)$ represents the supervoxels lying between the four bounding planes.
In practice, a supervoxel is considered to belong to the corridor if its center lies either between $p_1$ and $p_4$  or between $p_2$ and $p_3$, or is no further than $\kappa$ away from any of  them. 
When the angles are acute, the membership of a voxel is decided  by checking that the  dot product of the  voxel coordinates with the plane normals  have the same sign, provided that  the normals orientations are chosen so that they all point inside the corridor.

\subsubsection{Branch-and-bound}

To solve Eq.~\ref{eq:optimization} and find the optimal circular patch, we use a branch-and-bound approach.  
It involves  quickly eliminating entire  subsets of the parameter space $\Phi \times \Gamma$ using a bounding function~\cite{Lampert08},  a  recursive  search procedure,  and  a  termination criterion, which we describe below.

{\bf{Bounding  function}~~} Let us again consider the corridor $\Omega=[\phi_{\min},\phi_{\max})\times[\gamma_{\min},\gamma_{\max})$  bounded by  the  four  planes  $p_1$  to  $p_4$. 
Let  us  also  introduce  the  plane $p_0=(\alpha_1 \phi_{\min}+\beta_1 \phi_{\max}, \alpha_2 \gamma_{\min}+\beta_2 \gamma_{\max})$, where $\alpha_1+\beta_1 =  1, \alpha_2+\beta_2=1$ depicted by Fig.~\ref{fig:05-boundingFunction}c.
Given that  $U(\hat{s}_j)\geq0$ and  that Eq.~\ref{eq:optimization}  is linear  in  $U(\hat{s}_j)$,  the uncertainty  of $p_0$ will always be less or equal to that of $\Omega$.
This allows us  to bound the uncertainty  of any plane from above  and to search for the solution only within the most promising parameter intervals.

{\bf{Search procedure}~~} As in work of~\cite{Lampert08}, we maintain a priority queue $L$ of corridors. At each iteration, we  pop the corridor $\Omega_{\max}^j = [p_1^j, p_2^j,  p_3^j,  p_4^j]$  with  the  highest  uncertainty  $U(\Omega_{\max}^j)$ according to Eq.~\ref{eq:sectorUncertainty} and process it as follows.

We  introduce two  new angles  $\phi_0^j =  (\phi_{\min}^j+\phi_{\max}^j)/2$ and $\gamma_0^j  =  (\gamma_{\min}^j+\gamma_{\max}^j)/2$   and  split  the  original parameter intervals into two, as shown in Fig.~\ref{fig:05-boundingFunction}c.  
We compute  the  uncertainty  of four corridors: $[\phi_{\min},  \phi_0) \times  [\gamma_{min}, \gamma_0)$,  $[\phi_{\min},   \phi_0) \times  [\gamma_0,   \gamma_{\max})$,  $[\phi_0, \phi_{\max}) \times  [\gamma_{min}, \gamma_0)$  and  $[\phi_0, \phi_{\max}) \times  [\gamma_0, \gamma_{\max})$, and add them to the priority queue $L$. 

Note, that  we always  operate on  acute angles after  the first  iteration with initialization $[0; \pi)$, which allows us  to compute the uncertainty scores of corridors as discussed in Section~\ref{sec:planeSectorUncertainty}.

{\bf{Termination condition}~~} The search procedure terminates  when the bisector plane   $p_0   =    (\phi_0,   \gamma_0)$   of   the    corridor $C_i^r(\Omega_{\max}^j)$  touches  all the  supervoxels  from  the corridor.  
To fulfil this condition it  is enough to ensure that the  distance from any point 
in the corridor to a bisector plane is within the offset $2 \kappa$, that is,
\begin{equation}
\text{distance(} p_0,  \hat{s}_l \text{)} \le  2 \kappa , \forall  \hat{s}_l \in
\Omega_{\max}^j.
\end{equation}
Since  $U(p_0)$ is greater than the uncertainty of all the remaining corridors, which  is itself greater  than that of  all planes they  contain (including the boundary cases),  $p_0$ is guaranteed to  be the optimal plane  we are looking for.

\subsubsection{Global optimization}

Our branch-and-bound  search is relatively  fast for a single  voxel but not  fast enough  to  perform it for  all  supervoxels in  a  stack. 
Instead, we restrict our search to $t$ most uncertain supervoxels in the volume.

We  assume   that  the  uncertainty   scores  are  often  consistent   in  small neighbourhoods, which  is especially true  for the geometry-based  uncertainty of Section~\ref{sec:GeomUncert}. 
By doing so it enables us to find a solution that is close  to the optimal one  with a low value  of $t$. 
In this  way, the final algorithm first takes  all supervoxels $S$ with uncertainty $U$  and selects the top $t$ locations. 
Then, we  find the  best  plane for  each of  the top  $t$ supervoxels and choose the best plane among them.

\subsection{Illustration of search procedure in \num{2}D}
\label{sec:2Dprocedure}

\begin{figure}[t]
  \begin{center}
    \begin{tabular}{c@{\hskip -1mm} c@{\hskip -1mm} c@{\hskip -1mm}}
	  \includegraphics[width=0.34\linewidth]{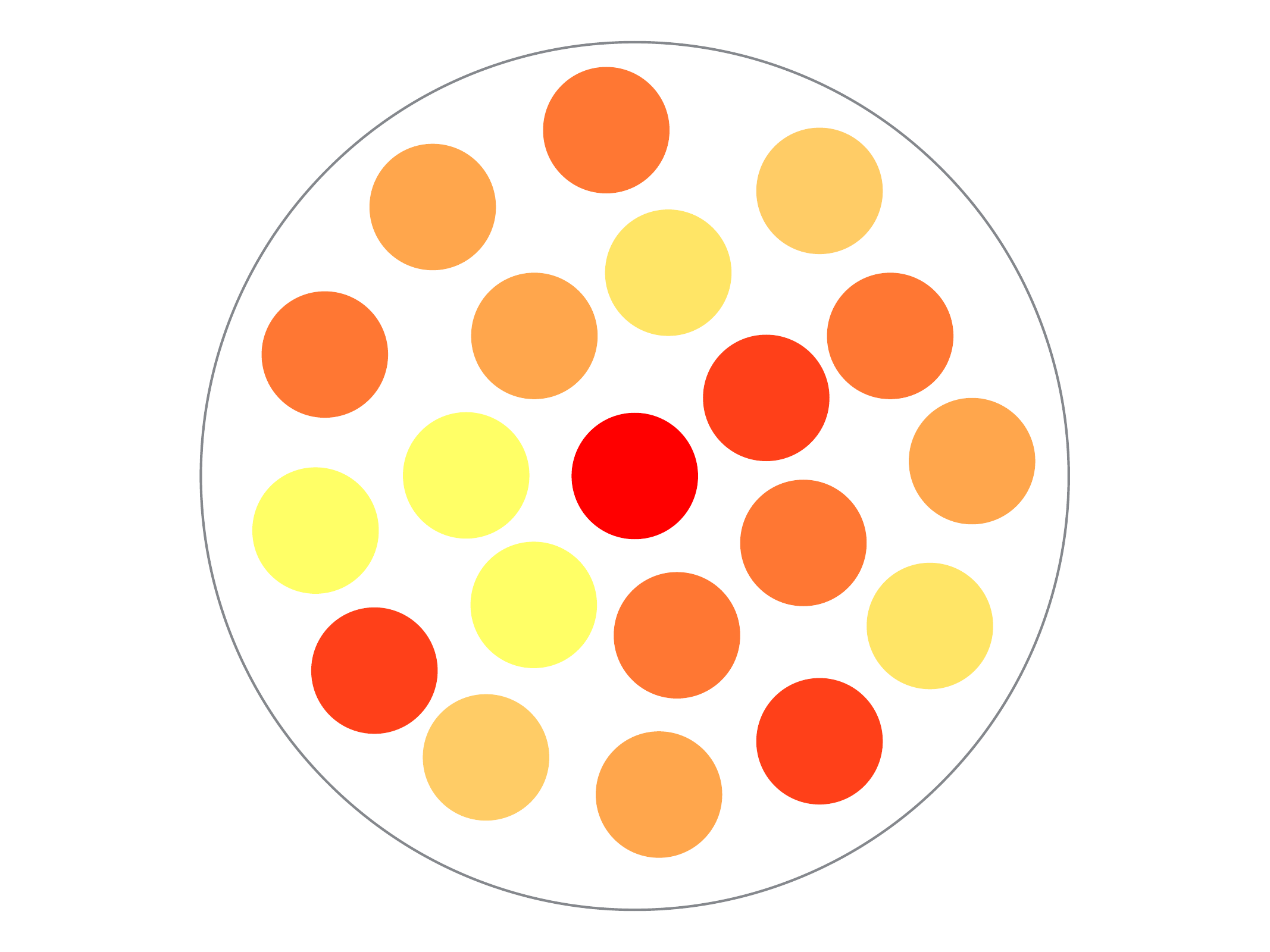} &
      \includegraphics[width=0.34\linewidth]{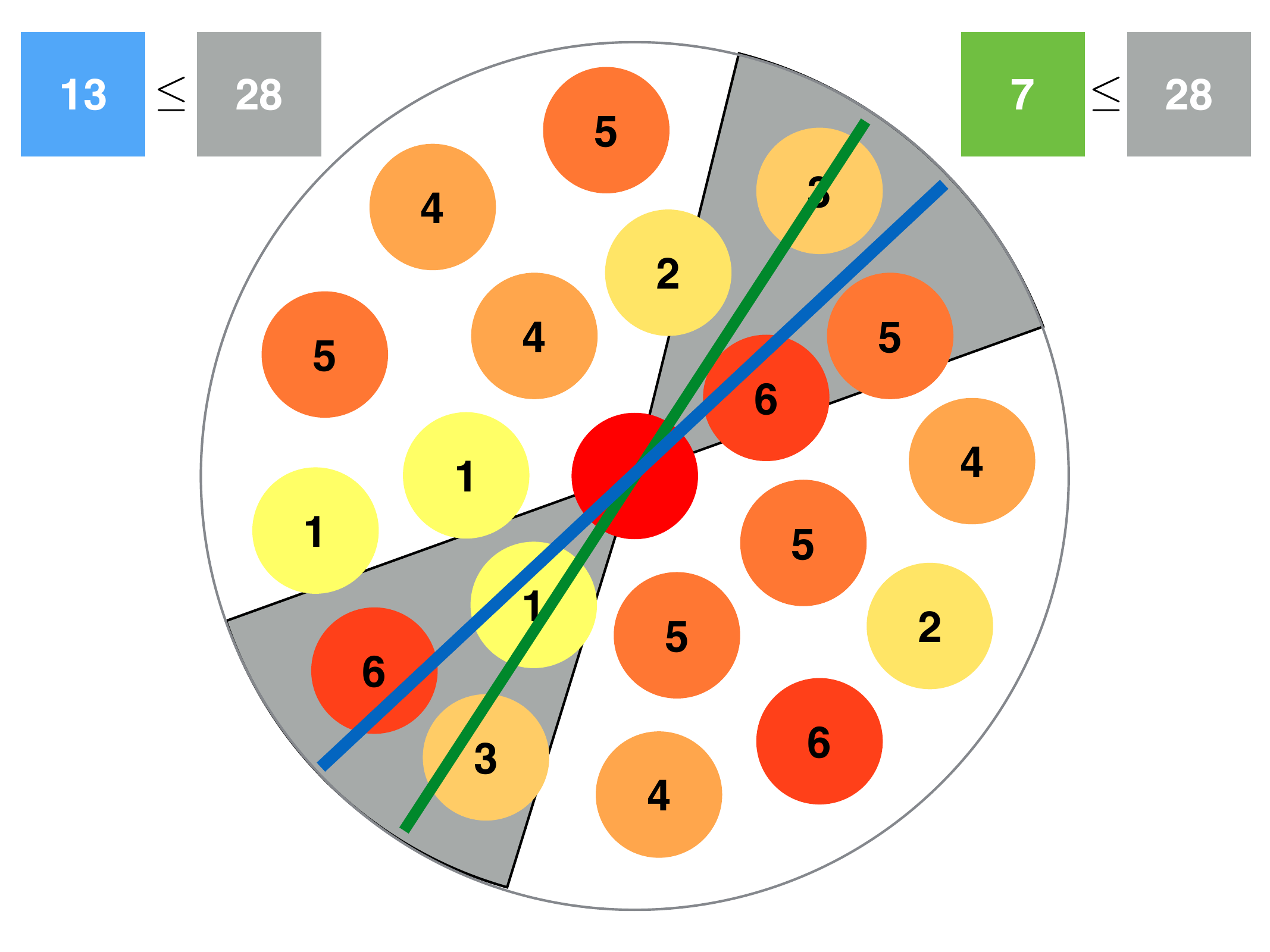} &
      \includegraphics[width=0.34\linewidth]{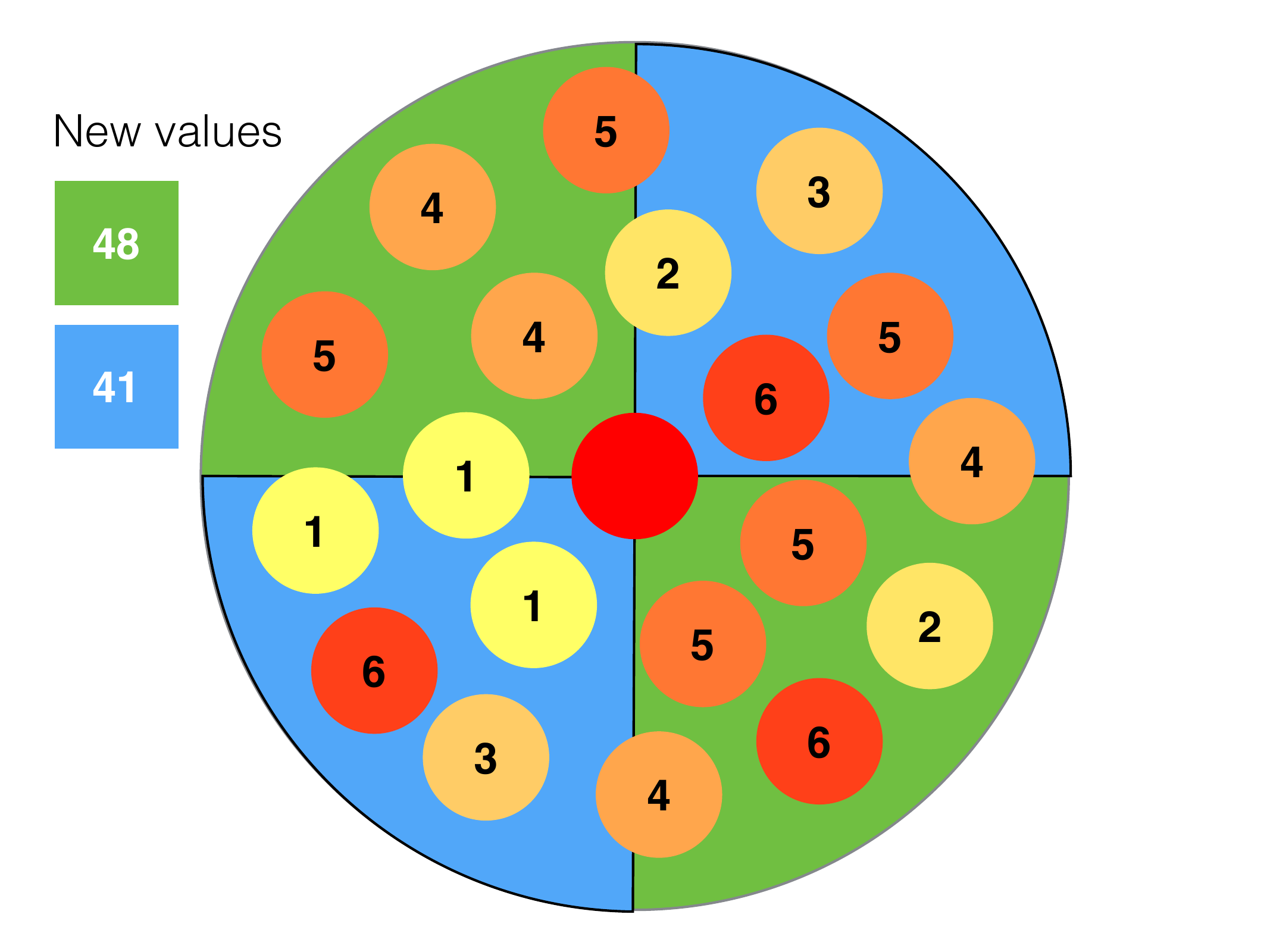} \\
      (a) & (b) & (c) \\
      \includegraphics[width=0.34\linewidth]{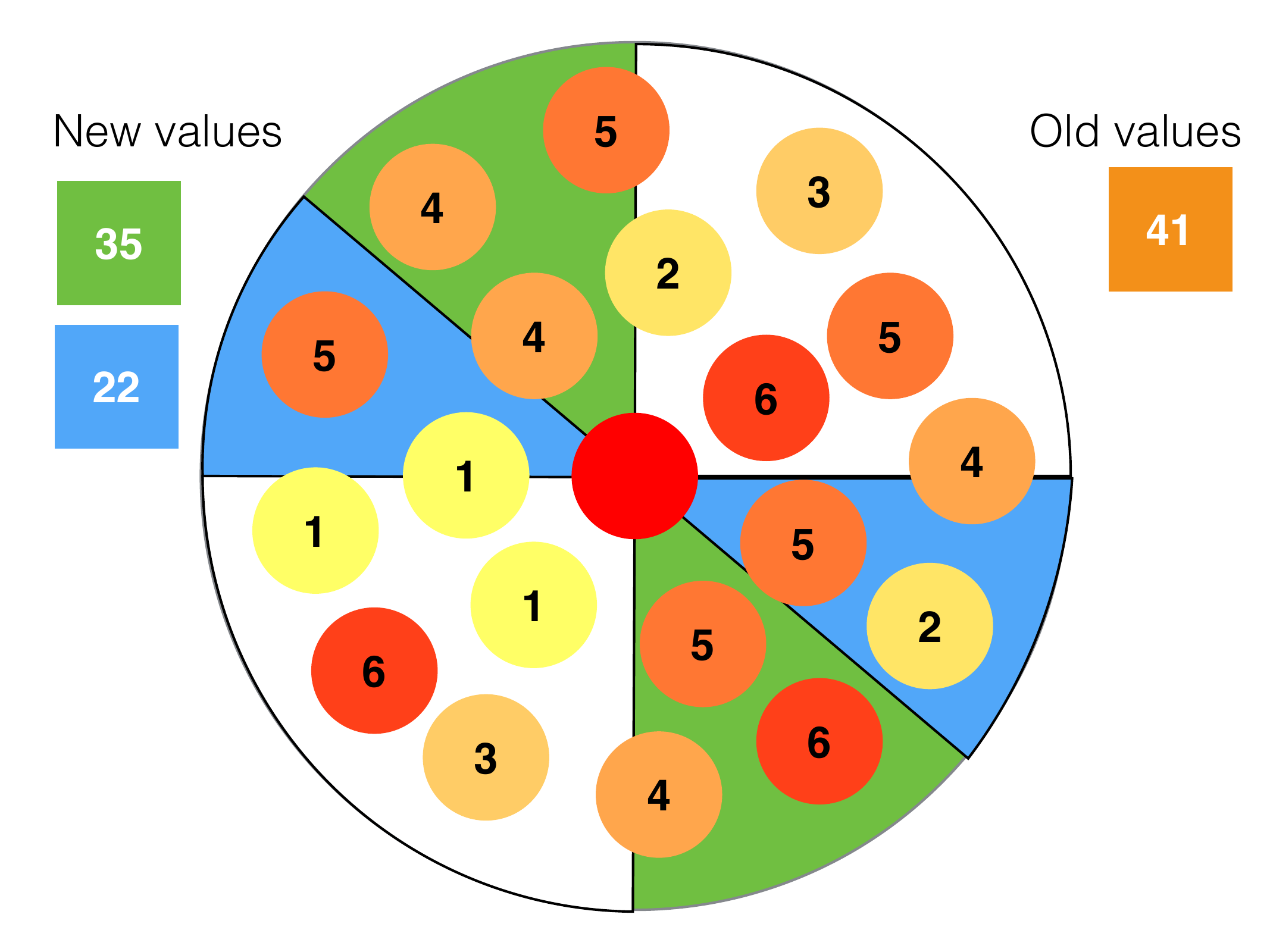} &
      \includegraphics[width=0.34\linewidth]{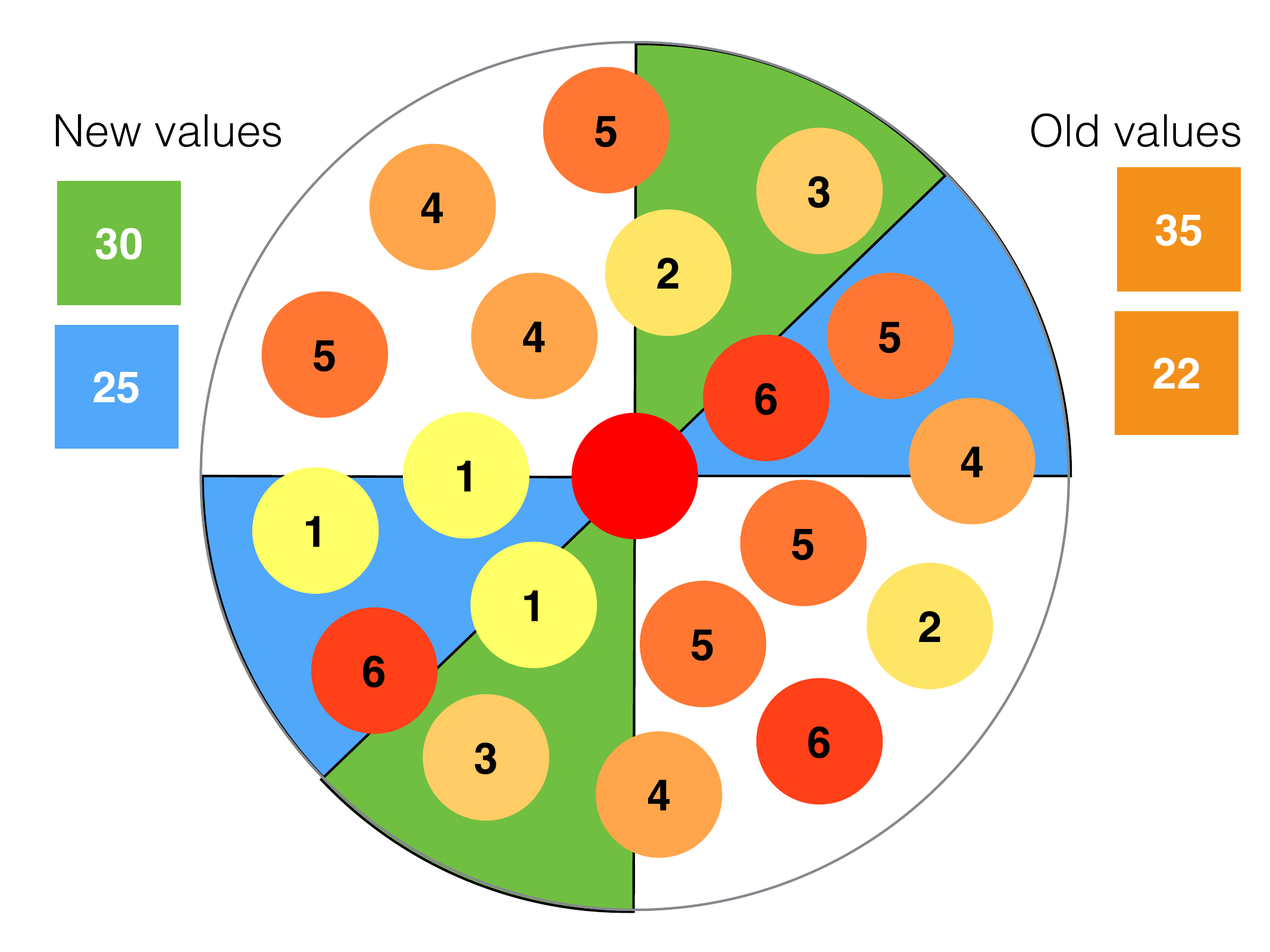} &
      \includegraphics[width=0.34\linewidth]{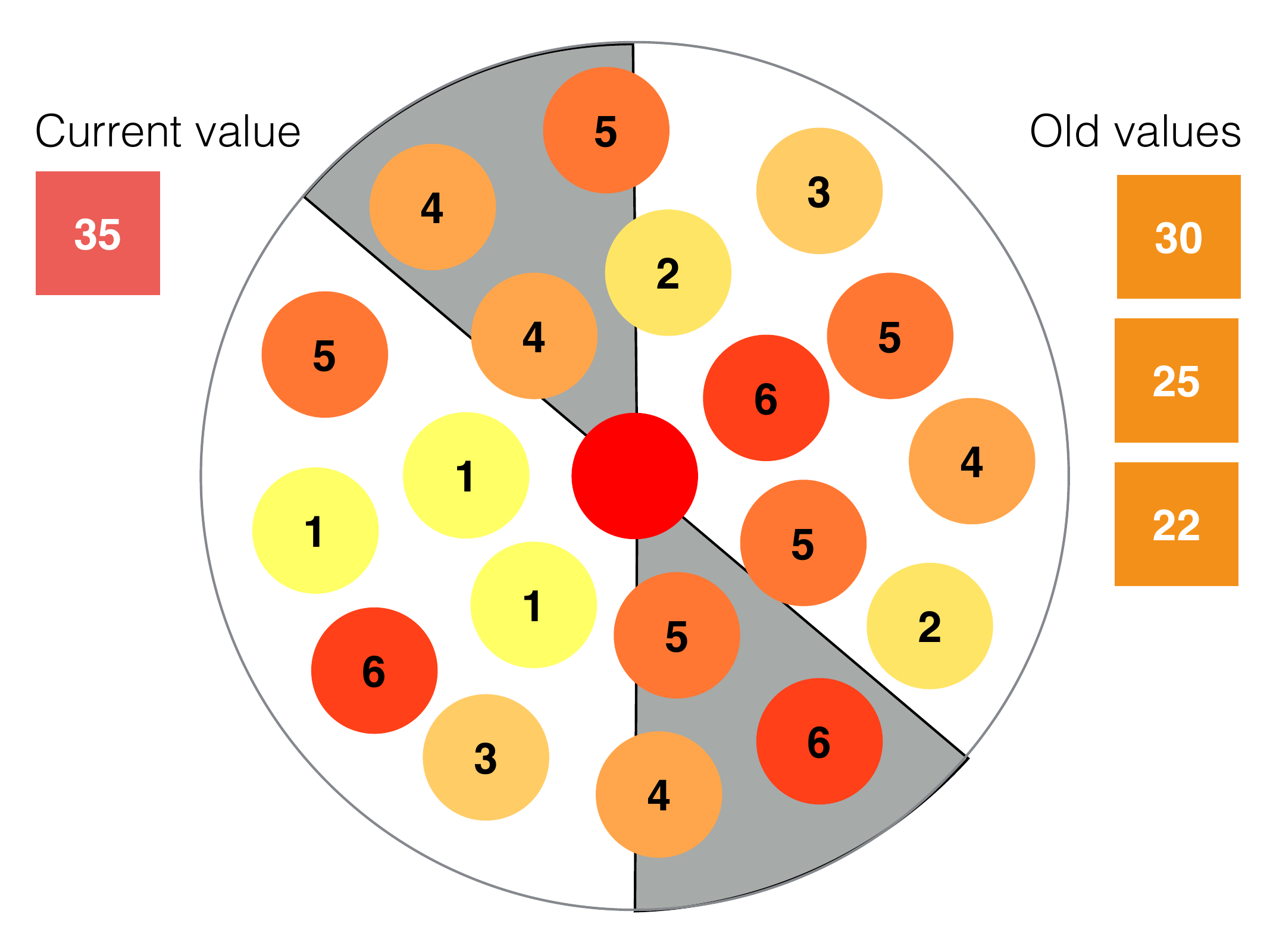} \\
      (d) & (e) & (f)
    \end{tabular}
  \end{center}
  \caption{Illustration of our branch-and-bound algorithm in \num{2}D.  
  a) The \num{2}D-equivalent of searching for a plane in sphere is searching for a line in circle. 
  b) The bounding function states that value of a sector is not smaller than the value of a line inside it.
  c) The procedure starts with splitting the whole parameter interval into \num{2} sectors. We compute the value of each sector and keep a priority queue of them.
  d) At each step of the procedure we divide the sector with the highest uncertainty in two new sectors by a bisector plane.
  e) and f) The procedure continues by splitting the sector with the highest value.}
  \label{fig:05-planes_in_2d}
\end{figure}

As it is difficult to represent graphically our branch-and-bound search procedure in \num{3}D, for illustration purposes we demonstrate it here on a \num{2}D example.
The \num{2}D-equivalent of searching for a plane in sphere is searching for a line in circle.

\begin{figure}[t]
\begin{center}
  \includegraphics[width=0.5\linewidth]{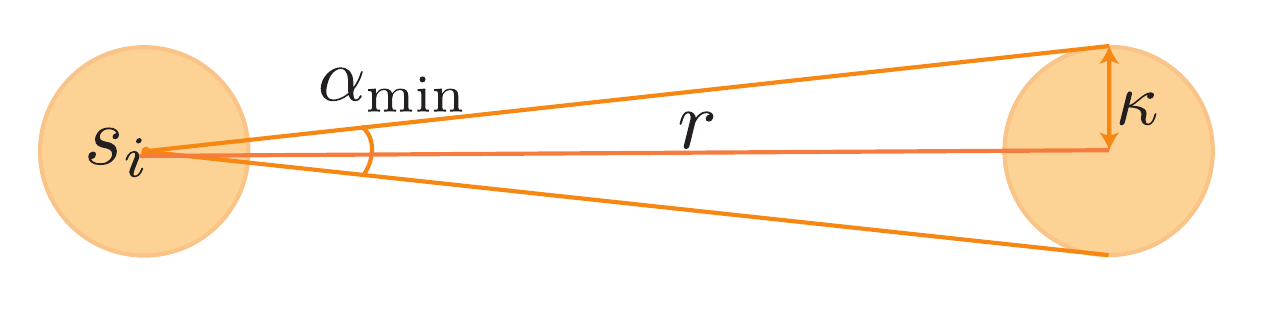}
\end{center}
  \caption{Termination condition of our branch-and-bound procedure in \num{2}D. 
  The search terminates when a sector can fit at most one superpixel at the perimeter of the original circle. 
  In this case, such a sector can be found as the one that exceeds the minimal angle $\alpha_{\min}$.}
   \label{fig:stop_2d}
\end{figure}

In Figure~\ref{fig:05-planes_in_2d}(a) we show a circle with superpixels approximated by circles and where the color of each superpixel indicates how uncertain it is, with red being the most uncertain and yellow the least uncertain.
Then, the task is to find a line of a maximum uncertainty, where the uncertainty of a line is defined as the sum of the uncertainties of the superpixels that it intersects.
For example, Figure~\ref{fig:05-planes_in_2d}(b) demonstrates that the score of blue line is $6+1+6=13$ and the score of the green line is $3+1+3=7$. 
A corridor in \num{3}D corresponds to a sector in \num{2}D.
An example of sector is shown in Figure~\ref{fig:05-planes_in_2d}(b) in grey with its score being $6+3+1+6+2+5+3=26$.
Figure~\ref{fig:05-planes_in_2d}(b) also illustrates the bounding function condition: the score of any line inside the sector is no bigger than the score of the sector that includes this line, in our case, $13 \le 26$ and $7 \le 26$.
The search procedure starts in Figure~\ref{fig:05-planes_in_2d}(c): We split the circle into blue and green sectors and compute their scores (\num{48} and \num{41}). 
All the scores encountered during the search procedure are stored in a priority queue. 
At every iteration, the sector with the highest score is selected from the queue and split into two.
For example, the green sector of Figure~\ref{fig:05-planes_in_2d}(c), whose score of \num{48} is the highest, is split into two new equal sectors resulting in \num{3} sectors depicted in Figure~\ref{fig:05-planes_in_2d}(d). 
The new uncertainty scores \num{35} and \num{22} are added to the priority queue. 
Next, the sector with the highest score is the blue sector of Figure~\ref{fig:05-planes_in_2d}(c). 
We split it into two new sectors in Figure~\ref{fig:05-planes_in_2d}(e). 
Next, we split the green sector of Figure~\ref{fig:05-planes_in_2d}(d) with the uncertainty score of \num{35}.
The procedure continues until the sector with the highest uncertainty can fit at most one superpixel at the perimeter of the original circle as shown in Figure~\ref{fig:stop_2d}. 

\section{Experiments}
\label{sec:exp}

In this  section, we evaluate our approach on  two different Electron
Microscopy (EM)  datasets and on one of Magnetic  Resonance Imaging (MRI) dataset.   We then demonstrate that \combun{} is also effective for natural \num{2}D images.
In multi-class MRI and multi-class natural \num{2}D images of faces the extended version of our approach also results in enhanced performance.

\subsection{General setup}
\label{sec:setup}

For  most of  our  experiments,  we   used  Boosted  Trees  selected   by  gradient
boosting~\cite{Sznitman13a,Becker13b} as our  underlying classifier.
Nothing in our method is specific to a classifier and thus in some of the experiments we also deal with Logistic Regression and Random Forest classifiers.
These general-purpose classifiers were chosen because they can be trained fast, they provide reasonable  probabilistic predictions even with very small amount of training data and many features.
However, there exist no closed form solution when new points are added. 
Thus, AL strategies such as expected model change or expected error reduction are not suitable to be applied with our classifier because they takes hours for one model update for a typical dataset size in our applications.
In Gradient Boosting, given  that  during  early  AL
iterations rounds, only limited amounts of training data are available, we limit the depth of our trees to \num{2} to avoid over-fitting.  Following standard practices,
individual trees are optimized using  \num{40}\%-\num{60}\% of the available training data
chosen at random and  \num{10} to \num{40} features are explored per  split. 

\subsubsection{Baselines}

For  each dataset,  we  compare  our approach  against  several baselines.   The simplest is Random  Sampling (\RS{}), which involves  randomly selecting samples to be  labelled.  It can be  understood as an indicator  of how difficult the learning task is.

In  practice,  the  most   widely  used AL  approach  is Uncertainty Sampling~\cite{Lewis94, Luo13, Elhamifar13, Yang15, Long15a, Sun15a}. 
To test several variations of it, we use several standard definitions of uncertainty described in~\cite{Settles12}. 
The first involves  choosing the sample with the  smallest posterior probability for its predicted class $b_1$, that is,
\begin{equation}
 \argmin_{s_i \in S_U} p_{\theta}(y_i=b_1 | \bx_i).
 \label{eq:minmax}
\end{equation}
Because of the structure of this strategy,  we  will refer  to  it  as Min max: \FMinMax{}.   
Uncertainty can  also be  defined by  considering the  probability difference between  the first and  second most  highly ranked classes  $b_1$ and $b_2$. 
The most uncertain sample is then taken to be
\begin{equation}
 \argmin_{s_i \in S_U} \{ p_{\theta}(y_i=b_1 | \bx_i) - p_{\theta}(y_i=b_2 | \bx_i) \}.
 \label{eq:minmargin}
\end{equation}
We will refer  to this Min margin strategy as \FMinmargin{}. Finally,  the AL procedure can
take into account the entire distribution  of scores over all classes, compute
the Total entropy $H$ of Sec.~\ref{sec:ShannonEntropy}, and select
\begin{equation}
 s^*= \underset{s_i \in S_U}{\arg\max} (H(s_i)),
 \label{eq:entStrategy}
\end{equation}
which we will refer to as \FU{}. 
Recall that \FMinMax{} and \FMinmargin{} cannot be easily combined with the geometric uncertainty because no upper-bound rule is applicable.

In the  case of binary  classification, \FMinMax{}, \FMinmargin{} and  \FU{} are strictly  equivalent   because  the  corresponding  expressions   are  monotonic functions of each other.  
In the  multi-class scenario, however, using one or the other can result in different behaviours, as shown in~\cite{Joshi09,Long15a,Yang15,Jain09,Korner06}.
According to~\cite{Settles12},  \FU{}   is  best   for  minimizing   the  expected logarithmic  loss while  \FMinMax{} and  \FMinmargin{} are  better suited  for minimizing the expected \num{0}/\num{1}-loss.

\subsubsection{Proposed strategies}

All    entropy-based   measures    introduced   in    Secs.~\ref{sec:FeatUncert} and~\ref{sec:GeomUncert} can be used in our  unified framework. 
Let $H^F$ be the specific uncertainty measure that we use in a given experiment.  
The strategy then is to select
\begin{equation}
 s^*= \underset{s_i \in S_U}{\arg\max} (H^F(s_i)).
\end{equation} 
Recall that  we refer to  the feature uncertainty \featun{}  strategy of Sec.~\ref{sec:UncertMeas} that relies on standard Total entropy as \FU{}. 
By analogy,  we will  refer  to those  that  rely on  the  Selection Entropy  and Conditional Entropy  of Eqs.~\ref{eq:SelectEntropy} and~\ref{eq:ConditEntropy} as  \FEntS{}  and \FEntC{},  respectively.   
Similarly,  when using  the combined  uncertainty  \combun{}  of  Sec.~\ref{sec:CombinedUncert},  we  will distinguish between \CU{}, \CEntS{}, and  \CEntC{} depending on whether we use Total Entropy, Selection Entropy, or Conditional Entropy.

Any strategy can be applied in a randomly chosen plane, which we will denote by adding {\bf p} to its name, as in \rPFU.
Finally, we will  refer to  the plane selection strategy of Sec.~\ref{sec:BatchGeom} in  conjunction with  either
\featun{} or  \combun{}  as \PFU{}, \pFEntS{}, \pFEntC{},  \PCU{}, \pCEntS{} and  \pCEntC{}, depending on whether uncertainty from \FU{}, \FEntS{}, \FEntC{}, \CU{}, \CEntS{}, or \CEntC{} is used in the plane optimization.
We will show that it does not require more than three corrections per iteration.
For performance evaluation purposes, we will therefore estimate that each user intervention for \PFU{}, \PCU{}, \pFEntS{}, \pCEntS{}, \pFEntC{}, \pCEntC{} requires either two or three inputs from the user whereas for other strategies it requires only one.

Note that \PFU{} is similar in spirit to the approach~\cite{Top11b} and can therefore  be taken as  a good  indicator of how  it would perform  on our data. However, unlike in~\cite{Top11b}, we do  not require the user to label the whole plane and retain our proposed interface for a fair comparison.

Before diving into the complex experimental setup with real problems that motivate our work, we first conduct a set of experiments where we study a) if the multi-class uncertainty criteria are efficient in general multi-class classification, and b) if the geometry-based uncertainty is applicable with various classifiers. 

\subsection{Multi-class active learning} 

\begin{figure}[h]
\begin{center}
   \includegraphics[width=0.75\linewidth]{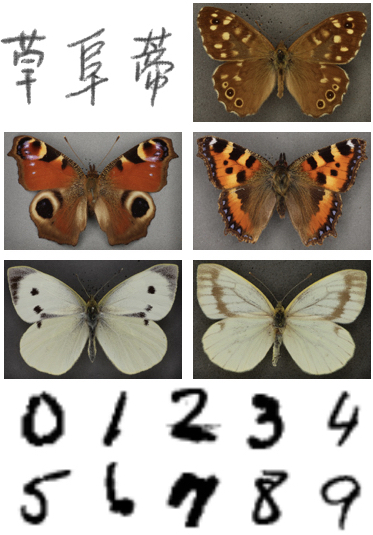}
\end{center}
\vspace{-5mm}
   \caption{ Sample images from the image-classification datasets {\it Chinese}, {\it Butterflies} and {\it Digits}.}
\label{fig:06-classification_datasets}
\end{figure}

\begin{figure*}[]
\begin{center}
\begin{tabular}{ccc}
   \includegraphics[width=0.3\linewidth]{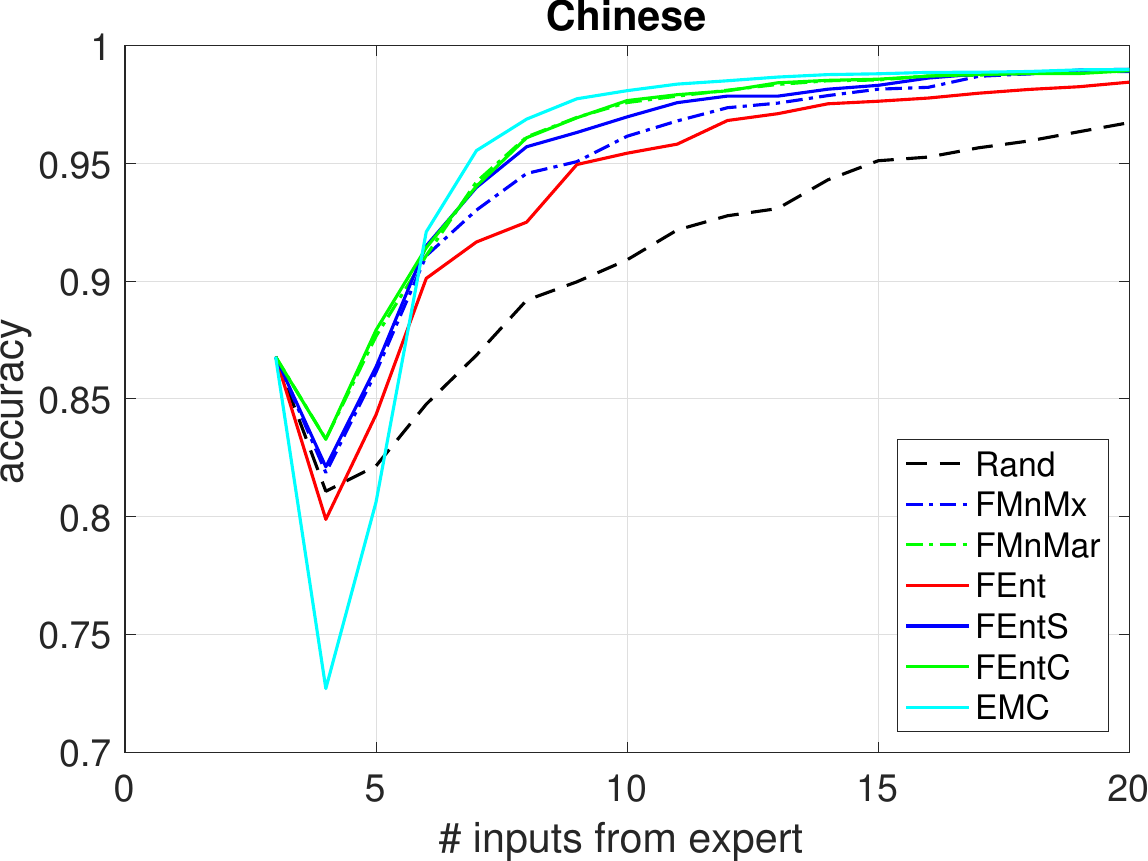} &
   \includegraphics[width=0.3\linewidth]{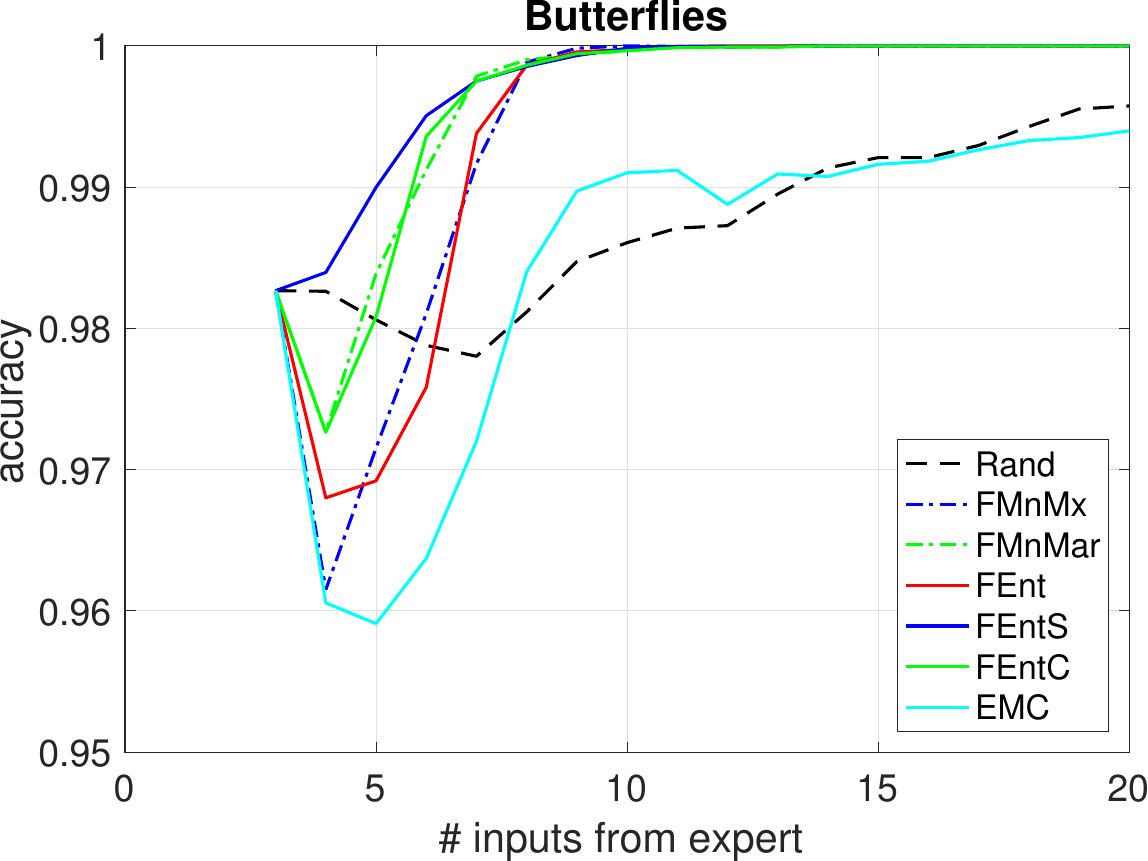} &
   \includegraphics[width=0.3\linewidth]{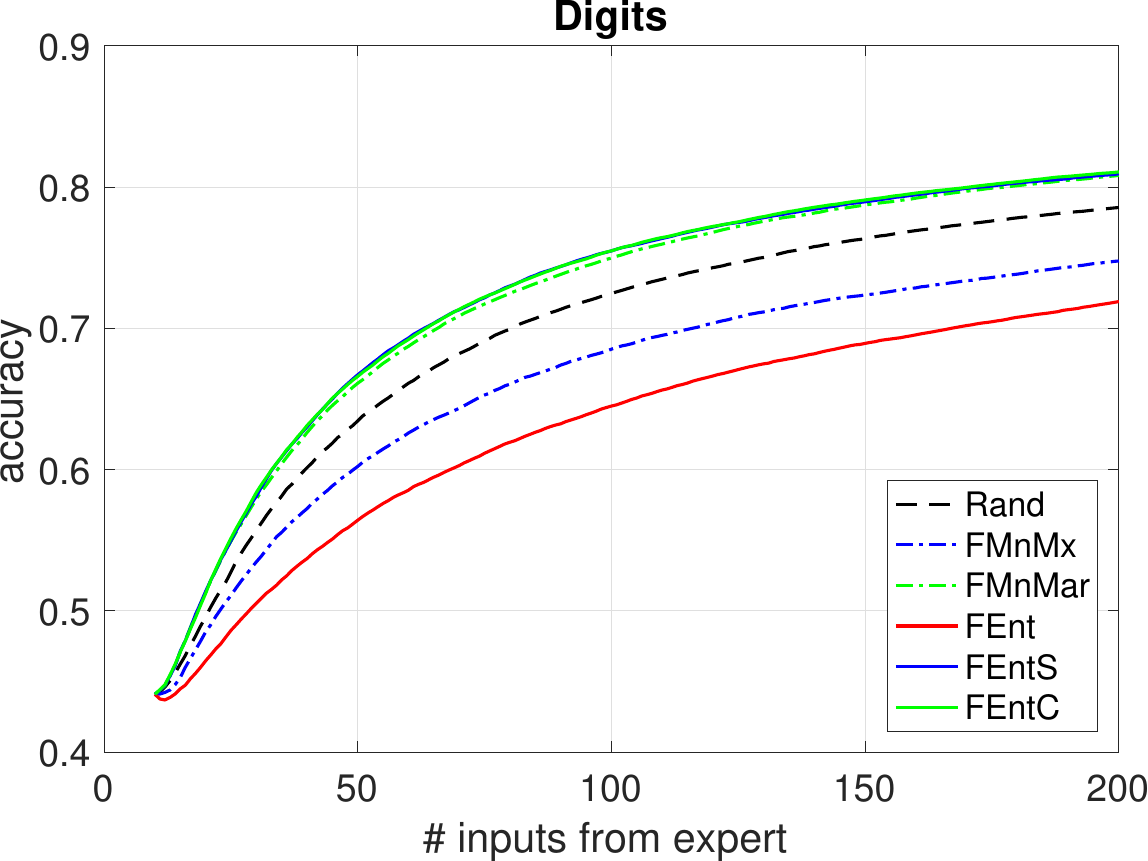} \\
\end{tabular}
\end{center}
\vspace{-4mm}
   \caption{Multi-class AL strategies applied to image classification tasks. Logistic regression is used as an underlying classifier, compare standard multi-class AL criteria to the newly introduced entropy-based criteria.}
   \label{fig:06-multiclassclass}
\end{figure*}

Recall from Sec.~\ref{sec:UncertMeas} that in multi-class scenarios, our different approaches to measuring \featun{} yield different results, as shown in Fig.~\ref{fig:06-multiintuition}. 
Therefore, even though all our strategies derive from the similar intuition, they favour different points. 
For example, \FMinmargin{} selects samples with small margin between the most probable classes irrespectively of the absolute values of the probabilities, whereas \FEntC{} allows for bigger margins for higher values. 
Selection Entropy \FEntS{} tends to avoid samples that look like they can  belong to any of the existing classes. This property can be useful to avoid querying outliers that look unlikely to belong to any class.

To study these differences independently of a full image segmentation pipeline, we first test the various strategies in a simple multi-class image classification task.  
We use them on the three datasets depicted by Fig.~\ref{fig:06-classification_datasets}. {\it Digits} is a standard MNIST collection with \num{10} hand-written digits and we use raw pixel values as features. {\it Chinese} comprises \num{3} classes from the a dataset of of  Chinese handwriting characters~\cite{Liu11}. {\it Butterflies} dataset contains \num{5} classes from British butterfly images from a museum collection~\cite{Johns15}. 
In the {\it Chinese} and {\it Butterflies} datasets, features are extracted using a Deep Net~\cite{Johns15,Jia14}.

We use a logistic regression classifier and test our various AL multi-class strategies including Expected Model Change ({\bf EMC})~\cite{Settles12, Vezhnevets12, Kading15}.
The results are shown in Fig.~\ref{fig:06-multiclassclass}. 
The strategies based on the Selection and Conditional Entropy perform either better or at least as well as the strategies based on the standard measures of uncertainty.
The performance of {\bf EMC} approach is not consistent and does not justify a high computational cost: \num{45} and \num{310} seconds per iteration in Chinese and Butterfly datasets with \num{4096} samples with \num{359} and \num{750} features correspondingly, against \num{0.005} and \num{0.01} seconds by Conditional Entropy.
The {\bf EMC} execution time grows with the AL pool size and thus, we did not run experiments with more than \num{10000} samples.
Many strategies exhibit a noticeable performance drop after a few iterations of AL. 
The reason for this is the imbalance of the class proportions in the training set.
AL starts with a small balanced datasets. 
Then, it is highly likely that the first annotation iterations will make the training set unbalanced. 
With little data, the negative influence of the class imbalance on the classifier is bigger than the advantage of adding more data. 
This effect is quickly eliminated when more data is collected, thus we do not concentrate on it much in the further experiments.

\subsection{Geometry-based uncertainty with different classifiers}

\begin{figure}[h]
\begin{center}
\begin{tabular}{c}
   \includegraphics[width=0.7\linewidth]{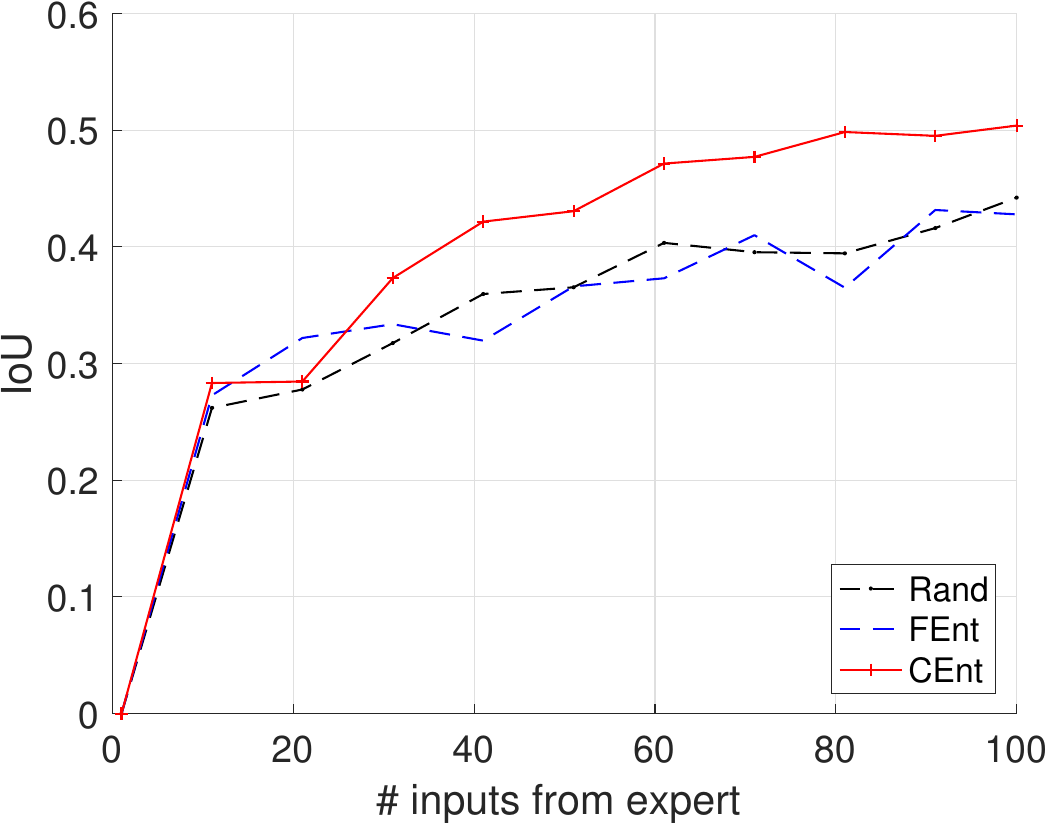}\\
  (a) Random Forest\\
   \includegraphics[width=0.7\linewidth]{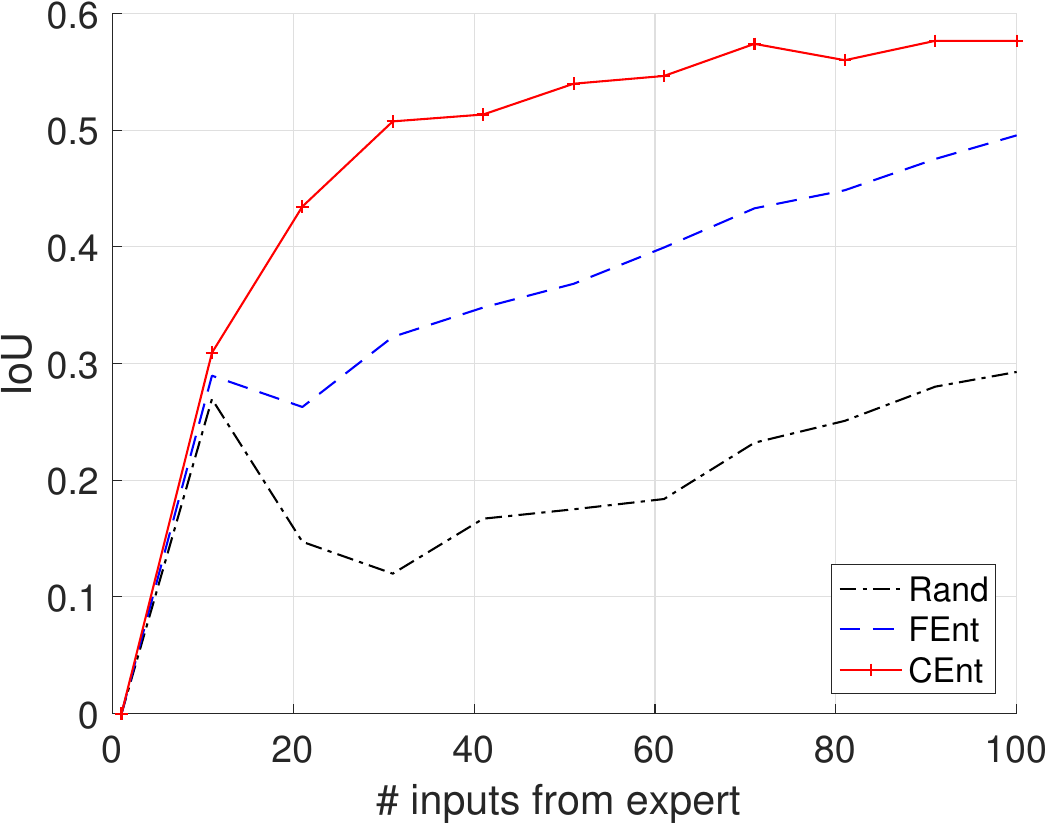}\\
  (b) Gradient Boosting\\ [-0.3cm]
\end{tabular}
\end{center}
   \caption{Comparison of performance of \FU{} and \CU{} with various classifiers. Despite the differences in scores between classifiers, geometric uncertainty consistently wins in performance.}
\label{fig:06-classifiers}
\end{figure}

The way how scores of the classifier are propagated in the Eq.~\eqref{eq:GeomProb} implies nothing about the classifier except for producing a probabilistic prediction. 
However, the applications with image data are characterised by a big number of features in classification and we found various tree-based models to be well suited for these tasks when only little training data is available. 
To check that our Geometric uncertainty can generalise well for various type of classifiers we conduced the experiment where we compare basic versions of \FU{} and \CU{} with two types of classifiers. 
For this experiment we use striatum datasets which will be presented in details further.
We use two types of classification models, Gradient Boosted Decision Trees and Random Forest.
The results are presented in Fig.~\ref{fig:06-classifiers}. 
We notice some variability in the results: while with Gradient Boosting the basic uncertainty sampling outperforms the passive data selection, with Random Forests they perform very similarly. 
Despite these differences, the proposed algorithm \CU{} outperforms \FU{} in both cases. 
In our next experiments we concentrate on using Gradient Boosting classifier as it demonstrated higher absolute scores in this initial study and it is reported to be successful in our applications~\cite{Becker13b}.

\subsection{Parameters of the strategies}
\subsubsection{Adaptive thresholding for binary AL}

\begin{figure}[h]
\begin{center}
\begin{tabular}{c}
   \includegraphics[width=0.6\linewidth]{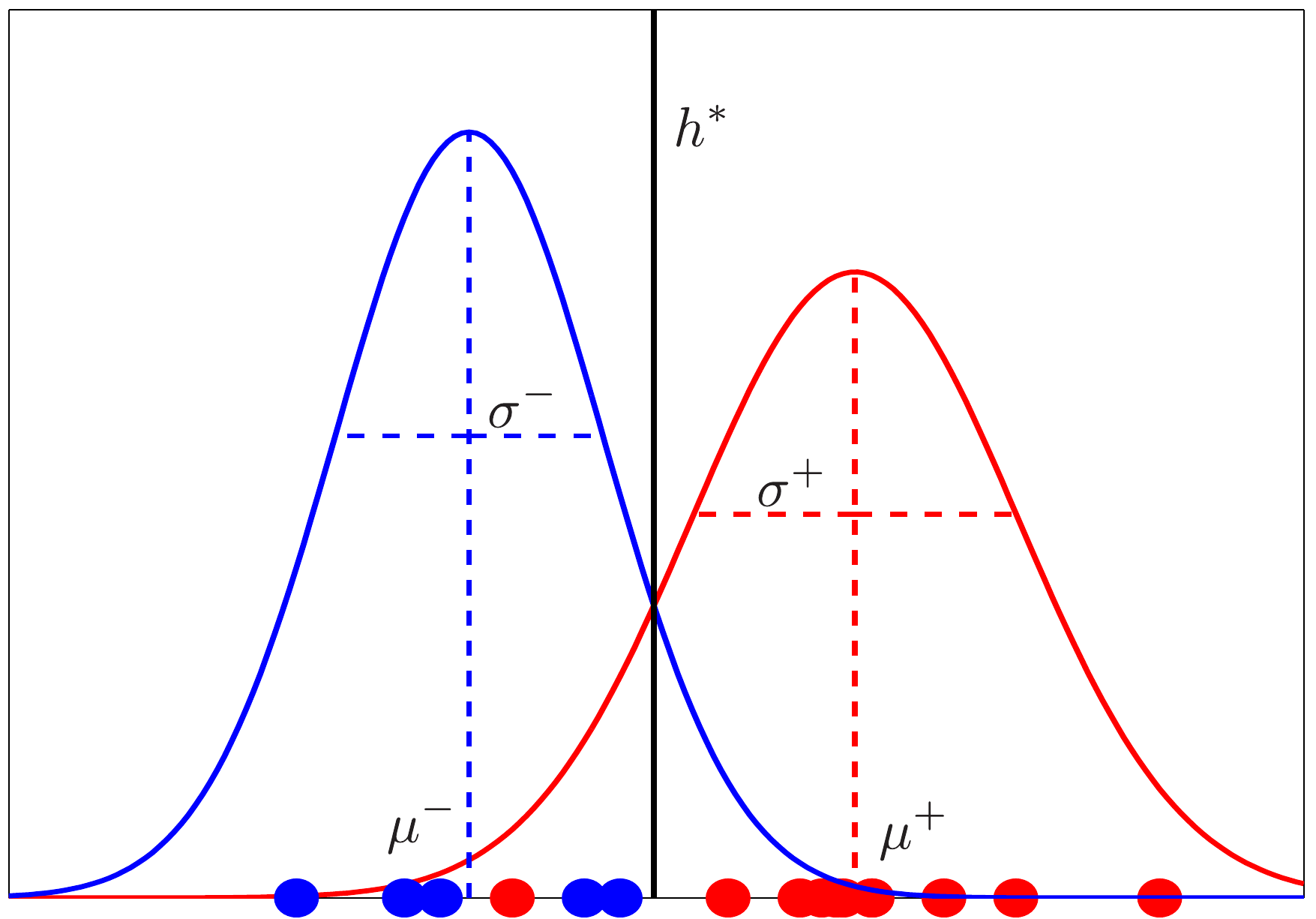}\\
  (a) \\
   \includegraphics[width=0.8\linewidth]{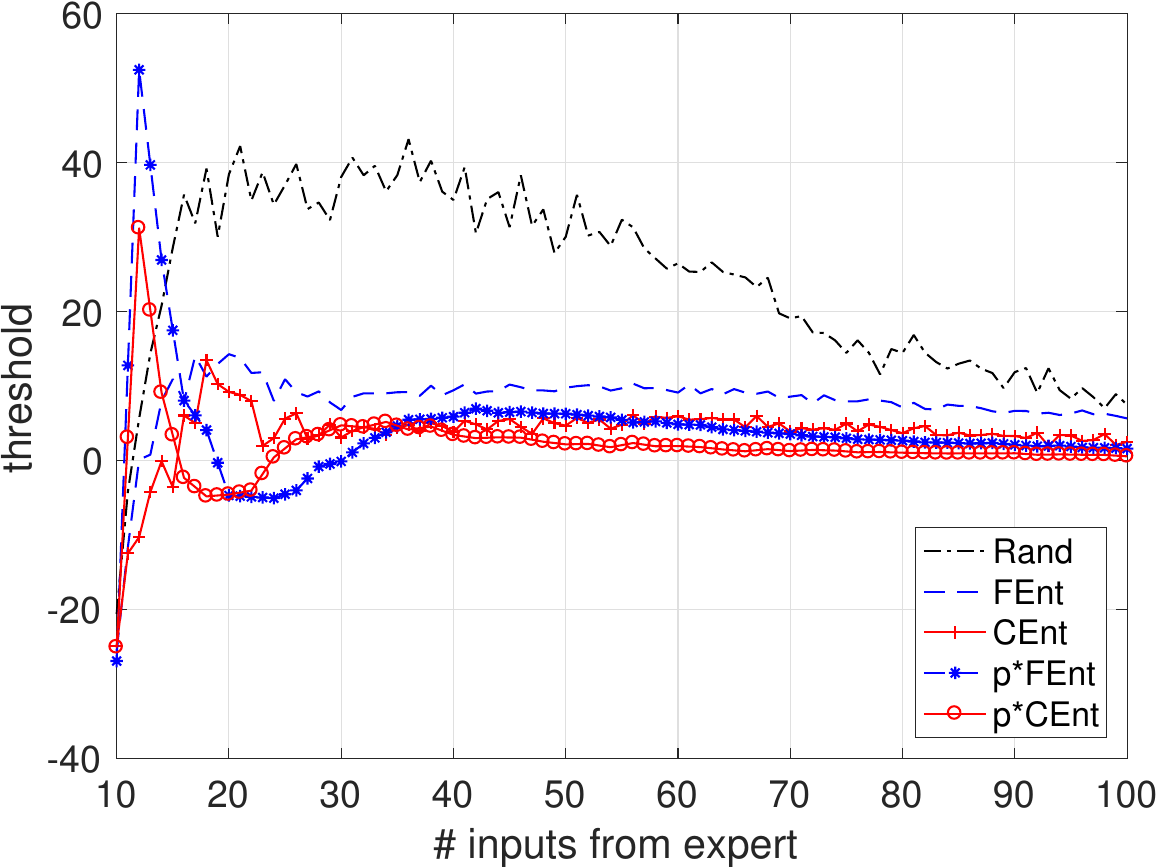}\\
  (b)\\ [-0.3cm]
\end{tabular}
\end{center}
   \caption{Threshold selection. (a) We estimate mean and standard deviation for classifier scores of positive class datapoints ($\mu^+$ and $\sigma^+$, data is shown in red) and negative class datapoints ($\mu^-$, $\sigma^-$, data is shown in blue) and fit 2 Gaussian distributions. Given their pdf, we estimate the optimal Bayesian error with threshold $h^*$.
   (b) Adaptive Thresholding convergence rate of classifier threshold for different AL strategies.}
\label{fig:06-threshold}
\end{figure}

The   probability  of   a  supervoxel   belonging  to   a  certain   class  from Sec.~\ref{sec:FeatUncert} is computed as
\begin{equation}
p_{\theta}(y_i=\hat{y} | x_i)= \frac{\exp^{-2 \cdot  (F_{\hat{y}}-h_{\hat{y}})}}{\sum_{y_j \in Y}\exp^{-2 \cdot  (F_{y_j}-h_{y_j})})},
\end{equation}
where  $F=\{F_{\hat{y}}  |  \hat{y}  \in  Y\}$  is  the  classifier  output  and $h=\{h_{\hat{y}}  | \hat{y}  \in Y\}$  is the  threshold~\cite{Hastie01}.  
Given enough  training data,  it can  be chosen  by cross-validation  but this  may be misleading  or  even impossible at early stages of AL.  
In  practice,  we observe  that the  optimal threshold  value varies  significantly for  binary  classification tasks and that the uncertainty measures are sensitive to it.
By contrast, in multi-class scenarios, the  threshold values remain close to \num{0} and the uncertainty-based strategies  are  comparatively   unaffected.   
In  our  experiments, we  therefore take the threshold to be \num{0} for multi-class  segmentation and  compute it  as follows  in the  binary case.   
We assume  that the  scores of training samples in each class  are Gaussian distributed with unknown parameters
$\mu$ and $\sigma$.  
We then find an optimal threshold $h^*$ by fitting Gaussian distributions to  the scores of positive  and negative classes and  choosing the value   that   yields   the   smallest    Bayesian   error,   as   depicted   by Fig.~\ref{fig:06-threshold}a.   
We refer to this approach as {\it Adaptive Thresholding} and     we use it in all our experiments.
Fig.~\ref{fig:06-threshold}b depicts  the value  of the selected  threshold as the amount of annotated data increases.   
Note that our various strategies yield different  convergence  rates, with the fastest  for  the  plane-based strategies, \PFU{} and \PCU{}.

\subsubsection{Geometric uncertainty}
The average radius of supervoxels $\kappa$ is \num{4.3} in EM dataset and \num{5.7} in MRI dataset. 
Parameter $\kappa$ is computed after setting the total number of supervoxels in a volume that is chosen according to the computational budget. 
In general, higher number of supervoxels can allow for better segmentation accuracy but it is computationally expensive as each of the supervoxels is treated as an additional datapoint for classification.
Besides, it is difficult for a human expert to annotate very small image patches reliably.
We set the number $k$ of nearest neighbours of Sec.~\ref{sec:GeomUncert} to be the average number
of  immediately adjacent  supervoxels  on average,  which is  between  \num{7} and \num{15} depending on  the resolution  of the  image and  size of  supervoxels.  However, experiments showed  that the algorithm  is not very  sensitive to the  choice of this parameter.
For  random walk  inference, $\tau_{max}=10$  iterations yield  the best learning  rates   in  the   multi-class  case   and  $\tau_{max}=20$   in  the binary-segmentation one.

\subsubsection{Plane selection}
\begin{figure}[h]
\begin{center}
\includegraphics[width=0.45\linewidth]{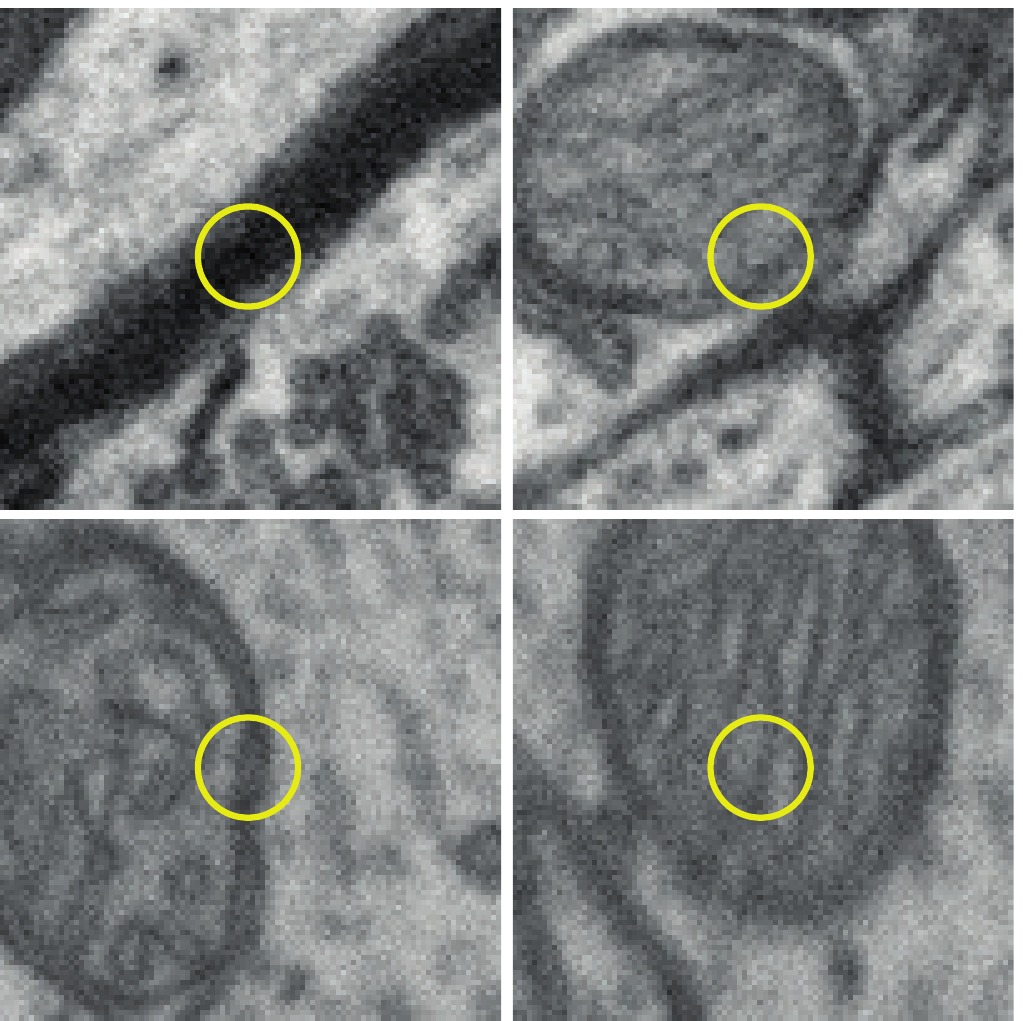} \hspace{5mm}
\includegraphics[width=0.45\linewidth]{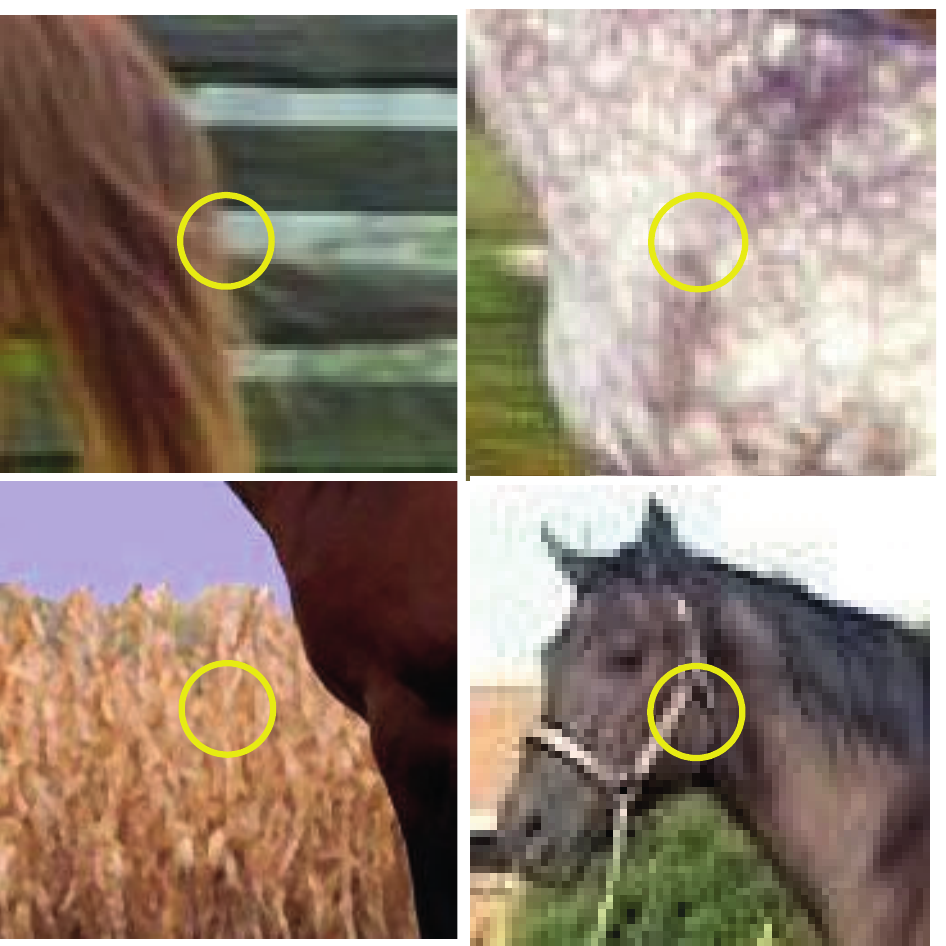}
\end{center}
   \caption{Circular patches  to be annotated  by the expert highlighted  by the yellow  circle in Electron Microscopy and natural images.  
The patches can be annotated either with a line that separates 2 classes or by correcting the mistakes in the current prediction, as shown in Fig.~\ref{fig:01-planeInterface}.}
\label{fig:06-selection}
\end{figure}

We restrict the size of each planar patch to be small enough to
contain typically not  more than \num{2} classes of objects and we explain what happens if this condition is not satisfied. 
To this end, we take  the radius $r$ of Sec.~\ref{sec:ParamSeachSpace} to  be between \num{10} and \num{15}.
Figs.~\ref{fig:01-planeInterface},~\ref{fig:06-selection}  jointly depict  what a potential user would  see for plane selection strategies given a small enough patch radius. 
Given a well  designed interface, it will typically require no  more than  one or  two mouse  clicks to  provide the  required feedback,  as depicted by  Fig.~\ref{fig:01-planeInterface}. 
The easiest way to annotate patches with only two classes is to indicate a line between them, and in situations when more than two classes co-occur in one patch, we allow users to correct mistakes in the current prediction instead.
All  plane selection strategies use the $t=5$ best
supervoxels in the optimization procedure as described in Sec.~\ref{sec:BatchGeom}. Further  increasing this  value does  not yield  any significant learning rate improvement.
The parameters of branch-and-bound plane selection procedure are the radius of the patch $r$ and the number $t$ of patch centers to consider. 
The parameter $r$ is chosen such that it results in a convenient interface for labelling, while
parameter $t$ is chosen depending on a given computational budget. 
Using branch-and-bound algorithm instead of gradient decent as suggested in~\cite{Top11b} avoids setting the the learning rate and other optimization-related parameters.

\subsubsection{Experimental protocol}

We start with \num{5} labelled supervoxels from each class and perform AL  iterations until we receive  \num{100} simulated user inputs in the binary case and \num{200} in the multi-class case.  
Each method starts with  the same random subset  of samples and each  experiment is repeated $N=40-50$  times.  
We  will  therefore  plot not  only  accuracy  results but  also indicate the variance of these results. 
We use half of the available data for independent testing and the AL strategy selects new training datapoints from the other half.

We  have access  to fully  annotated  ground-truth volumes  and we  use them  to
simulate the expert's  intervention in our experiments. This ground truth allows us to model several hypothetical strategies of human expert as will be shown in Sec~\ref{sec:humanExperiments}.
We  detail the specific features we used for EM, MRI, and natural images in the corresponding sections.

\subsection{Results on volumetric data}

\subsubsection{Results on EM data}
\begin{figure}[h]
 \begin{center}
  \hspace{-0.3cm}\includegraphics[width=0.7\linewidth]{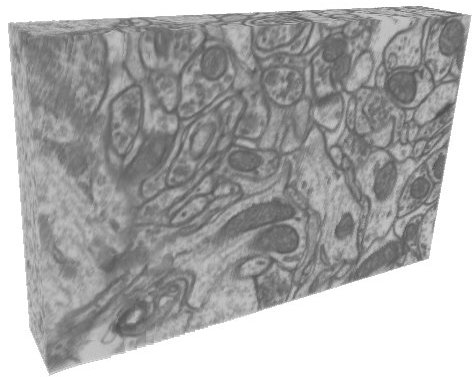}
 \end{center}
\vspace{-0.3cm}
\caption{Hippocampus volume for mitochondria segmentation.}
\label{fig:06-datasets}
\end{figure}

\begin{figure*}[]
\begin{center}
\includegraphics[width=0.4\linewidth]{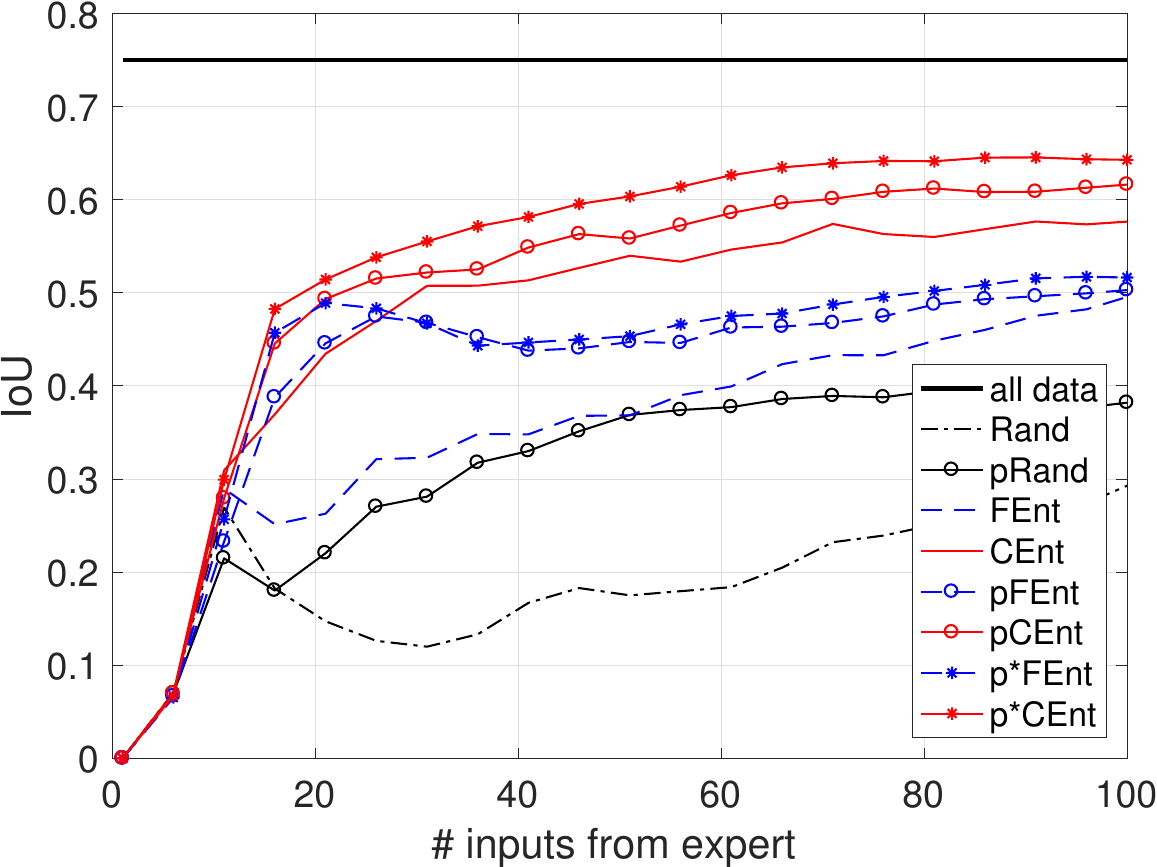} \hspace{10mm}
\includegraphics[width=0.4\linewidth]{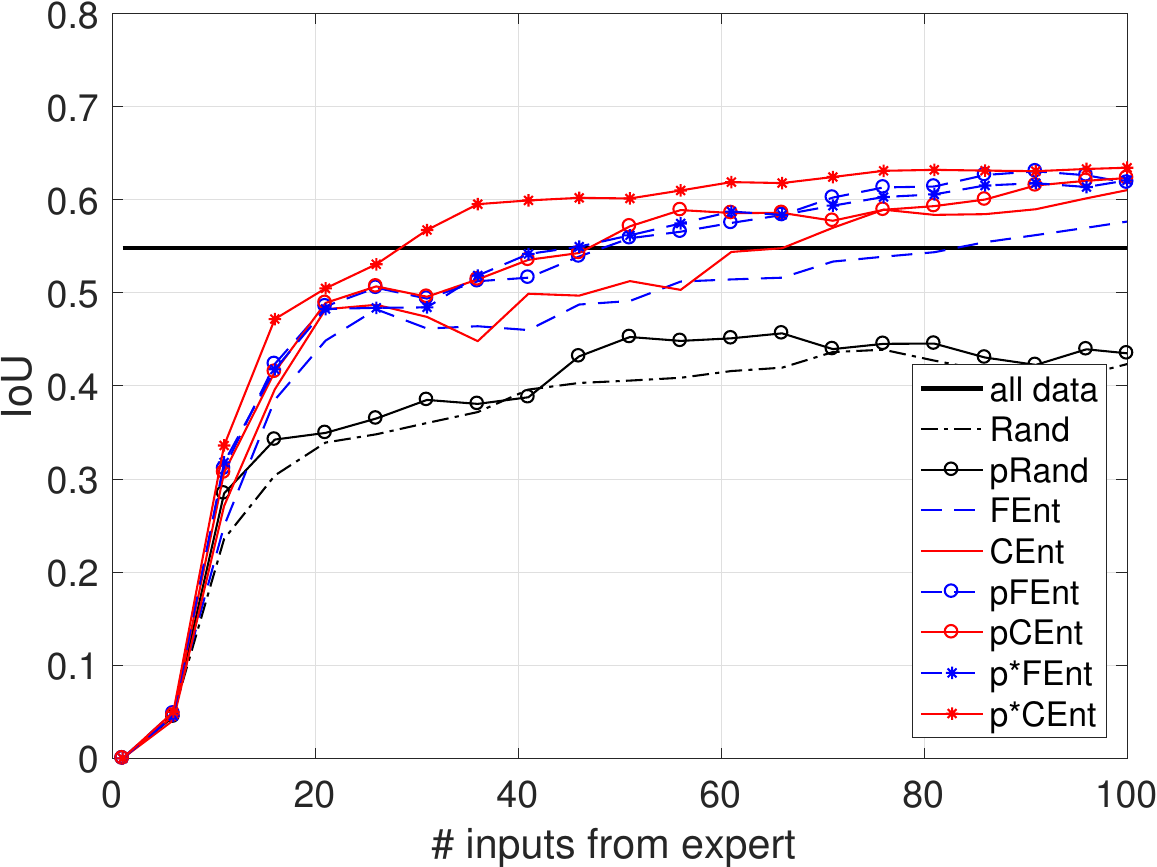}
\end{center}
\vspace{-5mm}
   \caption{Comparison of various AL strategies for (binary) mitochondria segmentation. Left: striatum dataset, right: hippocampus dataset.}
\label{fig:emresults}
\end{figure*}

First, we consider \num{3}D EM stacks of rat neural tissue from the striatum  and from the hippocampus\footnote{http://cvlab.epfl.ch/data/em}.  
One stack  of size \num{318x711x422}  (\num{165x1024x653} for the hippocampus) is  used for training and another stack of size \num{318x711x450} (\num{165x1024x883}) is  used to  evaluate the  performance. 
Their  resolution is  \num{5}nm in  all three spatial orientations.  
The slices  of Fig.~\ref{fig:01-fijiinterface} %as well as patches  in  Fig.~\ref{fig:06-selection}a 
come from  the striatum dataset
and the hippocampus volume is shown in Fig.~\ref{fig:06-datasets}.  
Since the image have the same resolution in all dimensions, they can be viewed equally well in all orientations and specialized tools have been developed for use by neuroscientists~\cite{Pietzsch15}.

The task is to segment mitochondria, which are the intracellular structures that supply the  cell with its energy  and are of great  interest to neuroscientists.
It is extremely laborious to annotate sufficient amounts of training data for learning segmentation algorithms to work satisfactorily.  
Furthermore, different brain areas  have different visual characteristics, which  means that the annotation process  must be
repeated often.   
The features for classification rely  on local texture and  shape information  using ray  descriptors  and intensity  histograms~\cite{Lucchi11b}.

In Fig.~\ref{fig:emresults}, we plot the performance of all the  approaches as a  function of  the annotation effort, where the performance is measured in terms of the intersection over union (IoU) score~\cite{Everingham10}, a  commonly used measure  for segmentation applications.   The horizontal lines at the  top depict the  IoU scores  obtained  by using  the {\it  whole} training set,  which comprises \num{276130}  and \num{325880} supervoxels for  the striatum and the hippocampus, respectively.  
\FU{} provides  a boost over \RS{} and \CU{} yields  a larger  one.  
Any strategy can be combined with a batch-mode AL that means that a \num{2}D plane is selected to be annotated.
For example, strategies \rpRS, \rPFU~and \rPCU~present to the user a randomly selected \num{2}D plane around the sample selected by \RS, \FU~and \CU.
Addition of a plane boosts the performance of all corresponding strategies, but further improvement  is obtained  by selecting an optimal plane by branch-and-bound algorithm in strategies \PFU{}  and \PCU{}.
The final strategy \PCU{} outperforms all the rest of the strategies thanks to the synergy of geometry-inspired uncertainty criteria and the intelligent selection of a batch.
Also note that the \num{100}  samples we  use  are two orders of magnitude smaller  than the  total  number of  available samples. Nevertheless AL provides a segmentation of comparable quality.

Somewhat surprisingly, in the hippocampus case, the classifier performance given
only 100  training data points  is {\it higher} that  the one obtained  by using
{\it all} the training  data. In fact, this phenomenon has  been reported in the
AL  literature~\cite{Schohn00}  and  suggests  that in some cases  a  well  chosen  subset  of datapoints can produce better generalisation performance than the complete set.

Recall  that the performance scores are  averaged over many  runs. 
In  Table~\ref{tab:06-variance},  we  give   the  corresponding variances. 
Note that  both using the geometric uncertainty and the selection  of optimal plane tend to reduce  variance, thus making the process more  predictable. 

\begin{table}
\caption{Variability of results (in the metric corresponding to the task) by different binary AL strategies. 80\% of the scores are lying within the indicated interval. \featun{} is  more variable that \combun{}, batch selection is less variable that single-instance selection and the batch-selection with an optimal plane cut combined with geometry-inspired uncertainty is the least variable. The best result is highlighted in bold.}
\hspace{-5mm}
\begin{center}
  \begin{tabular}{l@{\hskip 1mm} | l@{\hskip 2mm} l@{\hskip 2mm} l@{\hskip 2mm} l@{\hskip 2mm} l@{\hskip 2mm} l@{\hskip 0mm}}
    \toprule
    Dataset & \FU & \CU & \rPFU & \rPCU & \PFU & \PCU \\ \hline
    Striatum & $0.133$ & $0.105$ & $0.121$ & $0.094$ & $0.115$ & {\pmb {0.086}} \\ 
    Hippoc. & $0.117$ & $0.101$ & $0.081$ & $0.092$ & $0.090$ & {\pmb {0.078}} \\ 
    MRI & $0.076$ & $0.064$ & $0.078$ & $0.074$ & $0.073$ & {\pmb {0.048}} \\  
    Natural & $0.145$ & $0.140$ & $0.149$ & {\pmb {0.124}} & \ \ \ --- & \ \ \ --- \\ 
    \bottomrule
  \end{tabular}
  \end{center}
  \label{tab:06-variance} 
\end{table}

\subsubsection{Results on MRI data}
\label{sec:MRI}

\begin{figure}[h]
 \begin{center}
  \hspace{-0.3cm}\includegraphics[width=0.9\linewidth]{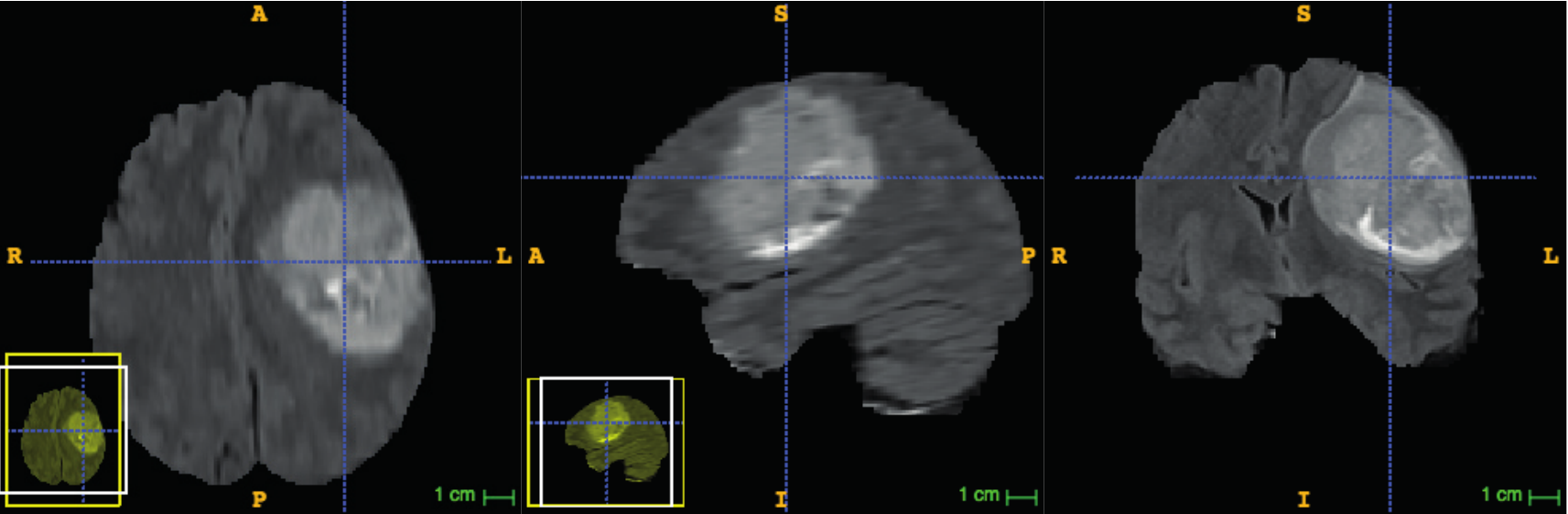}\\[-0.1cm]
 \end{center}
\vspace{-0.3cm}
\caption{MRI data for tumor segmentation (Flair image).}
\label{fig:06-datasets1}
\end{figure}

In this section,  we  consider multi-modal  brain-tumor  segmentation  in MRI  brain  scans an example of which is depicted in Fig.~\ref{fig:06-datasets1}.
Segmentation quality critically depends on the  amount of training data and only highly-trained experts can provide it. 
We use the  BRATS  dataset where T\num{1},  T\num{2}, FLAIR, and post-Gadolinium T\num{1} MR images  are  available  for  each  of \num{20}  subjects~\cite{Menze14}.   
We use  standard filters  such as  Gaussian, gradient  filter, tensor, Laplacian of  Gaussian and Hessian with different  parameters to compute the feature vectors we feed to the Boosted Trees.
Notice that we use different features compared to EM images that demonstrates that our method can be applied in different applications.

{\bf{Foreground-background segmentation}~~}
\begin{figure*}[]
\begin{center}
\includegraphics[width=0.4\linewidth]{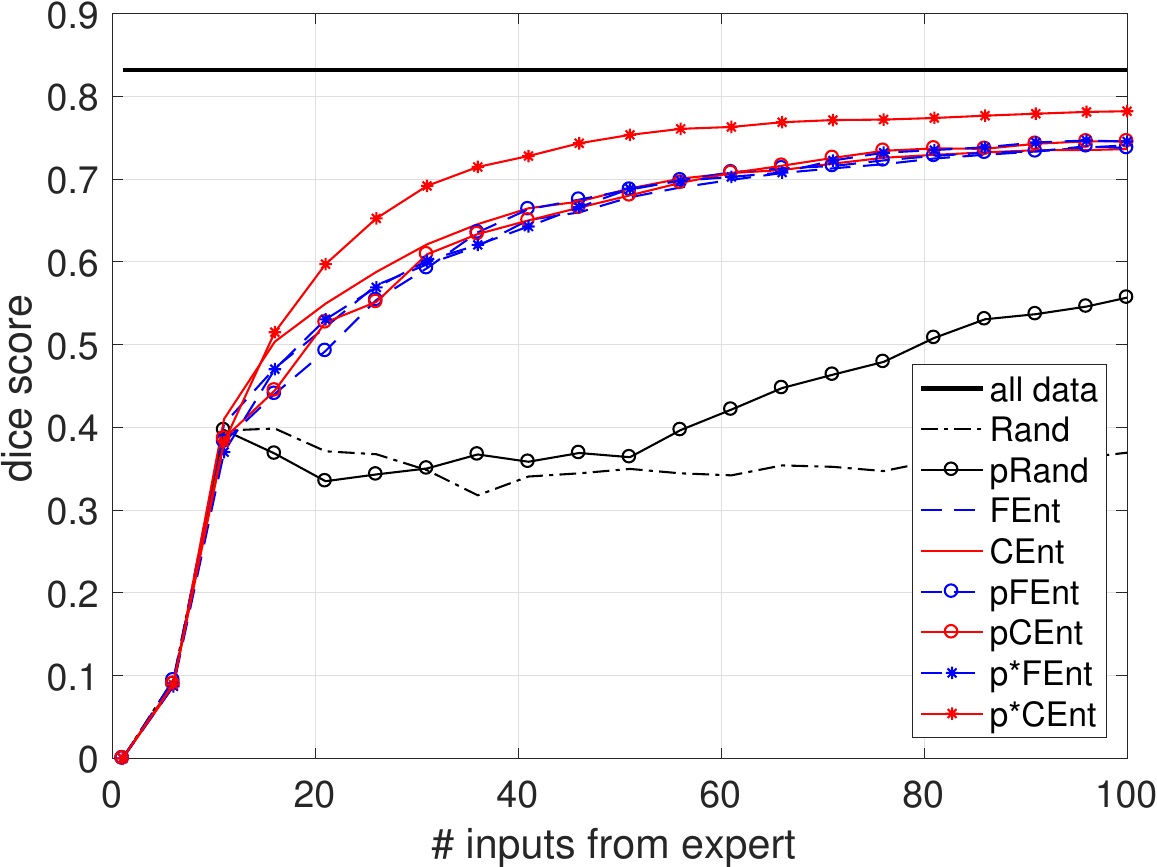} \hspace{10mm}
\includegraphics[width=0.4\linewidth]{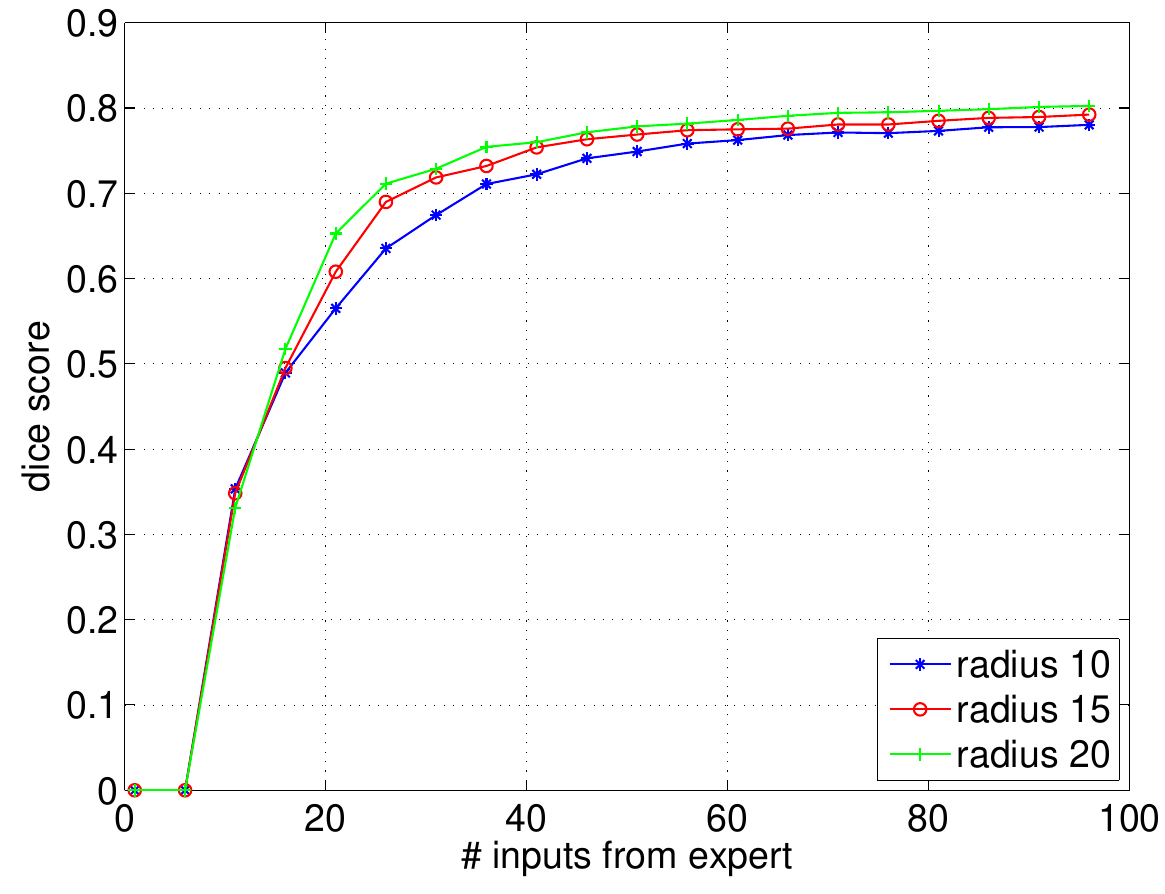}
\vspace{-3mm}
\caption{Comparison of various AL strategies for MRI data for binary tumor segmentation. Right: dice score for BRATS2012 dataset, left: $\PCU$ strategy with patches of different radius. }
\label{fig:mriresults}
\end{center}
\end{figure*}

We first consider segmentation of tumor versus healthy tissue.
In Fig.~\ref{fig:mriresults}, left we  plot the performance of all  the approaches as a function of the annotation effort where performance is measured in terms  of the dice score~\cite{Gordillo13}, a  commonly used quality measure for brain tumor segmentation. 
In Table~\ref{tab:06-variance} we give the corresponding variances. 
We observe the same  pattern as in  Fig.~\ref{fig:emresults}, with \PCU{} again resulting in the highest score.
Note that difference between \PCU{} and \rPCU{} is greater than between \PFU{} and \rPFU{} in all the experiments. 
This is the evidence of the synergy brought by the {\it geometric} uncertainty and the batch selection based on the {\it geometry}.

The patch  radius parameter $r$  of Sec.~\ref{sec:ParamSeachSpace} plays  an important role in  plane selection procedure. 
To  evaluate its  influence, we recomputed  our \PCU{} results \num{50}  times using three different  values for $r=10$, \num{15}  and $20$.  
The resulting plot is shown in  Fig.~\ref{fig:mriresults} on the right. 
As expected, with a  larger radii, the learning curve is slightly higher since more voxels are labelled  each time. However,  as the patches  become larger, it  stops being clear that  annotation can  be done  with small user effort and  that is  why we
limit ourselves to radius sizes of \num{10} to \num{15}.

\begin{figure}[h]
\begin{center}
   \includegraphics[width=0.4\linewidth]{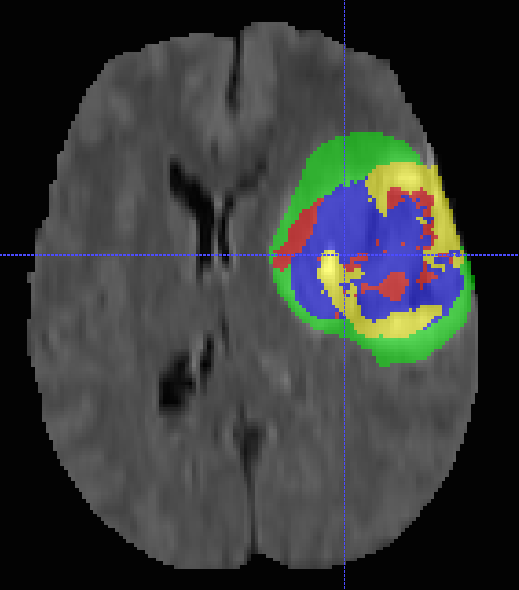}
\end{center}
   \caption{Example of ground truth from multi-class brain-tumor segmentation. Necrotic   center in red, edema in green, non-enhancing  gross   abnormalities in blue  and enhancing tumor  core in yellow. Best seen in color.}
   \label{fig:06-multiMRIgroundtruth}
\end{figure}

{\bf{Multi-class segmentation}~~}
\label{sec:multiMRI}
To test our multi-class approach, we use the full label set of the BRATS competition: healthy tissue (label \num{1}),
necrotic   center (\num{2}),  edema (\num{3}),   non-enhancing  gross   abnormalities (\num{4}),  and
enhancing tumor  core (\num{5}). 
Fig.~\ref{fig:06-multiMRIgroundtruth} shows a ground truth example for one slice in one of the volumes. 
Different classes are indicated in different colors. 
Note that the ground truth is highly unbalanced: we have \num{4000} samples of healthy tissue, \num{1600} of edema, \num{750} of enhancing tumor  core, \num{250} of necrotic   center and \num{200} of non-enhancing  gross   abnormalities in the full training dataset.
We use the evaluation protocol of  the BRATS competition~\cite{Menze14} to  analyse our results.   This involves evaluating how  well we
segment complete  tumors (classes \num{2}, \num{3}, \num{4}, and  \num{5}), core tumors (classes  \num{2}, \num{4}, and
\num{5}), and enhancing tumors (class \num{5} only).

\begin{figure}[!h]
\begin{center}
\includegraphics[width=0.8\linewidth]{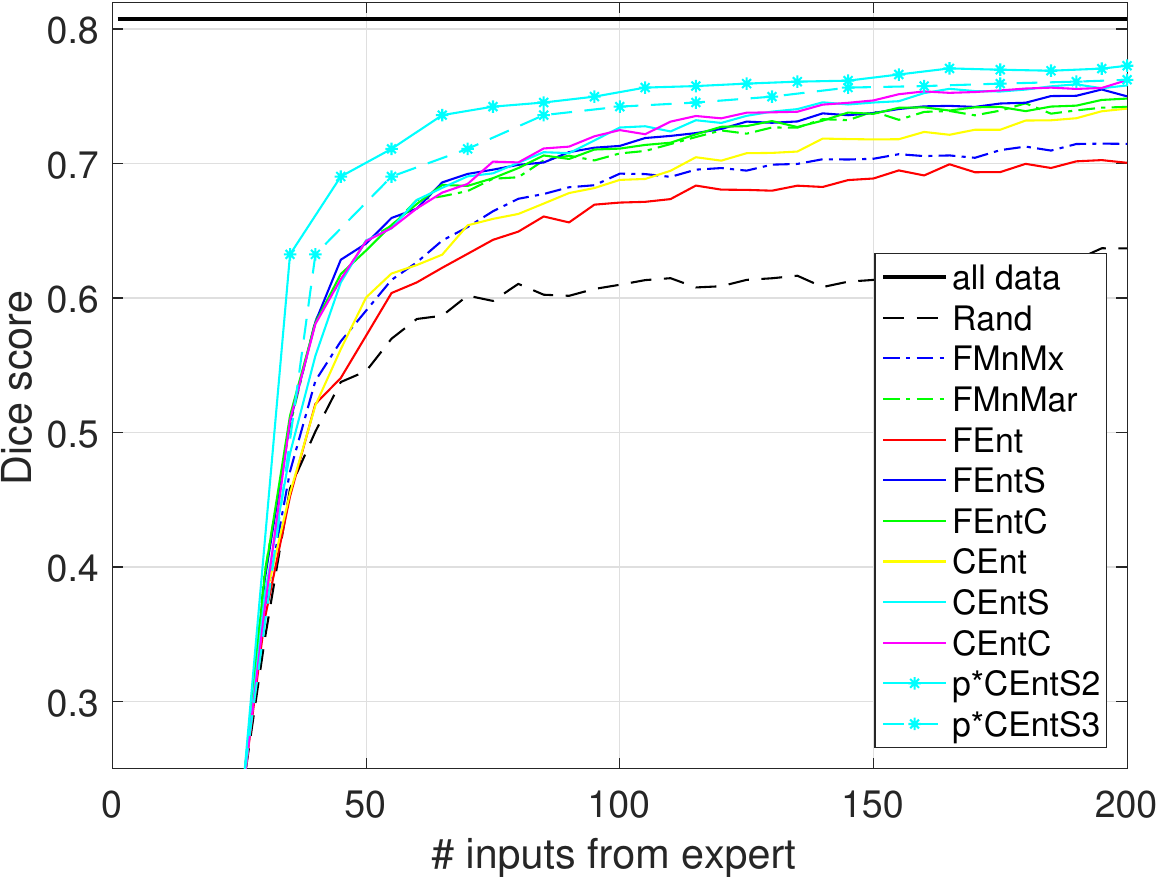}
\includegraphics[width=0.8\linewidth]{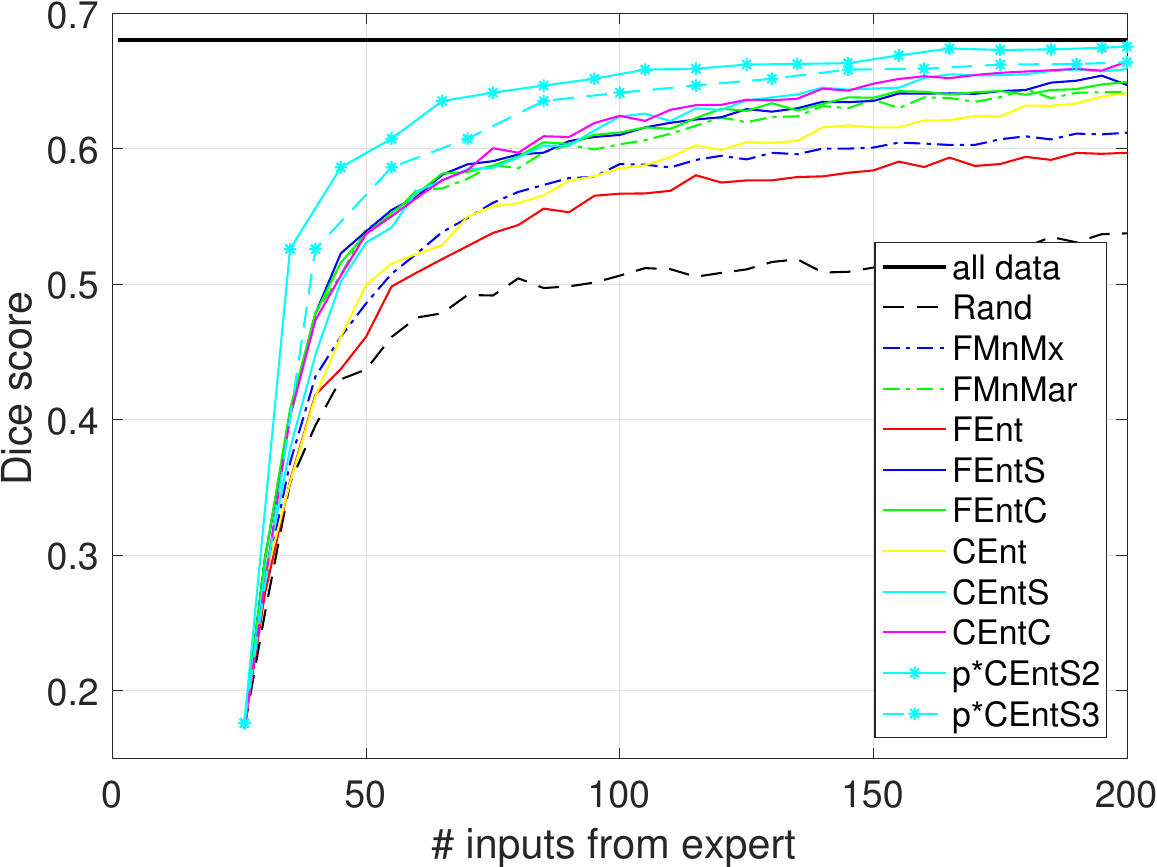}
\includegraphics[width=0.8\linewidth]{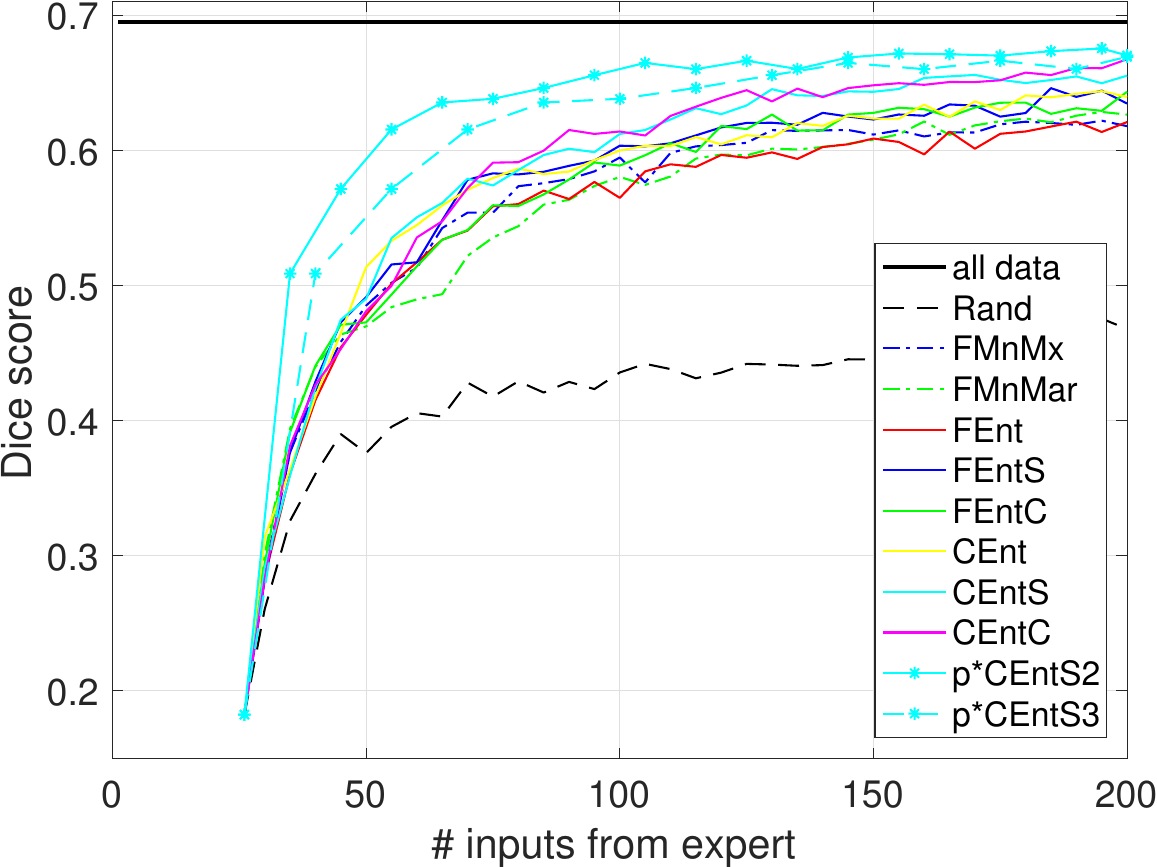}
\end{center}
\vspace{-5mm}
   \caption{Comparison of different AL strategies for multi-class MRI segmentation. Dice scores for three BRATS2012 tasks: complete tumor, tumor core, enhancing tumor.}
\label{fig:06-mrimultiresults}
\end{figure}

Fig.~\ref{fig:06-mrimultiresults} depicts  our results and those  of the selected baselines on these three tasks.  
As before, the results clearly indicate that AL provides a significant improvement over passive selection.  
In this case we do not show all the variants of batch-mode query selection for the benefit of the figure clarity.
Among the basic strategies, \FMinmargin{}  gives the best  performance in  subtasks \num{1} and  \num{2} and \FMinMax{} in subtask \num{3}.  
Our  entropy-based uncertainty strategies \FEntS{}~and  \FEntC{} perform  better  or equivalent to the corresponding  baselines \FMinMax{} and \FMinmargin{} as in the preliminary experiments.  
Next, the strategies with the geometric uncertainty
\CU{}, \CEntS{}  and \CEntC{} outperform their  corresponding \featun{} versions
\FU{}, \FEntS{} and  \FEntC{}, where the improvement depends on  the subtask and
the strategy.  
Note that \FEntS{} and  \FEntC{} as well as \CEntS{} and \CEntC{} perform  equally  well,  thus, they can be used  interchangeably.   
Further improvement is obtained  when each of the strategies is  combined with the optimal plane selection.

In practice, we observed that around \num{43}\% of selected patches contain more than two classes. 
In such cases, simply finding a line separating two classes  is not enough to annotate a patch. 
To handle such cases, we propose a different  annotation scheme.  
The current prediction on supervoxels is displayed to the annotator who needs to correct the mistakes in the prediction. 
We count the number of misclassified samples throughout the experiments and on average there were no more than \num{10.42}\% errors in the supervoxel classes, that is approximately \num{2.42} supervoxels per iteration.  
Thus, we show the learning curves for the plane-based strategy and we count one annotation iteration as either two and or three inputs from the user, with both variants dominating non-batch selection.

\begin{figure}[h]
\begin{center}
   \includegraphics[width=0.8\linewidth]{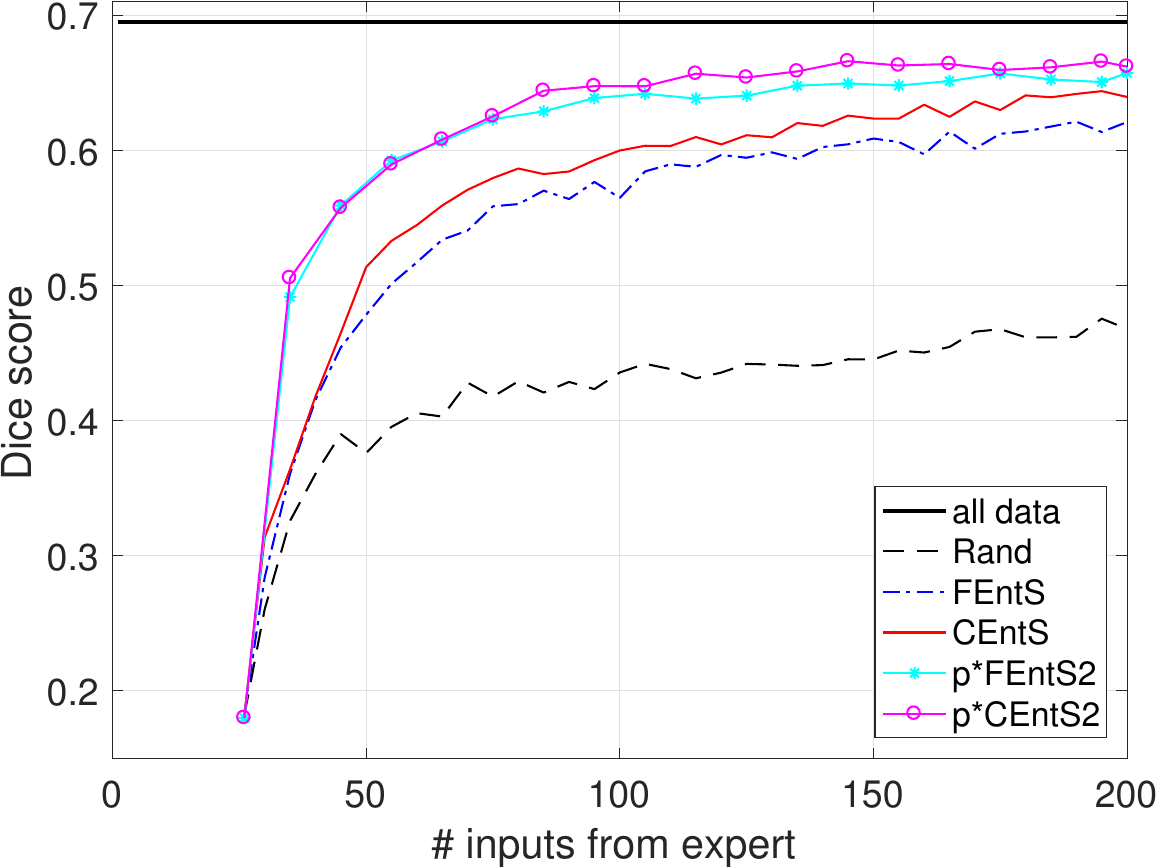}
\end{center}
\vspace{-5mm}
   \caption{Dice score for enhancing tumor segmentation. Performance of various strategies that have Selection Entropy at their basis.}
   \label{fig:06-mri_dice3_geometryplane}
\end{figure}

The  difference  between  competing  \combun{}  strategies  becomes negligible with a slight dominance of  Selection Entropy \pCEntS{} in subtasks \num{1} and \num{2} and Total entropy \PCU{} in the last subtask.  
In seven of nine cases, the \combun{} in  conjunction with  the plane selection  yields better  results than \featun{} with plane selection and in two of nine, they perform equally well.  For  illustrative purposes,  Fig.~\ref{fig:06-mrimultiresults} contains  only the best performing learning curve of \pCEntS{} and Fig.~\ref{fig:06-mri_dice3_geometryplane}   shows   the   performance   of   all strategies   based   on   the   Selection  Entropy   in   the   third   subtask.

\subsection{Results on natural images}
\label{sec:face}

Finally, we  turn to natural \num{2}D  images and replace supervoxels  by superpixels.
In  this case,  the  plane selection  reduces to a simple selection  of patches  in  the  image and we will refer to these strategies as \rPFU{} and  \rPCU{} because they do not involve the branch-and-bound selection procedure.
In  practice, we  simply  select superpixels with their \num{4} neighbours in binary segmentation and \num{7} in multi-class segmentation. 
This parameter is determined by the size of superpixels used in oversegmentation.
Increasing this number would lead to higher learning  rates in  the same  way as  increasing the  patch radius  $r$, but  we restrict it to a small value to ensure labelling of each patch can be done with little user intervention. 

\begin{figure}[h]
\begin{center}
   \includegraphics[width=0.8\linewidth]{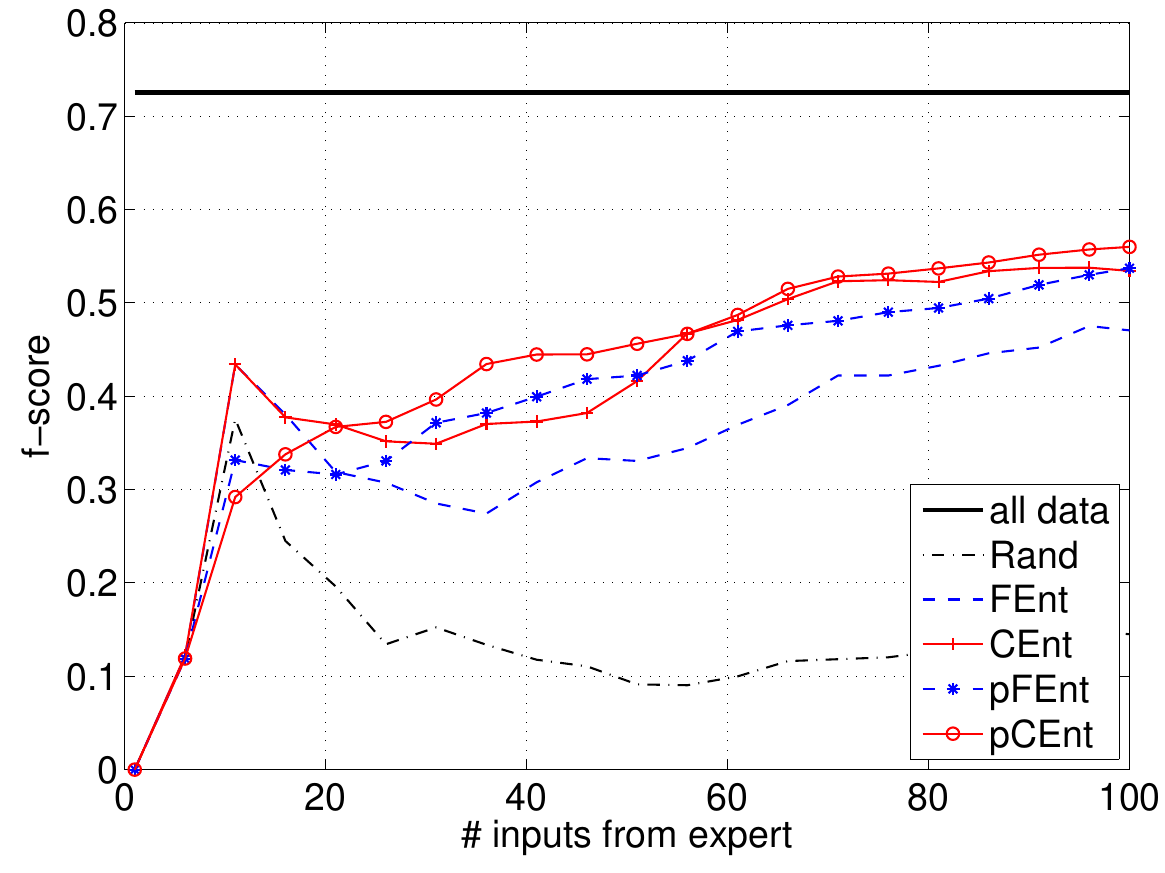}\\
\end{center}
\vspace{-5mm}
   \caption{Comparison of various AL strategies for binary segmentation of natural images.}
   \label{fig:06-naturalresults}
\end{figure}

{\bf{Foreground-background segmentation}~~}
We study the results of binary AL on the Weizmann horse database~\cite{Borenstein08} in Fig.~\ref{fig:06-naturalresults}  and   give  the  corresponding   variances  in Table~\ref{tab:06-variance}. 
To compute image features, we use Gaussian, Laplacian, Laplacian of Gaussian, Prewitt and  Sobel filters for intensity and color  values, gather first-order statistics such  as local standard deviation,  local range, gradient magnitude and direction histograms, as well as SIFT features.
The pattern is again similar to the one observed in Figs.~\ref{fig:emresults} and~\ref{fig:mriresults}, with  the difference between
\CU{} and \rPCU{}  being smaller due to  the fact that \num{2}D  batch-mode approach does not involve any optimization of patch selection.  
Note, however,  that while the  first few iterations result in the reduced scores for all methods, plane-based methods are able to recover from this effect quite fast. 

\begin{figure}[h]
  \begin{center}
    \begin{tabular}{cc}
      \includegraphics[width=0.3\linewidth]{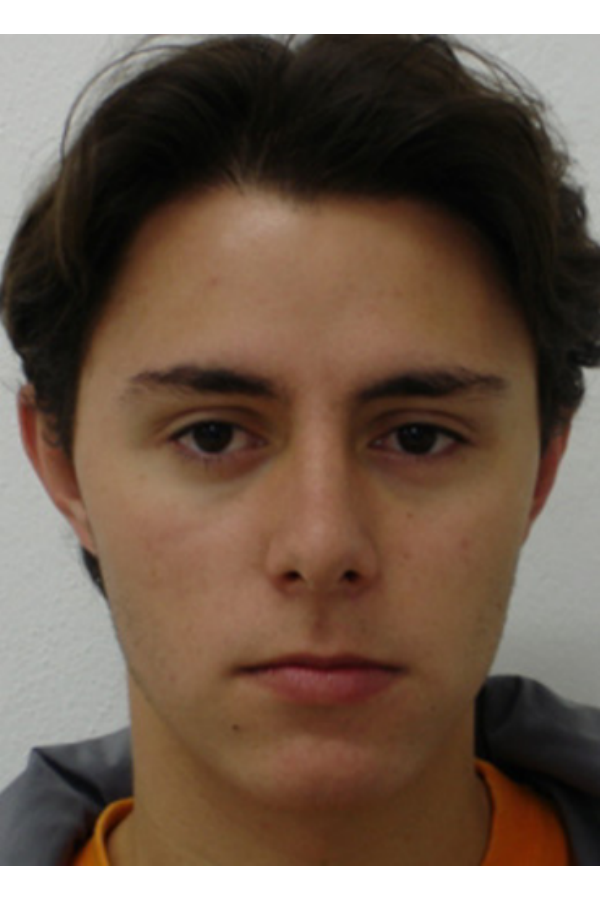}&
      \includegraphics[width=0.3\linewidth]{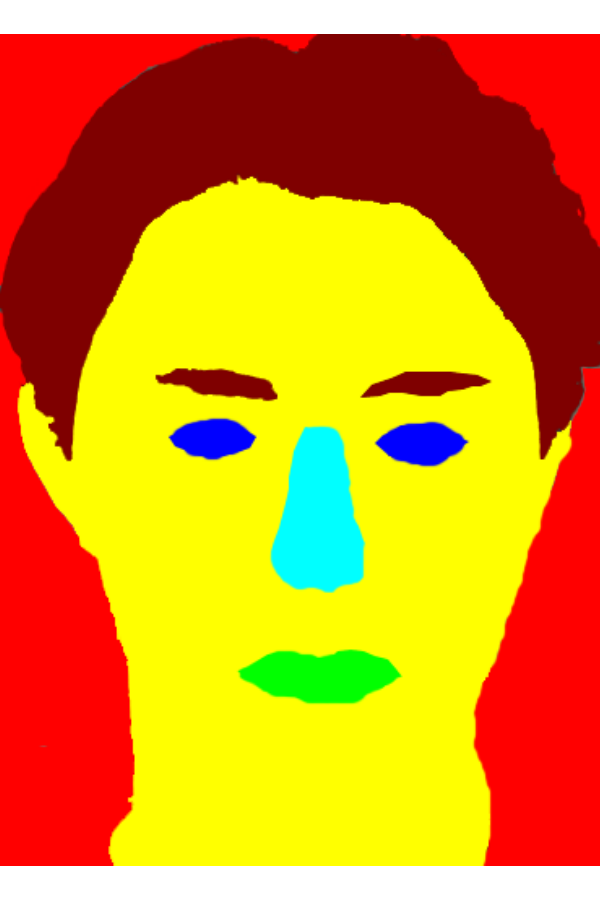}\\
      [-2mm]
      (a)&(b)
    \end{tabular}
  \end{center}
  \caption{Dataset for face segmentation (a) Example of an image from face segmentation dataset (b) Ground truth annotation for the given image. Different classes are indicated in different colors. Best viewed in color.}
\label{fig:06-facedatset}
\end{figure}

{\bf{Multi-class face segmentation}~~}
We apply the multi-class AL to the task of segmenting human faces~\cite{Khan15b}.
We  distinguish \num{6} classes: background, hair, skin, eyes, nose, and mouth.  
Fig.~\ref{fig:06-facedatset} demonstrates an example of an image from the dataset with the corresponding ground truth annotation. 
Notice again that we are dealing with unbalanced problem, obviously classes `eyes', `nose', `mouth' are a minority compared to `background', `skin' and `hair'. 
We use the same features as for the Weizmann    horse    database plus HOG features.

As in the case of multi-class MRI images, we must handle cases in which more than two classes are present in a single patch. 
However, this only happens in $0.84\%$ of the selected patches because three classes do no often co-occur in the same small neighborhood. 
Thus, we can still use the simple line separation heuristic depicted by Fig.~\ref{fig:01-planeInterface} in most iterations and leave a user an opportunity to use a standard brush for rare refined annotations.

\begin{figure}[]
\begin{center}
   \includegraphics[width=0.8\linewidth]{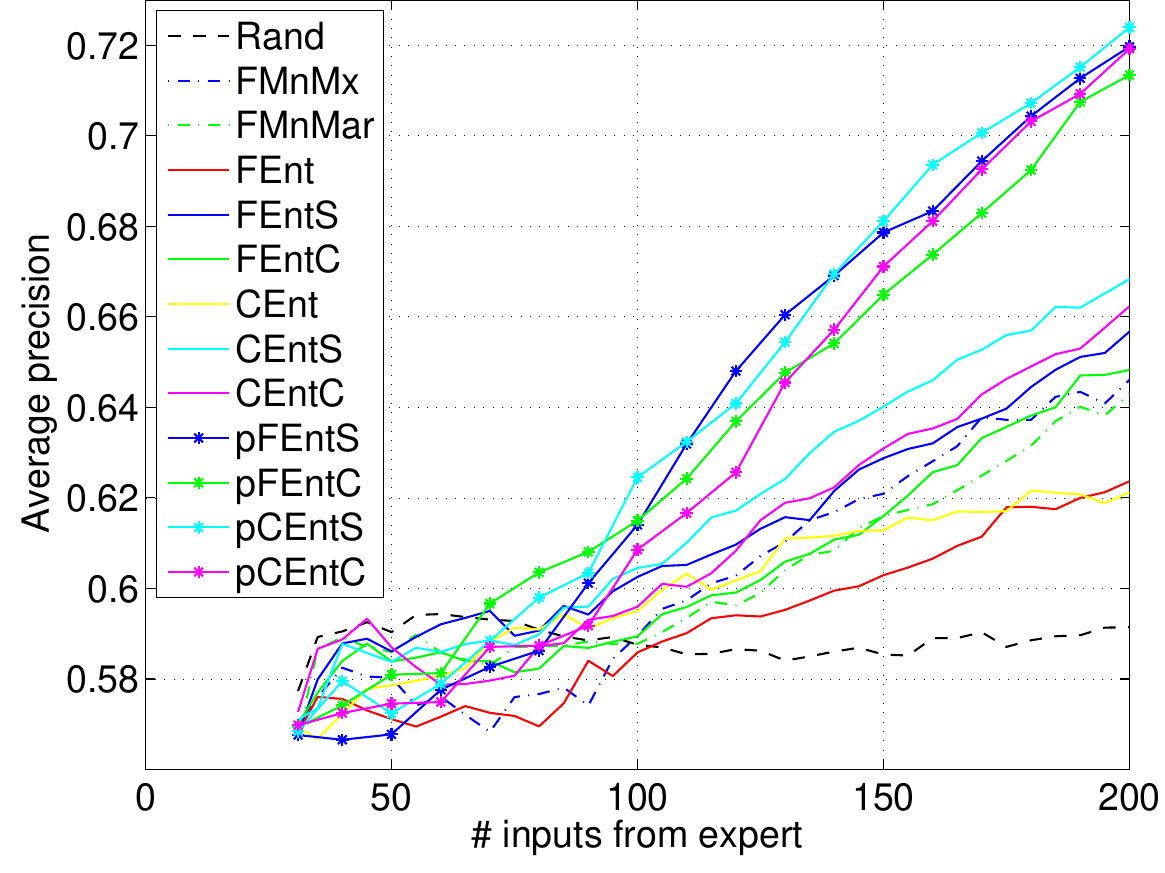}
\end{center}
\vspace{-5mm}
   \caption{Comparing several  AL strategies for multi-class face  segmentation. }
   \label{fig:06-faces}
\end{figure}

In Fig.~\ref{fig:06-faces} we  compare our  results to those  of the  baselines in terms of precision averaged over each one of the \num{6} classes.
This measure was chosen because it is better suited for capturing the performance in smaller classes and, thus reflects better the performance in segmentation with unbalanced classes. 
To ensure that performance on dominant classes is not sacrificed, we monitor the score of total precision (but omit the figure for conciseness), that shows similar performance for all AL strategies.
Entropy-based algorithms \FEntS{} and \FEntC{} are better than the standard \FMinMax{} and \FMinmargin{}, respectively.
Moreover, selection that is based on the entropy allows for a combination with \combun{} and brings further improvement in average precision with the strategies \CU{}, \CEntS{} and \CEntC{}.
Next, each of the strategies can be used in conjunction with patch selection that allows for further improvement in the learning rate. 
We show the patch selection results only for Selection Entropy and Conditional Entropy and skip Total Entropy as it performs poorly in total precision. 
As we can see, the combination of plane selection with \combun{} strategies demonstrates better results at the end of learning experiments with the best result obtained by \rpCEntS.

\subsection{Active learning or human intuition}
\label{sec:humanExperiments}

Before we conclude the experiments, we would like to motivate why AL is important and why we cannot rely on human selecting the data for annotation manually. 
First advantage of AL is that it eliminates the cognitive cost for a human user who would otherwise need to decide which datapoint is informative for a classifier. 
For this, the human annotator would need to have a good understanding of the underlying classification algorithm. 
Besides, in the next experiment we show an example that demonstrates that not all human-intuitive strategies are useful for a classifier.

To design a human-intuitive selection strategy, we study distances to the closest class boundary for selected samples. 
For this purpose we count how many samples lie within radius of \num{10} pixels from the boundary for \num{2} strategies: \RS{} and \CEntS{} in the face dataset.
We observe that \CEntS{} strategy samples \num{7.4}\% more datapoints in this area than \RS{}.
More superpixels in this area illustrate the effect of geometric component that prefers regions in the non-smooth areas of the prediction.
Then, an intuitive strategy could be to first label patches  at the boundary between  classes. 
We implemented the selection  strategies that  simulate such user behaviours and we refer to it as {\em boundary} strategy. 

To design another human-intuitive strategy, we notice that as part of the AL query selection procedure, we predict the segmentation for the whole training volume at every iteration.  
Given this prediction, a human expert could manually identify   patches  that are worth labelling. 
For example, he might  first correct the most obvious mistakes. 
Thus, we simulate such a strategy {\em max error} by selecting first  the most  confidently but  wrongly classified  samples.

\begin{figure}[H]
\begin{center}
   \includegraphics[width=0.8\linewidth]{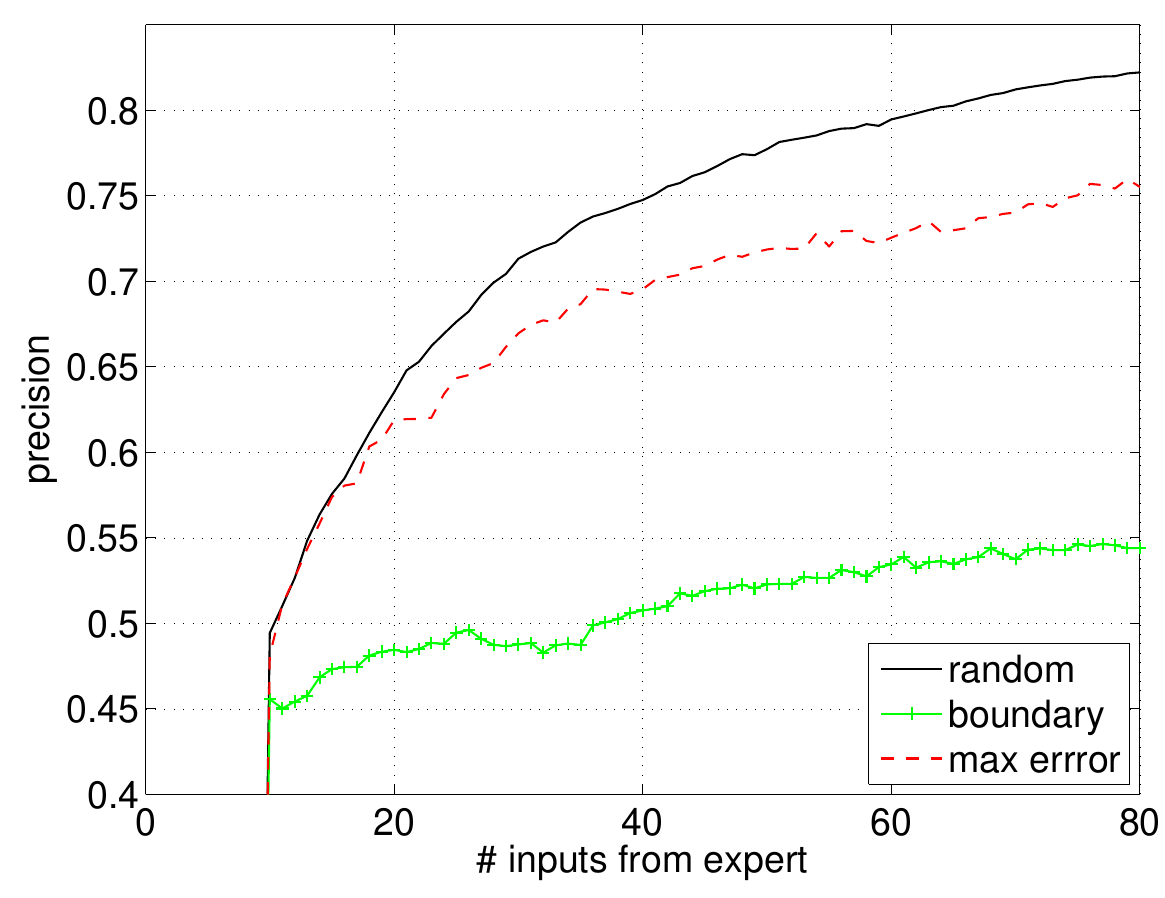}
\end{center}
\vspace{-5mm}
   \caption{Hypothetical human expert selection strategies. 
   We demonstrate that strategies that are intuitive for a human annotator do not result in better performance than passive sampling.}
   \label{fig:06-moral}
\end{figure}

We ran fifty  trials using each of these two  strategies on the face segmentation problem of  Sec.~\ref{sec:face}. Fig.~\ref{fig:06-moral} depicts the results. 
Surprisingly, the human strategies perform  much worse than even passive data selection,  that confirms  the  difficulty of  the  AL  problem.   The
heuristics we proposed  derive from our intuitive understanding  of the problem.
However,  applying these  intuitions is  not straightforward  for a  human user. 
For example, it turns out that selecting samples that have the highest error
leads to selecting outlier samples or those that have the most contradictory appearance.
Thus, intelligent and automated query selection is  necessary to determine how uncertain the
classifier is and  what smoothness prior should be used  when selecting the next
samples to be labelled.

\section{Conclusion}

In  this paper we introduced an approach to exploit image geometry priors to increase the effectiveness of AL in image segmentation application.  
We propose entropy-based uncertainty measures for multi-class classification that can be combined with geometric priors in a principled way.
In the segmentation of \num{2}D and \num{3}D images, our approach leverages the uncertainty information on the prediction at an image patch and at its neighbours. 
For \num{3}D image stacks, it adds an ability to select a \num{2}D planar patch where annotations are easier to perform.
We demonstrate the effectiveness of our approach on several datasets featuring MRI, EM, and natural images and both for  foreground-background and multi-class segmentation. 

We conclude that intuitions about geometrical properties of images are useful to answer the question {\em ``What data to annotate?''} in image segmentation. 
Besides, by reducing the annotation task from cumbersome \num{3}D annotations to \num{2}D annotations, we provide one possible answer to the question {\em ``How to annotate data?''}.
Moreover, we observe that addressing these two questions jointly can bring additional benefits to the annotation method.
Finally, we notice that the human intuitions may not always result in a desirable behaviour.

Our hand-crafted annotation strategy brings us impressive cost savings in image segmentation. 
Yet, we realise that it would be impossible to design a selection strategy for every new problem at hand. 
Besides, our last experiment shows that not all human intuitions perform as expected. 
To overcome these limitation and systematically search in the space of possible strategies, in future  work, we will strive to replace  the heuristics we have introduced by AL strategies that are themselves learned and take into account image priors in annotation process.
\vspace{10mm}
\section*{Acknowledgements}

This project has received funding from the European Union’s Horizon 2020 Research and Innovation Programme under Grant Agreement No. 720270 (HBP SGA1).
We would like to thank Lucas Maystre for asking many questions that helped to improve this article. It is important to note that the research on multi-class AL started as a semester project by Udaranga Wickramasinghe. Finally, we would like thank Carlos Becker for proofreading and comments on the text.

\bibliographystyle{IEEEtran}
\bibliography{string,vision,learning,biomed,optim,misc}

\end{document}